\documentclass[twoside,11pt]{article}

\usepackage{blindtext}

\usepackage[abbrvbib, preprint]{jmlr2e}

\usepackage[utf8]{inputenc} 
\usepackage{url}            
\usepackage{booktabs}       
\usepackage{nicefrac}       
\usepackage{microtype}      
\usepackage{xcolor}         
\usepackage{subcaption}     
\usepackage{amsmath}        

\usepackage{caption}
\usepackage{lastpage}
\jmlrheading{23}{ September 2025}{1-\pageref{LastPage}}{1/21; Revised 5/22}{9/22}{21-0000}{Lukas Silvester Barth and Paulo von Petersenn}

\ShortHeadings{Probabilistic and nonlinear compressive sensing}{Barth and von Petersenn}
\firstpageno{1}

\begin{document}

\title{Probabilistic and nonlinear compressive sensing}

\author{\name Lukas Silvester Barth \email lukas.barth@mis.mpg.de \\
       \addr Max Planck Institute for Mathematics in the Sciences\\
       Leipzig, Germany
       \AND
       \name Paulo von Petersenn \email vonpetersenn@mis.mpg.de \\
       \addr Max Planck Institute for Mathematics in the Sciences\\
       Leipzig, Germany}

\editor{My editor}

\maketitle

\begin{abstract}
We present a smooth probabilistic reformulation of $\ell_0$ regularized regression that does not require Monte Carlo sampling and allows for the computation of exact gradients, facilitating rapid convergence to local optima of the best subset selection problem. The method drastically improves convergence speed compared to similar Monte Carlo based approaches. 
Furthermore, we empirically demonstrate that it outperforms compressive sensing algorithms such as IHT and (Relaxed-) Lasso across a wide range of settings and signal-to-noise ratios. 
The implementation runs efficiently on both CPUs and GPUs and is freely available at \href{https://github.com/L0-and-behold/probabilistic-nonlinear-cs}{github.com/L0-and-behold/probabilistic-nonlinear-cs} .

We also contribute to research on nonlinear generalizations of compressive sensing by investigating
when parameter recovery of a nonlinear teacher network is possible through compression of a student network. 
Building upon theorems in \citep{fefferman1993recoveringNNfromOutputs}, we show theoretically that the global optimum in the infinite-data limit enforces recovery up to certain symmetries. 
For empirical validation, we implement a normal-form algorithm that selects a canonical representative within each symmetry class. 
However, while compression can help to improve test loss, we find that exact parameter recovery is not even possible up to symmetries.
In particular, we observe a surprising rebound effect where teacher and student configurations initially converge but subsequently diverge despite continuous decrease in test loss.
These findings indicate fundamental differences between linear and nonlinear compressive sensing.
\end{abstract} 


\section{Introduction and background}
\label{sec:introduction}

This section introduces the historical and mathematical background of compressive sensing and its nonlinear generalizations. We motivate and outline our contributions while explaining differences from related work.

\subsection{Compressive sensing}
\label{eq:compressiveSensing}

Compressive sensing \citep{OrthogonalMatchinPursuit1993,tibshirani1996regression,CandesTao, Donoho, blumensath2008iterative,foucart_mathematical_2013} explores techniques for reconstructing signals from far fewer data samples than generally required by the Nyquist-Shannon sampling theorem \citep{ShannonNyquist}. 
The key insight is exploiting the assumption that signals have (at least approximately) sparse representations in some basis of the signal's function space.

This research has found applications across many important domains: 
image processing \citep{metzler2016denoising,otazo2015low}, 
audio and speech processing \citep{abrol2015voiced,giacobello2009retrieving,george2015audio}, 
wireless communication \citep{liu2014cdc}, 
power amplifiers \citep{chen2017nonlinear}, 
face recognition \citep{nagesh2009compressive,qiao2010sparsity}, 
manifold learning \citep{davenport2010joint}, 
nano-scale circuits \citep{zhang2011virtual}, 
genetic analysis \citep{dai2008compressive,tang2011compressed}, 
LiDAR \citep{Howland2011Photon}, 
and many others \citep{rani2018systematic}. 
More recently, it has been applied to sparsification of neural ODEs and neural PDEs \citep{sahoo2018learning,brunton2019,datadrivendiffeq2021,maddu2022stability}, enabling the learning of sophisticated dynamical systems.

The compressive sensing problem is typically formulated as linear regression with sparsity constraints or regularization. 
To see this, consider a function $f:[0,1] \to \mathbb{R}$ (more general domains are possible) corresponding to a signal, and let $\{f_j\}_{j\in \mathbb{N}}$ be a basis of functions over the domain of $f$. We say $f$ is approximately sparse in this basis if there exists a comparatively small set $J$ and coefficients $\{\theta_j\}_{j\in J}$ such that $f(x) \approx \sum_{j\in J}\theta_j f_j(x)$ for all $x$.

Given $n$ samples $\{(x_i,f(x_i))\}_{i\in \{1,\ldots,n\}}$ without knowing $f$ explicitly, we can attempt to reconstruct $f$ by fixing a large finite set $K$ with $|K|=p$ that we assume to contain $J$, then finding the minimizer of:
\begin{equation}
    \begin{split}
        L_\lambda(\theta) := \sum_i \bigg(f(x_i)-\sum_{j\in K} \theta_j f_j(x_i) \bigg)^2 + \lambda \ell_0(\theta),
    \end{split}
    \label{eq:compressiveSensingObjective}
\end{equation}
where $\ell_0(\theta)=|\{~i~|~\theta_i\ne 0~\}|$ is the ``norm'' that counts the number of non-zero components of $\theta$. Defining the matrix $F$ by $F_{ij}:=f_j(x_i)$ and the vector $y$ by $y_i:=f(x_i)$, eq.~\eqref{eq:compressiveSensingObjective} becomes
\begin{equation}
    \begin{split}
        L_\lambda(\theta) = ||y-F\theta||_2^2 + \lambda \ell_0(\theta),
    \end{split}
    \label{eq:RegLinReg}
\end{equation}
the aforementioned regularized linear regression problem.

Several variations of this problem exist. Equation \eqref{eq:RegLinReg} is the ``Lagrangian form'' of the constrained optimization problems:
\begin{equation}
    \begin{split}
        \text{minimize}_\theta\quad ||y-F\theta||_2^2 \quad\text{subject to}\quad \ell_0(\theta)\leq k,
    \end{split}
    \label{eq:bestSubset}
\end{equation}
\begin{equation}
    \begin{split}
        \text{minimize}_\theta\quad \ell_0(\theta) \quad\text{subject to}\quad ||y-F\theta||_2^2 \leq \eta.
    \end{split}
    \label{eq:bestL0}
\end{equation}
Problem \eqref{eq:bestSubset} is the ``best subset selection'' problem, dating back to at least \citep{Beale1967} and \citep{hocking1967selection} according to \citep{hastie2017extended}. 
Problem \eqref{eq:bestL0} is the ``$\ell_0$ minimization problem,'' shown to be NP-hard in \citep{natarajan1995sparse} and \citep[Section 2.3]{foucart_mathematical_2013}. 
While problems \eqref{eq:bestSubset} and \eqref{eq:bestL0} are equivalent to each other\footnote{
To see this, let $\theta^*(k)$ be the global minizer of \eqref{eq:bestSubset} with the smallest $\ell_0$ norm among all global minimizers. Then, if we solve \eqref{eq:bestL0} with $\eta = ||y-F\theta^*(k)||_2^2$, one global optimum will be at $\theta^*(k)$.
Conversely, let $\theta^*(\eta)$ be the global minimizer of \eqref{eq:bestL0} with the smallest least-squares-error among all global minimizers. Then, if we solve \eqref{eq:bestSubset} with $k=\ell_0(\theta^*(\eta))$, one global optimum will be at $\theta^*(\eta)$. 
}, they are generally non-equivalent\footnote{However, one could amend the Lagrangian form to make them equivalent, either by solving a minimax problem, $\min_\theta \max_{u\ge 0} ||y-F\theta||_2^2 + u (\ell_0(\theta)-k)$, or by changing the Lagrangian form to $M_\lambda^k(\theta)=||y-F\theta||_2^2 + \lambda \max(0,\ell_0(\theta)-k)$. In the latter case, whenever $\lambda > ||y-F\theta^*(k)||_2^2$, where $\theta^*(k)$ is a global minimizer of \eqref{eq:bestSubset}, one global minimizer of $M_\lambda^k(\theta)$ becomes $\theta^*(k)$.} to \eqref{eq:RegLinReg} due to the non-convexity of the $\ell_0$ norm \citep{bertsimas2016best}.

A convex relaxation of the problems above can be obtained by replacing the $\ell_0$ norm with the $\ell_1$ norm, as done in the \textit{Lasso} (least absolute shrinkage and selection operator), cf.~\citep{tibshirani1996regression}. By convex duality, the resulting $\ell_1$ relaxations of \eqref{eq:RegLinReg}, \eqref{eq:bestSubset}, and \eqref{eq:bestL0} are all equivalent, cf.~\citep[Proposition 3.2]{foucart_mathematical_2013}. One major achievement of compressive sensing was proving that the convex $\ell_1$ relaxation can recover the true support set of $\theta$ with high probability under certain assumptions. 
For example, \citep[Remark around eq.~(9.40)]{foucart_mathematical_2013} present an impressive estimate for Gaussian matrices $F \in \mathbb{R}^{n\times p}$: If $\theta\in \mathbb{R}^p$ has at most $s$ non-zero entries, then for large $p$ and moderately large $s$, recovery of the correct support set can be guaranteed with very high probability when $n > 2s \ln(p/s)$. More restrictive results hold for non-Gaussian matrices.

Nevertheless, even when the correct support set can be recovered, the shrinkage effect of the $\ell_1$ norm introduces bias on the weights $\theta$ during optimization. As summarized in \citep{yin2020probabilistic}, ``\citep{johnson2015risk} show that the predictive risk of $\ell_1$ regularized linear regression cannot outperform $\ell_0$ regularized regression by more than a small constant factor and in some cases is infinitely worse under assumptions of the design matrix.'' 
Furthermore, \citep{meinshausen2007relaxed} demonstrated that $\ell_1$ regularized regression lacks posterior consistency--specifically, ``it is impossible to have both consistent variable selection and optimal rates for independent predictor variables''--and instead proved consistency for an amended version called the \textit{Relaxed Lasso}. 
A simplified version of the latter also exhibited superior empirical performance in \citep{hastie2017extended}. 

However, even the Relaxed Lasso inherits several disadvantages from the original Lasso estimator: (1) Support set recovery is only guaranteed under certain conditions that can be quite restrictive and difficult to verify \citep{buhlmann2011statistics};
    (2) The optimization path depends heavily on sampled data, leading to unstable estimation \citep{meinshausen2010stability};
    (3) As \citep{bertsimas2016best} note, ``in order to deliver a model with good predictive accuracy, Lasso brings in a large number of nonzero coefficients including noise variables. [...] Upon increasing the degree of regularization, Lasso sets more coefficients to zero, but in the process ends up leaving out true predictors from the active set.'';
    (4) $\ell_1$ regularization lacks the favorable statistical properties of $\ell_0$ regularization \citep{foster1994risk, greenshtein2006best,zhang2014lowerboundsperformancepolynomialtime}.

For these reasons, there remains ongoing interest in efficient procedures that can find good local optima of the original $\ell_0$ regularized regression problems. 
The established \textit{Forward Stepwise} (FS) method \citep{efroymson1966stepwise,draper1998applied} easily gets stuck in poor local optima and performs poorly in low signal-to-noise regimes \citep{hastie2017extended}. Furthermore, we show in Section \ref{sec:runtimeBenchmarks} that its runtime scales nonlinearly with the sample size. 
The more recent \textit{Iterative Hard Thresholding (IHT)} \citep{blumensath2008iterative} can also become trapped in local minima when the Restricted Isometry Property (RIP) does not hold and also has difficulties in low signal-to-noise regimes. Furthermore, it is quite sensitive to initial conditions and step size \citep{cartis2014newimprovedquantitativerecovery}. 
While \citep{bertsimas2016best} used \textit{mixed integer optimization (MIO)} methods to significantly accelerate global solution procedures, their method struggles with lower signal-to-noise ratios \citep{hastie2017extended} and remains far too slow for practical use with larger datasets. 

Recent progress in scalable non-global solution procedures has been made by \citep{yin2020probabilistic} and \citep{kunes2023gradient}, who developed a probabilistic reformulation of \eqref{eq:RegLinReg} that makes the problem amenable to gradient descent. Nevertheless, their procedure requires Monte Carlo sampling, which leads to much slower convergence than $\ell_1$ regularization methods.

\subsection{Nonlinear generalizations}
\label{sec:NonlinearGeneralizations}

As shown in \citep{barth2025efficientcompression}, the regularized regression problem can be understood as an approximation of a special case of the minimum description length principle \citep{rissanen1978modeling}. Since minimum description length is the maximum a posteriori estimator of Solomonoff's predictive prior \citep{solomonoff1964formal,poland2004convergence}, compressive sensing can be subsumed under Solomonoff induction \citep{solomonoff1978complexity,hutter2005universal}, where one exploits the low complexity of a data-generating distribution to reconstruct it from few samples. 

In essence, if the data generating distribution is assumed to be Gaussian with linear mean and constant variance, then description length minimization reduces to \eqref{eq:RegLinReg}. To see this, note that the description length of data $y$ given $x$, when encoded via arithmetic coding \citep{pasco1976source,RissanenArithmeticCoding} with the conditional distribution $\mu_\theta$ (parameterized by $\theta\in\mathbb{R}^p$), is given by 
\begin{equation}
    \begin{split}
        L_\lambda(\theta) := \lambda \ell(\mu_\theta) - \log_2(\mu_\theta(y|x)),
    \end{split}
    \label{eq:minDescLength}
\end{equation}
where $\ell$ measures the length of the program that computes the distribution $\mu_\theta$. To approximate $\ell$ by $\ell_0$ is often a good choice because if we assume that parameters are independently drawn from some distribution, then statistically the complexity scales with the number of parameters. Furthermore, in comparison to most smooth regularizers, the $\ell_0$ norm does not induce any shrinkage on the weights as explained above.
If we assume that the parameterized distribution is a Gaussian family and approximate $\ell$ by $\ell_0$, eq.~\eqref{eq:minDescLength} becomes:
\begin{equation}
    \begin{split}
        L_\lambda(\theta) = \lambda \ell_0(\theta) + \frac{1}{2}\log_2(2\pi\sigma^2) + \frac{||y-F_\theta(x)||_2^2}{2\sigma^2\ln(2)}.
    \end{split}
    \label{eq:GaussianLoss}
\end{equation}
Here the mean $F_\theta$ can be a nonlinear function of the parameters $\theta$ (and of the input $x$), such as a neural network. However, when we hold the standard deviation $\sigma$ constant during optimization and assume $F_\theta$ to be linear in $\theta$, i.e.~$F_\theta(x)=\sum_j \theta_j f_j(x)$, we recover \eqref{eq:compressiveSensingObjective} and \eqref{eq:RegLinReg} for appropriate $\lambda$. 
Thus we consider \eqref{eq:GaussianLoss} a conceptually clean generalization of compressive sensing, applicable when the conditional distribution, which generates the observed data, is assumed to be a family of Gaussians.
Even more generally, one can consider the description length for arbitrary distributions, as mentioned before but for us \eqref{eq:GaussianLoss} remains the relevant case here.
In \citep{barth2025efficientcompression}, we refined and compared multiple methods that can find local optima of (approximations of) objective \eqref{eq:GaussianLoss}. We briefly summarize these methods in Section \ref{sec:nonlinearL0Regularization} and use them for a number of nonlinear compressive sensing experiments in Subsection \ref{sec:TeacherStudentExperiments}.

Having identified this generalization, a natural question arises: to what extent do the strong reconstruction results from linear regression theory translate to the nonlinear problem \eqref{eq:GaussianLoss}? Results from Solomonoff induction theory and minimum description length learning alone cannot answer this question because they only guarantee that a posterior distribution ultimately converges to the true distribution (posterior consistency) but do not guarantee recovery of the model parameters that generated the data. 

Several other articles have explored certain forms of nonlinear generalizations of compressive sensing. In \citep{chen2017nonlinear}, the authors considered objectives of the form $\lambda \ell(\theta) + ||y-g(F \theta)||_2^2$, where $F$ is a matrix and $g$ is a ``nonlinear distortion function.'' In their experiments, they used $g(x)=x/(1+(x/\alpha)^{2\beta})^{1/(2\beta)}$ and optimized the objective using proximal gradient descent. However, this approach is less general than \eqref{eq:GaussianLoss} because in $g(F \theta) = g(\sum_j \theta_j f_j(x))$, the weights still act linearly on the basis functions, with nonlinearity applied only to the result.

In \citep{blumensath2013nonlinear}, the author considers a generalization where $y$ is assumed to equal $\Phi(\theta)$ up to noise, with $\Phi$ being any nonlinear function. In contrast to \eqref{eq:GaussianLoss}, there is no explicit dependence on the input $x$.\footnote{Note that \citep{blumensath2013nonlinear} uses $x$ instead of $\theta$, but we maintain our original notation where $x$ denotes the input and $\theta$ represents the optimized weights.} However, for fixed $x$, we could potentially absorb this into the definition of $\Phi$ via the correspondence $\Phi(\theta)=F_\theta(x)$.
Instead of imposing direct regularization constraints on $\theta$, this approach assumes $\theta$ to lie within some arbitrary closed subset $\mathcal{A}$. 
The suggested optimization algorithm is, again, a form of proximal gradient descent, where the proximal operator projects the gradient back to the feasible region $\mathcal{A}$. \citep{blumensath2013nonlinear} proves that $\theta$ converges to the true solution $\theta^*$ under the constraint that the linearization $\Phi_{\theta^*}$ of $\Phi$ for all $\theta_1,\theta_2,\theta^*\in \mathcal{A}$ satisfies the so-called Restricted Isometry Property (RIP):
\begin{equation}
    \label{eq:RIPcondition}
    \begin{split}
        \alpha ||\theta_1-\theta_2||_2^2 \le ||\Phi_{\theta^*}(\theta_1-\theta_2)||_2^2 \le \beta ||\theta_1-\theta_2||_2^2 ,
    \end{split}
\end{equation}
where $\Phi_{\theta^*}$ is the matrix corresponding to the differential of $F$ at $\theta^*$. 
However, we will show that this condition is far too restrictive when $F_\theta$ is as nonlinear as, for example, a neural network.

Finally, recovery of nontrivial nonlinear behavior can be achieved by applying compressive sensing techniques to the right-hand side of dynamical systems or PDEs, as in \citep{brunton2014compressive, sahoo2018learning,brunton2019,datadrivendiffeq2021,maddu2022stability}. 
For example, if we assume that data approximately lies on the solution of an evolutionary equation of the form
\begin{equation}
    \begin{split}
        z'(x) \equiv
         \frac{dz(x)}{dx} = f(z),
    \end{split}
    \label{eq:ODE}
\end{equation}
then, if $z$ and $f$ are unknown but we are given samples $\{(x_i, z'(x_i))\}_{i\in\{1,\cdots,n\}}$, we can attempt to reconstruct $f$ (and then integrate to obtain $z$) by expanding $f$ in an appropriate function basis and reduce the problem to regularized regression of the form \eqref{eq:compressiveSensingObjective}.

However, for this approach to work, the right-hand side $f$ must remain linear in the optimized coefficients, and one either needs direct access to samples of the derivatives of the dependent variable $\{z'(x_i)\}_i$ or must use collocation methods to estimate derivatives \citep{RoeschRackauckasStumpf} from samples $\{z(x_i)\}_i$. While these methods are highly interesting, they ultimately reduce to the linear case.

\subsection{Our contributions}

As explained in Section \ref{eq:compressiveSensing}, there is ongoing interest in obtaining more efficient procedures to approximately solve the $\ell_0$ regularized regression problem. In Section \ref{sec:probReformulation}, we prove the existence of a closed-form solution to the probabilistic reformulation originally proposed in \citep{yin2020probabilistic}. This allows us to compute exact gradients without Monte Carlo sampling.

In Section \ref{sec:ComparisonWithMonteCarloMethods}, we demonstrate that this approach drastically improves convergence speed compared to \citep{yin2020probabilistic} and \citep{kunes2023gradient}.
Moreover, in Section \ref{sec:compressiveSensingExperiments}, we compare our method with \textit{Iterative Hard Thresholing (IHT)} \citep{blumensath2008iterative}, \textit{Forward Stepwise} \citep{efroymson1966stepwise}, \textit{Lasso} \citep{tibshirani1996regression}, and \textit{Relaxed Lasso} \citep{meinshausen2007relaxed}, using the implementations of \citep{chu2020IHT} and \citep{hastie2017extended}. The comparison shows that our method performs best across a wide range of settings and signal-to-noise ratios.

In Section \ref{sec:FeffermanMarkelTheory}, we contribute to the theory of active set recovery for the nonlinear objective \eqref{eq:GaussianLoss}, assuming for the first time to our knowledge that $F_\theta$ corresponds to a fully nonlinear class of Multi-Layer Perceptrons (MLPs) with arbitrarily many layers and tanh activation functions. Building upon Fefferman and Markel's work \citep{fefferman1993recoveringNNfromOutputs}, we show that in the large data limit and under certain ``generic'' conditions, parameter recovery is guaranteed up to certain symmetries of the network architecture.
In Section \ref{sec:TeacherStudentExperiments}, we present experiments using a student-teacher setup that allows us to investigate whether the student can recover the teacher's parameters in practice. We use a normal-form algorithm to select a canonical representative within each symmetry class to ensure fair comparison up to symmetries. We can indeed observe that compression can help to improve the test error. However, even accounting for symmetries, we observe a surprising rebound effect that indicates a fundamental difference from the theory in the large data limit. This interesting phenomenon suggests that it may not be possible to prove reconstruction theorems as strong as those in the linear case.
\section{Probabilistic compressive sensing}
\label{sec:probReformulation}

In \citep{yin2020probabilistic}, it was shown that 
\begin{equation}
    \begin{split}
        &\min_{\theta\in\mathbb{R}^p}\left\{||y-F\theta||_2^2 
        + \lambda \ell_0(\theta) \right\} 
        = \min_{\gamma\in [0,1]^p,~w\in \mathbb{R}^p}\mathbb{E}_{z\sim\pi_\gamma}\left\{||y-F(wz)||_2^2 
        + \lambda \ell_0(z) \right\},
    \end{split}
    \label{eq:linearcase}
\end{equation}
where $\pi_\gamma$ is the Bernoulli distribution over $\{0,1\}^p \ni z$, parameterized by $\gamma \in [0,1]^p$, and $wz$ denotes componentwise multiplication.

The advantage of the right-hand side is that optimization is performed over parameters with respect to which the objective is piecewise differentiable, making it amenable to gradient descent in principle. However, writing out the expected value results in a sum with $2^p$ terms (since $z\in\{0,1\}^p$), meaning the combinatorial explosion that makes the problem NP-hard still occurs, albeit in a different form. Nevertheless, this reformulation enables approximation of the sum via Monte Carlo sampling.

This can be accomplished using the following well-known identity:
\begin{equation}
    \begin{split}
        \frac{\partial}{\partial \gamma_k}\mathbb{E}_{z\sim \pi_\gamma}[g(z)] &= \mathbb{E}_{z\sim \pi_\gamma}\left[g(z)\frac{\partial \log \pi_\gamma(z)}{\partial \gamma_k}\right].
    \end{split}
    \label{eq:derivPiAlternative}
\end{equation}
Monte Carlo methods estimate this quantity by sampling $K \ll 2^p$ right-hand side terms. This estimator is also used for policy gradient learning in reinforcement learning, where it is known as the \textit{Reinforce} estimator \citep{williams1992simple}. However, a standard limitation of such sample estimators is their potentially high variance. To address this issue, the DisARM estimator was developed (apparently independently) in \citep{yin2020probabilistic} and \citep{dong2020disarm}, which can be viewed as the Rao-Blackwellization of ARM \citep{yin2018arm}. Nevertheless, DisARM performs poorly near the boundaries of the parameter space, where its variance explodes and its gradients become so sparse that convergence is prevented. This poses a significant problem in compressive sensing and $\ell_0$ regularization, where operation close to these boundary regions is precisely what is required most of the time. Consequently, \citep{kunes2023gradient} introduced the BitFlip estimator, which offers low variance even at boundaries but produces very sparse gradients, thus requiring many function evaluations. Therefore, all these methods (and even their combinations) suffer from either imprecision or high computational costs. 

We introduce a method that eliminates the need for Monte Carlo sampling entirely. To begin, note that $\mathbb{E}_{z\sim \pi_\gamma}[\sum_i a_i z_i] = \sum_i a_i \gamma_i$. Furthermore, since $\ell_0(z)=\sum_i z_i$, we find that
\begin{equation}
    \begin{split}
        \mathbb{E}_{z \sim \pi_\gamma}[\ell_0(z)] = \sum_{i} \gamma_i,
    \end{split}
    \label{eq:l0probterm}
\end{equation}
a result already used in \citep{louizos2017learning}. However, when considering the expectation value of nonlinear functions of $z$, a similar closed form cannot be derived. What has not been observed previously, to our knowledge, is that an efficient reformulation can still be achieved when the terms within the expectation are at most quadratic, as in \eqref{eq:linearcase}:
\begin{theorem}
    \label{lem:quadratic}
    \begin{equation*}
        \begin{split}
            &\mathbb{E}_{z\sim\pi_\gamma}\bigg[\sum_{i}\bigg(y_{i} - \sum_j F_{ij}w_jz_j\bigg)^2~\bigg] 
            = \sum_{i}\bigg(y_{i} - \sum_j F_{ij}w_j\gamma_j\bigg)^2 
            + \sum_{i,j} F_{ij}^2 w_j^2 \gamma_j (1-\gamma_j).
        \end{split}
    \end{equation*}
\end{theorem}

\begin{proof}
    This proof employs index notation. For readers who prefer matrix notation, we provide an alternative proof in Appendix \ref{app:alternativeProof}.
    
    We first expand the square:
    \begin{equation}
        \begin{split}
            \sum_{i}\bigg(y_{i} - \sum_j F_{ij}w_jz_j\bigg)^2
            = \sum_i \bigg(y_i^2 - 2y_i \sum_j F_{ij}w_jz_j + \bigg(\sum_j F_{ij}w_jz_j\bigg)^2\bigg).
        \end{split}
    \end{equation}
    Next, we exchange $\mathbb{E}_{z\sim\pi_\gamma}$ with the linear sums and use the fact that $\mathbb{E}_{z\sim\pi_\gamma}(z_j)=\gamma_j\cdot 1 + (1-\gamma_j)\cdot 0=\gamma_j$ to obtain
    \begin{equation}
        \begin{split}
            &\mathbb{E}_{z\sim\pi_\gamma}\bigg[\sum_{i}\bigg(y_{i} - \sum_j F_{ij}w_jz_j\bigg)^2~\bigg] \\
            &= \sum_i \left(y_i^2 - 2y_i \sum_j F_{ij}w_j\gamma_j + \mathbb{E}_{z\sim\pi_\gamma}\bigg[\bigg(\sum_j F_{ij}w_jz_j\bigg)^2~\bigg]\right)
        \end{split}
        \label{eq:proofLemma23_eq_1}
    \end{equation}
    For the remaining expectation over $\big(\sum_j F_{ij}w_jz_j\big)^2$, note that the terms will be of the form $F_{ij_1}F_{ij_2}w_{j_1}w_{j_2}z_{j_1}z_{j_2}$. Whenever $j_1\ne j_2$, then such terms are again separately linear in $z_{j_1}$ and $z_{j_2}$ and can be replaced by $\gamma_{j_1}\gamma_{j_2}$ terms.\footnote{To see this, note that $\mathbb{E}_{z\sim\pi_\gamma}[z_{j_1}z_{j_2}] = \gamma_{j_1}\gamma_{j_2} \cdot 1\cdot 1 + (1-\gamma_{j_1})\gamma_{j_2}\cdot 0\cdot 1 + \gamma_{j_1}(1-\gamma_{j_2})\cdot 1\cdot 0 + (1-\gamma_{j_1})(1-\gamma_{j_2})\cdot 0\cdot 0 = \gamma_{j_1}\gamma_{j_2}$.
    }
    Only when $j_1=j_2$ do we get $\mathbb{E}_{z\sim\pi_\gamma}[z_{j_1}z_{j_2}] = \mathbb{E}_{z\sim\pi_\gamma}[z_{j_1}^2] = \gamma_{j_1}\cdot 1^2 + (1-\gamma_{j_1})
    \cdot 0^2 = \gamma_{j_1}$, meaning the expectation linearizes the squares. If the sum over $j$ has $p$ terms, then the square of the sum has exactly $p$ terms of the form $z_{j_k}^2$, and we therefore obtain
    \begin{equation}
        \begin{split}
            \mathbb{E}_{z\sim\pi_\gamma}\bigg[\bigg(\sum_j F_{ij}w_jz_j\bigg)^2~\bigg] = \bigg(\sum_j F_{ij}w_j\gamma_j\bigg)^2 - \sum_j F_{ij}^2 w_j^2 \gamma_j^2 + \sum_j F_{ij}^2 w_j^2 \gamma_j
        \end{split}
    \end{equation}
    When reinserting this into \eqref{eq:proofLemma23_eq_1}, we can recombine the first three terms of the result into a square and finally obtain
    \begin{equation}
        \begin{split}
            \mathbb{E}_{z\sim\pi_\gamma} & \bigg[\sum_{i}\bigg(y_{i} - \sum_j F_{ij}w_jz_j\bigg)^2~\bigg] \\ 
            &= \sum_{i}\left(\bigg(y_{i} - \sum_j F_{ij}w_j\gamma_j\bigg)^2 + \sum_{j} F_{ij}^2w_j^2\gamma_j(1-\gamma_j)\right)
        \end{split}
    \end{equation}
\end{proof}

Note that this approach also works with higher-order polynomials: one can always replace the higher powers with linear terms. However, the higher the power of the sum, the more terms arise. In particular, for analytic functions with infinite Taylor series or non-polynomial nonlinearities, the technique is not applicable or useful.

The crucial point is that we end up with merely twice the number of summands instead of exponentially more. Combining \eqref{eq:l0probterm} with Theorem \ref{lem:quadratic} allows us to derive the following result:

\begin{corollary}
    \label{cor:EGP}
    \begin{equation}
        \begin{split}
            &\min_{\theta\in\mathbb{R}^p}\left\{||y-F\theta||_2^2 
            + \lambda \ell_0(\theta) \right\} \\
            &= \min_{w\in \mathbb{R}^p, ~\gamma\in [0,1]^p} 
            ||y - F(w\gamma)||_2^2 
            + \sum_{i,j} F_{ij}^2 w_j^2 \gamma_j (1-\gamma_j)
            + \lambda \sum_{i}\gamma_i.
        \end{split}
        \label{eq:EGP}
    \end{equation}
\end{corollary}

This result now allows us to directly apply stochastic gradient descent (SGD) to the objective in \eqref{eq:EGP} without any sampling. Our empirical results in Section \ref{sec:ComparisonWithMonteCarloMethods} show that this approach is drastically faster than Monte Carlo-based methods. 
To keep $\gamma$ within $[0,1]^p$, we simply clip it to this interval after every gradient step, which turns out to be highly effective.
Standard theorems for SGD then guarantee convergence to a local optimum of this procedure, cf.~\citep{SGDconvergence}.

One significant advantage of this method is that, since SGD is the standard optimization procedure in deep learning, a vast amount of efficient optimization algorithms has been developed that run on arbitrarily many GPUs and efficiently handle large amounts of data. We call this new method for compressive sensing \textbf{Exact Gradient Pruning (EGP)} because it is, to our knowledge, the only method that allows for the computation of exact gradients of the objective in \eqref{eq:RegLinReg}.

\subsection{Combination with \texorpdfstring{$\ell_1$}{L1} and \texorpdfstring{$\ell_2$}{L2} norms}
\label{sec:L1L2combinationsEGP}

Our method can be easily combined with $\ell_1$  and $\ell_2$-regularization. To this end, note that
\begin{equation}
    \begin{split}
        \mathbb{E}_{z \sim \pi_\gamma}[\ell_1(wz)] &= \mathbb{E}_{z \sim \pi_\gamma}\bigg[\sum_i|w_iz_i|\bigg]= \sum_i|w_i|~\mathbb{E}_{z \sim \pi_\gamma}[z_i]
        = \sum_i|w_i|\gamma_i.
    \end{split}
    \label{eq:PL1}
\end{equation}
For the (squared) $\ell_2$-norm, we obtain
\begin{equation}
    \begin{split}
        \mathbb{E}_{z \sim \pi_\gamma}\left[\ell_2(wz)\right] &=  \sum_iw_i^2~\mathbb{E}_{z \sim \pi_\gamma}\left[z_i^2\right]=  \sum_iw_i^2\gamma_i.
    \end{split}
    \label{eq:PL2}
\end{equation}
Combining these equations with Corollary \ref{cor:EGP} yields the elegant expression:
\begin{equation}
    \begin{split}
        \min_{\theta \in \mathbb{R}^p} L_{\lambda_0,\lambda_1,\lambda_2}(\theta) := || y - & F\theta||_2^2 + \lambda_0 \ell_0(\theta)+ \lambda_1 \ell_1(\theta)+ \lambda_2 \ell_2(\theta) \\
        \overset{\eqref{eq:EGP},\eqref{eq:PL1},\eqref{eq:PL2}}{=} & \min_{w\in \mathbb{R}^p, ~\gamma\in [0,1]^p} 
        ||y - F(w\gamma)||_2^2
        + \sum_{i,j} F_{ij}^2 w_j^2 \gamma_j (1-\gamma_j)\\
        & \qquad +  \sum_{i}\gamma_i\left(\lambda_0 + \lambda_1|w_i| + \lambda_2w_i^2\right).
    \end{split}
\end{equation}
Since $|w_i|$ and $w_i^2$ are piecewise differentiable, one can simply apply gradient descent to the entire expression to find a local optimum.

This seamless combination enables the simultaneous utilization of the strengths of all three norms if desired. The $\ell_1$ norm can help in noisy environments by preventing excessively large weights, while the $\ell_2$-norm can ensure that weights do not have overly disparate magnitudes. However, since these norms introduce shrinkage bias, one should use rather small $\lambda$ coefficients during optimization. The $\ell_0$ norm then ensures that weights become exactly zero and provides unbiased sparsity.

\subsection{Convergence to good local optima}
\label{sec:convToGoodOptima}

Even though Corollary \ref{cor:EGP} provides an exciting new approach to optimize \eqref{eq:RegLinReg}, the objective \eqref{eq:EGP} remains highly non-convex and is generally equivalent to an NP-hard problem. To address this challenge and ensure convergence to good local optima with low test loss, we have implemented a series of refinements, which we describe in this subsection.

One common approach when facing non-convex optimization problems is to run the problem many times in parallel with different initial conditions or parameter settings to increase the chance to find a good local optimum. To do this efficiently, we promote $w$ and $\gamma$ to $p \times K$ matrices and promote the $\lambda$'s to $K$-dimensional vectors, each entry of which is multiplied with one column of $\gamma$ or $w$. The objective then becomes
\begin{equation}
    \begin{split}
        L_\lambda(w,\gamma) = ||y - & F(w\gamma)||_2^2 
        + \sum_{i,j,k} F_{ij}^2 w_{jk}^2 \gamma_{jk} (1-\gamma_{jk})\\
        &+  \sum_{i,k}\gamma_{ik}(\lambda_{0,k} + \lambda_{1,k}|w_{ik}| + \lambda_{2,k}w_{ik}^2).
    \end{split}
    \label{eq:L0andL1}
\end{equation}
The gradients of this objective, $\partial L_\lambda(w,\gamma) / \partial w_{jk}$, do not depend on $w_{mk'}$ for any $m\in\{1,\cdots,p\}$ as long as $k'\ne k$, because the summands over $k$ do not interact. The same holds for $\gamma_{jk}$. Therefore, applying gradient descent to this objective effectively optimizes each column independently. In this way, we obtain a simple and effective approach--essentially using matrix multiplication, which is fast on GPUs--to solve the problem $K$ times in parallel for different initial conditions and parameter settings.

Below we also discuss the important questions when to trigger convergence and how to select the coefficients most likely to generalize well.
Initially, we simply used a validation set to determine a validation loss at every $n$-th iteration and triggered convergence once the validation loss increased. Thereafter we selected the model with the lowest validation loss. This already yielded good results. However, in Appendix \ref{app:refinedModelSelection} we explain another selection method that we found to be even more effective for EGP, and that we used throughout the experiments in Section \ref{sec:ComparisonWithMonteCarloMethods} and \ref{sec:compressiveSensingExperiments}.

\subsubsection{Convergence}
\label{sec:convergence}

During training, the estimate of $\theta$ is computed every $n$-th update step. Convergence is then triggered if a minimum number of epochs has elapsed and if two additional criteria are met:
\begin{enumerate}
    \item the deviation of the loss $\mathcal{L}(\theta)$ at epoch $N$ from the previous loss at epoch $N-1$ has been observed to be not greater than a minimum deviation paramater $\Delta$,
    \item the active set of the best-performing column of $\gamma$ has converged. Convergence is determined by checking if $\max_i|\gamma_{iK}-\gamma_{iK}^{\text{(prev)}}|<\epsilon$, where $K=\text{argmin}_k \mathcal{L}_k$, $\epsilon\ge 0$ is a hyperparameter and $\gamma_{iK}^{\text{(prev)}}$ is the best-performing column computed in the previous iteration. 
\end{enumerate}

We also implemented a finetuning procedure, which is optionally applied to the result of the regularized optimization process. Finetuning \citep{lecun1989optimal} (also sometimes called retraining) uses gradient descent to minimize the least squares loss without regularization until the validation loss increases, amounting to a form of relaxed $\ell_0$ regularization.

Finally, we remark that we experimented with many other approaches that we eventually discarded. For example, we modified the implementation of the Adam optimizer \citep{kingma2014adam} to allow for custom learning rates for each column of our matrices, and we experimented with alternative convergence criteria that evaluated stability of the loss by measuring the derivative of a smooth curve fitted to the loss curve using an Epanechnikov kernel. Ultimately, however, these and other amendments proved unnecessary.

\section{Nonlinear \texorpdfstring{$\ell_0$}{L0} regularization}
\label{sec:nonlinearL0Regularization}

In this section, we briefly summarize the main methods that we used in \citep{barth2025efficientcompression} to efficiently find good local optima of (approximations of) objective \eqref{eq:GaussianLoss}.

We investigated 3 distinct approaches, summarized below:
\begin{enumerate}
   \item Probabilistic minimax pruning (PMMP): This can be thought of as a nonlinear generalization of EGP. By introducing additional variables and rewriting the nonlinear objective as a constrained optimization problem, we obtained a formulation which is at most quadratic in $\ell_0$ regularized variables:
   \begin{equation*}
       \begin{split}
           \min_{\theta,w,z}\left\{ \lambda \ell_0(z) + \mathcal{L}(f_\theta, x) \big|\theta=wz \right\} 
           = \min_{\theta,w,z}\max_u\left\{ \lambda \ell_0(z) + \mathcal{L}(f_\theta, x) + u \cdot (\theta-wz)^2 \right\}.
       \end{split}
       \label{eq:constrained}
   \end{equation*}
   The right-hand side can then be rewritten in a probabilistic form, employing (a simplified version of) Theorem \ref{lem:quadratic}.
   \item Relaxed $\ell_1$ regularization (R-L1): Similar to the linear case, one can replace the $\ell_0$ norm by the $\ell_1$ norm. The gradients pull less relevant parameters to $0$ and a threshold procedure can be used to set small parameters to $0$ at the end of the training. In order to compensate for the shrinkage effect of the $\ell_1$ norm, a subsequent finetuning phase thereafter retrains the remaining non-zero parameters of the pruned network. This is similar to Relaxed Lasso 
   \citep{meinshausen2007relaxed,hastie2017extended} and seems to be a popular approach in the nonlinear setting, cf.~\citep{sahoo2018learning}.
   \item Differentiable relaxation of $\ell_0$ regularization (DRR): 
   This method was proposed in \citep{gu2009l0}, \citep{mummadi2019group} and \citep{oliveira2024compression}. 
   It employs the following smooth approximation of the $\ell_0$ norm:
   \begin{equation}
      \begin{split}
         \ell_0(\theta) \approx \sum_i\left(1-\exp\left(-\beta|\theta_i|\right)\right)
      \end{split}
   \end{equation}
   Apart from this, it works like R-L1: After training, parameters below a certain threshold are deleted and finetuning refines the network.
\end{enumerate}
We remark that even though we did not introduce R-L1 and DRR, we refined them with several techniques (like binary search pruning and random gradient pruning) described in more detail in \citep{barth2025efficientcompression}.

Our experiments revealed that the refined versions of  R-L1 and DRR are most competitive among a wide variety of methods across datasets from quite different domains (image classification, regression and next-token prediction tasks). Though PMMP enjoys the advantage that it is the only reformulation whose global minima are equivalent to those of the original objective, it can sometimes face optimization challenges related to the fact that optimal points are saddle points instead of minima. Nevertheless, we believe that the method has a lot of potential that future research might be able to unleash.

In this article, we employ our refined version of DRR to conduct nonlinear compressive sensing experiments in Subsection \ref{sec:TeacherStudentExperiments}.

\section{Active Set Recovery for Multi Layer Perceptrons}
\label{sec:FeffermanMarkelTheory}

Multi-layer perceptrons (MLPs) serve as fundamental building blocks in modern machine learning architectures. Understanding their parameter structure and the relationship between network weights and learned functions remains crucial for developing interpretable and efficient neural networks.
A central challenge in neural network analysis is parameter recovery: given the input-output behavior of an MLP, can we identify the underlying parameters that generated this behavior? 

\begin{figure}[ht]
   \centering
   \begin{subfigure}[t]{0.33\textwidth}
       \includegraphics[width=\linewidth]{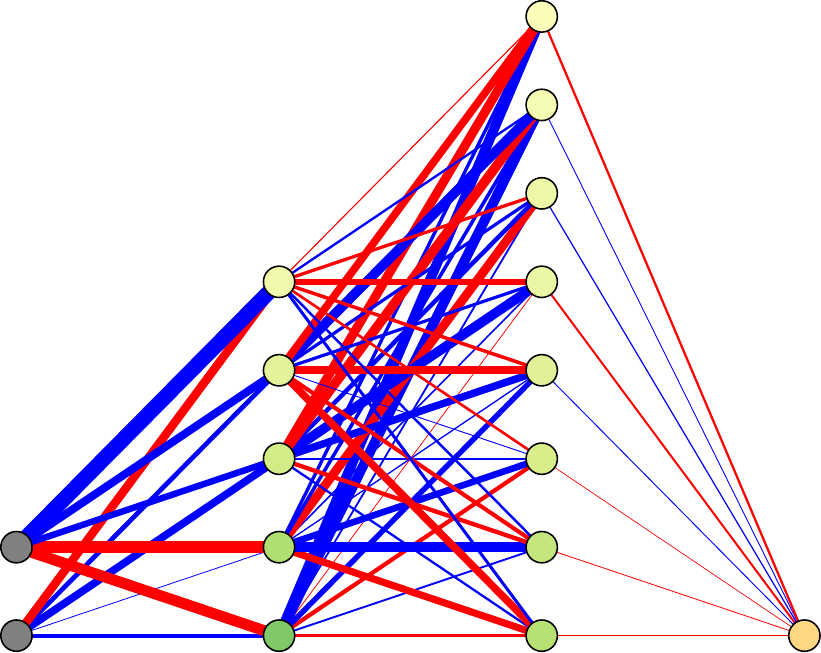}$~$
       \caption*{(a)}
   \end{subfigure}
   \hfill
   \begin{subfigure}[t]{0.33\textwidth}
       \includegraphics[width=\linewidth]{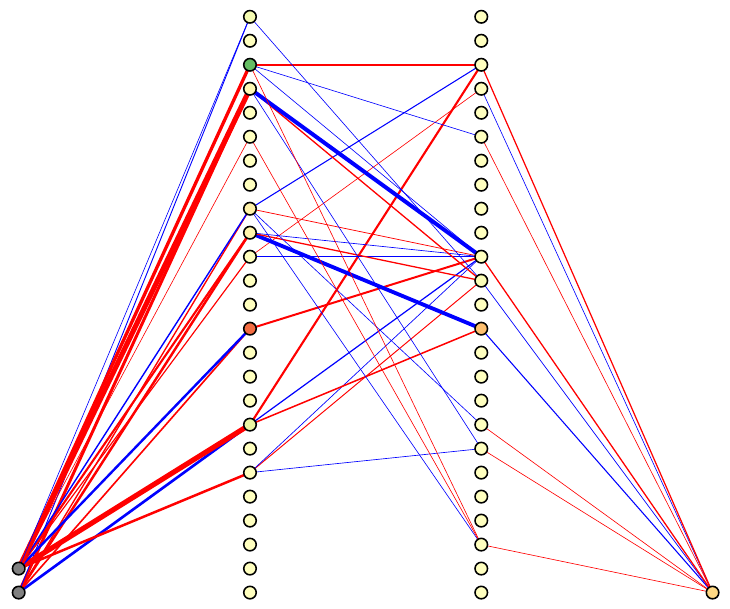}$~$
       \caption*{(b)}
   \end{subfigure}
   \hfill
   \begin{subfigure}[t]{0.29\textwidth}
       \includegraphics[width=\linewidth]{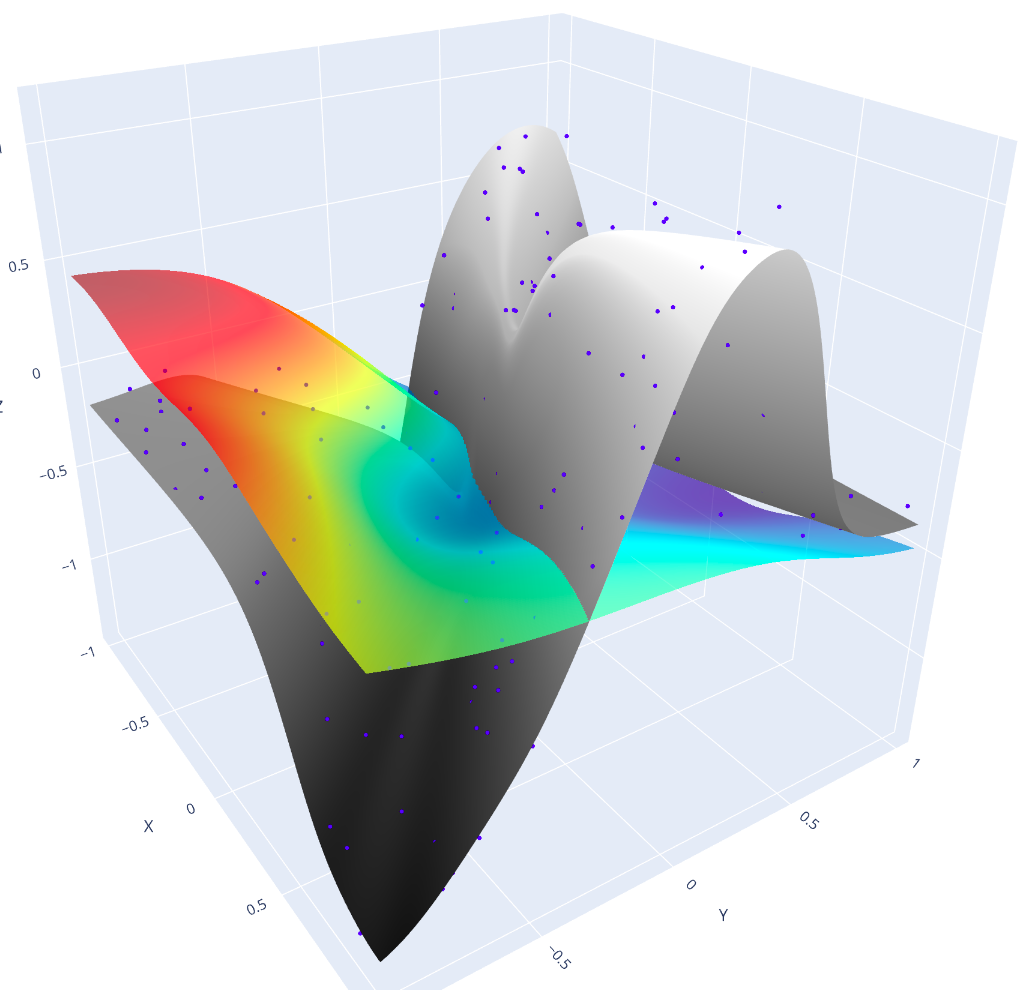}$~$
       \caption*{(c)}
   \end{subfigure}
   \caption{(a) Typical teacher network. (b) Pruned student network. (c) Teacher function (grey), student function (rainbow color, before training) and teacher-generated training dataset (blue dots).}
   \label{fig:TeacherStudentSetup}
\end{figure}

We consider a teacher-student setting where a teacher MLP with parameters $\theta_T$ generates data samples $X = \{(x_i, y_i)|i=1\dots n\}$ such that $y_i = f_{\theta_T}(x_i)$. We aim to recover $\theta_T$ by solving the optimization problem
\begin{equation}
   \begin{split}
      \text{minimize}_{\theta \in M}\left\{~\lambda \ell_0(\theta)+ \mathcal{L}(\theta, X)~\right\},
   \end{split}
   \label{eq:optimizationObjectiveTeacherStudent}
\end{equation}
where $\mathcal{L}$ is a suitable loss function like the MSE loss or the Gaussian loss described in \eqref{eq:GaussianLoss} and $M$ is a class of MLPs, the ``student networks''. We illustrate this setting in Figure \ref{fig:TeacherStudentSetup}.
However, unlike in linear models, the solution (i.e.~the best student) is not uniquely identifiable. 
In the best case, the solution is unique up to certain known symmetry operations.

To exemplify this problem, consider MLPs where the output is computed through layer-wise manipulation of neuron activations with the two equations:
\begin{align}
   z^l_k &= \sum_j w^l_{jk} y^{l-1}_j + b_k^l \\
   y^l_k &= \phi\left(z_k^l\right)
\end{align}
Here, $l$ denotes the layer index and $k$ the neuron index. We call $z^l_k$ the pre-activation of neuron $k$ in layer $l$, and the corresponding activation $y^l_k$ is calculated by the element-wise non-linear function $\phi(\cdot)$.
$w^l_{jk}$ denotes the weight connecting the $j$-th neuron in layer $l-1$ to the $k$-th neuron in layer $l$, while $b^l_k$ is the bias of the $k$-th neuron in layer $l$.
Given the (vector-valued) input $z^0$, we compute the (vector-valued) output $z^L$.
With this definition, we observe that any permutation $\sigma$ of the neuron indices in a given layer, excluding the input and output layers, does not change the computed function:
\begin{equation}
   \label{eq:permutations}
   \sum_j w^l_{jk} y^{l-1}_j + b_k^l = \sum_j w^l_{\sigma(j) k} y^{l-1}_{\sigma(j)} + b_{k}^l 
\end{equation}
Furthermore, depending on the activation function, additional symmetry operations on the network parameters are always permitted.
Consider the case where $\phi$ is odd; then mutual sign flips of all parameters connected to a neuron do not change the computed function:
\begin{equation}
   \label{eq:sign_flip}
   w^{l+1}_{jk} \phi\left( \sum_j w^{l}_{jk} y_j^{l-1} + b^l_k \right) = - w^{l+1}_{jk} \phi\left( \sum_j - w^{l}_{jk} y_j^{l-1} - b^l_k \right)
\end{equation}

While compressive sensing provides theoretical guarantees for active set recovery in linear problems through $\ell_0$  and $\ell_1$ regularization, extending these guarantees to nonlinear neural networks remains challenging. 
The fundamental issue is the beforementioned many-to-one mapping from parameters to functions in neural networks, which creates ambiguity in parameter identification.

We further find that existing nonlinear extensions of compressive sensing, such as the nonlinear extension of iterative hard thresholding, are insufficient for our purposes. As mentioned in the introduction, we find condition \eqref{eq:RIPcondition} too restrictive to apply to MLPs:
When we relate $\Phi_\theta$ to a multi-valued real function $f_\theta(x)$ computed by an MLP, where $\theta$ are the network parameters and $x$ is the input, both interpreted as vectors, the Restricted Isometry Property (RIP) for $\Phi$ translates to:
\begin{equation}
 \label{eq:RIP-for-f}
 \alpha \|\theta_1-\theta_2\|_2^2 \leq \left\| \left(\frac{\partial f_\theta(x)}{\partial \theta}\bigg|_{\theta=\theta^\star}\right) (\theta_1-\theta_2)\right\|_2^2 \leq \beta \|\theta_1-\theta_2\|_2^2
\end{equation}

The convergence theorem in \citep{blumensath2013nonlinear} requires $\beta \leq 1/ \mu < 3/2 \alpha$, where $\mu$ is the learning rate.
Suppose that $\mathcal{A}$ is large enough so that $(\theta_1 - \theta_2)$ can be any vector $\theta$ (this naturally occurs when considering students constrained only to a certain architecture), and suppose that $\alpha$ and $\beta$ are such that the inequality in eq.~\eqref{eq:RIP-for-f} is maximally weak. Then the RIP condition is satisfied if the derivative of $f_\theta$ satisfies the following condition:
\begin{equation}
 \|\theta\|_2 \leq \sqrt{\frac{1}{\mu}} \left\| \left(\frac{\partial f_\theta(x)}{\partial \theta}\bigg|_{\theta=\theta^\star}\right) \theta \right\|_2 \leq \sqrt{\frac{3}{2}} \|\theta\|_2
\end{equation}

We believe that the left-hand inequality does not hold in most practical situations.
A necessary condition for the inequality to hold is that the smallest singular value of $\sqrt{1 / \mu} \left(\partial f_\theta(x) / \partial \theta \big|_{\theta=\theta^\star}\right)$ is neither smaller than one nor larger than $\sqrt{3/2}$. 
For our simulations presented in Section \ref{experiments}, we found that the typical and minimal values were several orders of magnitude smaller than one, meaning that the results from nonlinear IHT do not apply to these experiments.

However, we can address some of the challenges regarding the identifiability of functions computed by MLPs and their parameterizations by drawing from Fefferman and Markel's work \citep{fefferman1993recoveringNNfromOutputs}. They developed networks, that we call \textit{Fefferman-Markel networks}, which are MLPs with $\tanh$ activation functions satisfying the following ``generic'' conditions:

\begin{enumerate}
\item All parameters are non-zero. 
\item No two biases in a layer have the same absolute value.
\item Relative weights are not too rational: Let $w^l$ be the weight matrix in layer $l$ with output dimension $D_l$. Then we require
  $ (w^l_{jk})/(w^l_{j'k}) \neq p/q $ for any $ p,q \in \mathbb{N}, ~1 \leq q \leq 100 D_l^2 $.
\end{enumerate}
We denote the space of parameterizations of the Fefferman-Markel networks as $\mathcal{F}$.
We consider the two symmetry relations: (1) neuron permutations in a layer, see eq.~\eqref{eq:permutations}, and (2) sign-flip of all incoming and outgoing parameters to a node, see eq.~\eqref{eq:sign_flip}.

Under the three conditions formulated above, Fefferman and Markel proved that any two parameterizations computing the same function must be related by the two stated symmetry operations. 

A direct consequence of their theorem is that in the limit of infinite data $n \to \infty$, the global minimum of our regularized optimization problem must yield parameters equivalent to $\theta_T$ up to the mentioned symmetries.
The practical significance is that in this controlled setting, we can study to what extent and under what conditions results from compressive sensing theory generalize to nonlinear problems, potentially informing methods that yield interpretable MLPs if we find that optimization recovers the ground-truth parameters rather than an arbitrary functional approximation.
We implement this framework, accounting for the symmetry relations and the listed conditions. 

We can identify two MLPs that differ only by neuron permutations and sign flips.
The two symmetry operations, see eq.~\eqref{eq:permutations} and eq.~\eqref{eq:sign_flip}, define an equivalence relation $\sim_\text{sym}$. 
Equipping $\mathcal{F}$ with $\sim_\text{sym}$ yields a space of MLPs for which functional equivalence and parameter equivalence are one-to-one.
In practice, we implement an algorithm that applies the symmetries in a deterministic way, such that all representatives of an equivalence class under $\sim_\text{sym}$ are mapped to the same canonical representative, the ``normal form''.
For this, we flip the weights of all neurons that have negative biases as in Equation \eqref{eq:sign_flip}, so that all biases are positive. 
Then we permute the neurons in each layer such that they are ordered by bias magnitude. If two neurons have the same bias, they are then ordered according to the first incoming weight magnitude, and so on.  

In the following, we formalize these remarks and provide mathematical proofs.

\begin{theorem}[
\cite{fefferman1993recoveringNNfromOutputs}]
\label{thm:fefferman-markel}
   For any $ \theta, \theta' \in \mathcal{F} $, it holds that if
   $ f_{\theta} = f_{\theta'}$ (the MLPs compute the same function), then $ \theta ~\sim_\text{sym}~ \theta' $.
\end{theorem}

Our goal is to explore whether it is possible to extract the active set of parameters of the teacher network up to $\sim_\text{sym}$ from a sampled training set using ($\ell_0$ regularized) optimization.
Using the introduced setup of Fefferman-Markel networks, we find that this is in principle possible, at least in the limit of a large dataset.

We specify that in the sampling process of the training set $X$, the inputs $x_i$ are drawn from the multi-dimensional uniform distribution $x_i \sim \mathcal{U}(\mathcal{X})$, where $\mathcal{X}$ denotes the domain of the MLPs we study. We assume that the domain is finite (which is always true in practice, where floating point numbers are used).
Furthermore, let $\mathcal{L}$ denote the loss $\mathcal{L}(f, X) = \sum_i l(y_i, f(x_i))$ with the following conditions on the error function $l$:
\begin{enumerate}
\item Definiteness: $l(x, y) = 0 \Leftrightarrow x = y$
\item Positivity: $l(x, y) \geq 0$
\end{enumerate}

Most typically employed loss functions, such as MSE, MAE and entropy loss, satisfy these requirements.
With this, we can state the following proposition.

\begin{proposition}[Student-Teacher Isomorphism from Loss Minimization]
\label{prop:nn_isomorphim_loss_minimization}
   Let $\theta_{\text{opt}}$ be a solution of the optimization problem
   $ \theta_{\text{opt}} = \arg \min_{\theta \in \mathcal{F}} \mathcal{L}(f_\theta, X), $
   where $\mathcal{L}$ is defined as above.
   Then it holds that:
   $ \theta_{\text{opt}} ~\sim_{\text{sym}}~ \theta_T $
\end{proposition}

To prove the proposition, we will first introduce a helper lemma and then proceed with the main proof.

\begin{lemma}[Function Equality from Zero Error]
\label{lemma:function_equality_from_zero_error}
 Let $f, g$ be functions with finite domain $\mathcal{X}$, and let $l$ be an error function satisfying:
 \begin{enumerate}
 \item $l(x, y) = 0 \Leftrightarrow x = y$
 \item $l(x, y) \geq 0$
 \end{enumerate}

 If $ \sum_{x \in \mathcal{X}} l(f(x), g(x)) = 0, $ then $ f = g \text{ on } \mathcal{X}. $
\end{lemma}

\begin{proof} {\bf - Lemma \ref{lemma:function_equality_from_zero_error}}

Suppose $ \sum_{x \in \mathcal{X}} l(f(x), g(x)) = 0. $ Since $l(x, y) \geq 0$ for all $x, y$, each term in the sum must be zero. 
By the first property of $l$, this implies $f(x) = g(x)$ for all $x \in \mathcal{X}$. Therefore, $f = g$ on $\mathcal{X}$.

\end{proof}

\begin{proof}
{\bf - Proposition \ref{prop:nn_isomorphim_loss_minimization}}

We define $D = D(X = \{(x_i, y_i) | i = 1 \dots n\}) := \{ \xi \in \mathcal{X} | \exists i,\ 1 \leq i \leq n\ \text{s.t.}\ x_i = \xi \}$.
Since the $x_i$ are sampled from $\mathcal{U}(\mathcal{X})$, and the domain is finite, we have:
$ D = \mathcal{X} \text{ as } n \to \infty \text{ with } \mu \text{-prob. } 1 \text{ where } \mu \sim U(\mathcal{X}) $

Note that
$ \min_{\theta \in \mathcal{F}} \mathcal{L}(f_\theta, X) = 0, $
since:
$ \mathcal{L}(f_{\theta_T}, X) = \sum_i l(t(x_i), t(x_i)) = 0 $
From this, it immediately follows that $\mathcal{L}(f_{\theta_{\text{opt}}}, X) = 0$.

Now, consider:
$ \mathcal{L}(f_{\theta_{\text{opt}}}, X) = \sum_i l(f_{\theta_{\text{opt}}}(x_i), t(x_i)) \geq \sum_{\xi \in D} l(f_{\theta_{\text{opt}}}(\xi), t(\xi)) = 0 $
As $n \to \infty$, $D = \mathcal{X}$ with $\mu$-probability 1. 
Applying Lemma \ref{lemma:function_equality_from_zero_error}, we conclude that $f_{\theta_{\text{opt}}} = f_{\theta_{T}}$ on $\mathcal{X}$ with $\mu$-probability 1.
This satisfies the conditions of Theorem \ref{thm:fefferman-markel}, which implies:
$ \theta_{\text{opt}} = \arg \min_{\theta \in \mathcal{F}} \mathcal{L}(f_\theta, X) ~\sim_{\text{sym}}~ \theta_T $

\end{proof}

Proposition \ref{prop:nn_isomorphim_loss_minimization} holds for a rather general class of error functions, but not yet for regularized objectives.
The following proposition extends it to this class.

\begin{proposition}[Neural Network Isomorphism with $\ell_0$ Regularization]
\label{prop:nn_isomorphim_loss_minimization_l0}
   Let $\theta_{\text{opt}}$ be a solution of the optimization problem:
   $ \theta_{\text{opt}} = \arg \min_{\theta \in \mathcal{F}} \mathcal{L}(f_\theta, X) + \ell_0(\theta) $

   Then, in the limit $n \to \infty$, it holds that:
   $ \theta_{\text{opt}} ~\sim_{\text{sym}}~ \theta_T $
\end{proposition}

\begin{proof}

We will show that this optimization problem has the same solution as the one described in Proposition \ref{prop:nn_isomorphim_loss_minimization} in the limit $n \to \infty$.
Let $\theta_{\text{opt}}$ be optimal. We consider two cases:
\begin{enumerate}
\item $\mathcal{L}(f_{\theta_{\text{opt}}}, X) = 0$
\item $\mathcal{L}(f_{\theta_{\text{opt}}}, X) \neq 0$
\end{enumerate}

In the first case, the problem reduces to that of 
Proposition \ref{prop:nn_isomorphim_loss_minimization}.

The second case can be shown to violate the inequality
\begin{equation}
\label{eq_inequality}
0 \leq \min_{\theta \in \mathcal{F}} \mathcal{L}(f_\theta, X) + \ell_0(\theta) \leq \mathcal{L}(f_{\theta_T}, X) + \|\theta_T\|_0 < \infty
\end{equation}
in the $n \to \infty$ limit:
Assume $\mathcal{L}(f_{\theta_{\text{opt}}}, X) \neq 0$. Then $ b = l(f_{\theta_{\text{opt}}}(\xi), f_{\theta_T}(\xi)) > 0 $ for some $\xi$ that is an input occurring in the dataset for some input-target pairing $(\xi, y_i) \in X$.
Let $ c \equiv c(\xi , X) := |\{ i : 1 \leq i \leq n \text{ and } x_i = \xi\}|. $
Since $x_i \sim U(\mathcal{X})$, 
$ c \to \infty \text{ as } n \to \infty \text{ with } \mu\text{-prob. } 1, $ 
where $\mu \sim U(\mathcal{X})$.

Now a simple lower bound estimate yields:
\begin{align*}
\mathcal{L}(f_{\theta_{\text{opt}}}, X) + \|\theta_{\text{opt}}\|_0
 &\geq \mathcal{L}(f_{\theta_{\text{opt}}}, X) 
 \geq c \cdot b 
 \to \infty
\end{align*}
This implies that eq.~\eqref{eq_inequality} is violated.
Therefore, in the limit $n \to \infty$, case 1 always holds.

\end{proof}
\section{Experiments}
\label{experiments}

In this section, we present numerical experiments to compare our probabilistic method against a series of baselines. Furthermore, we show results that elucidate the theory about teacher student networks.
All of our implementations were done in the high-performance programming language julia, cf.~\citep{bezanson2017juliaProgrammingLanguage}.  Our hardware is detailed in Appendix \ref{app:computeresources}.
Our code is publicly available at \href{https://github.com/L0-and-behold/probabilistic-nonlinear-cs}{github.com/L0-and-behold/probabilistic-nonlinear-cs}.

\subsection{Comparison with Monte Carlo methods}
\label{sec:ComparisonWithMonteCarloMethods}

In this subsection, we compare Exact Gradient Pruning (EGP), as described in Section \ref{sec:probReformulation}, with Monte Carlo based probabilistic methods as described at the beginning of the same section. More concretely, we compare our implementation with the one provided by \citep{kunes2023gradient} because their code already includes various different Monte Carlo estimators, namely Reinforce-loo by \citep{kool2019buy}, DisARM by \citep{yin2020probabilistic,dong2020disarm}, as well as BitFlip-1 and UGC by \citep{kunes2023gradient}.

We consider two settings for the comparison, denoted by
\begin{equation}
    \begin{split}
        M_1 = (n=&60,~ p=200,~ s=3,~ \kappa=0.5,~ \lambda=2),\\
        M_2 = (n=&200,~ p=2000,~ s=30,~ \kappa=0.75,~ \lambda=15).
    \end{split}
\end{equation}

These parameters are then used in the following way to generate the necessary quantities for the optimization problem \eqref{eq:RegLinReg}:
\begin{equation}
    \begin{split}
        F \in \mathbb{R}^{n \times p},~ F_{ij} &\stackrel{\text{iid}}{\sim} \mathcal{N}(0, 1),\qquad
        \beta \in \mathbb{R}^{p},~ \beta_{i \le s}  \stackrel{\text{iid}}{\sim} \mathcal{N}(0, 1),~  \beta_{i > s} = 0,\\
        \text{noise} \in \mathbb{R}^{n}, ~\text{noise}_{i}& \stackrel{\text{iid}}{\sim} \mathcal{N}(0, 1),\qquad
        y \in \mathbb{R}^{n}, ~y = F \beta  + \kappa ~\text{noise}
    \end{split}
\end{equation}

\begin{figure}[ht]
    \centering
    \begin{subfigure}[t]{.49\textwidth}
        \includegraphics[width=\textwidth]{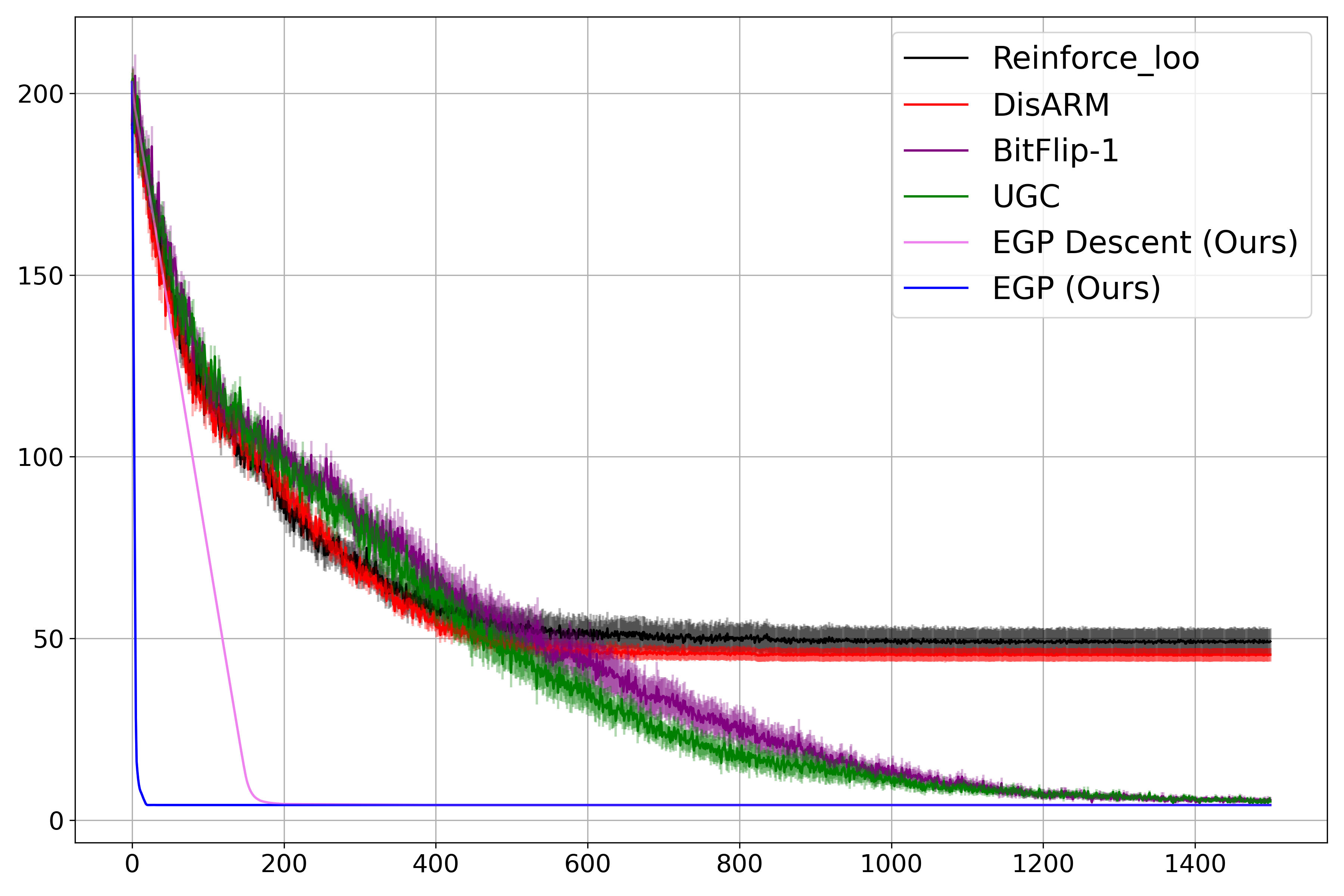}
        \caption{Setting $M_1$}
        \label{fig:MonteCarloMethodComparison1}
    \end{subfigure}
    \hfill
    \begin{subfigure}[t]{.49\textwidth}
        \includegraphics[width=\textwidth]{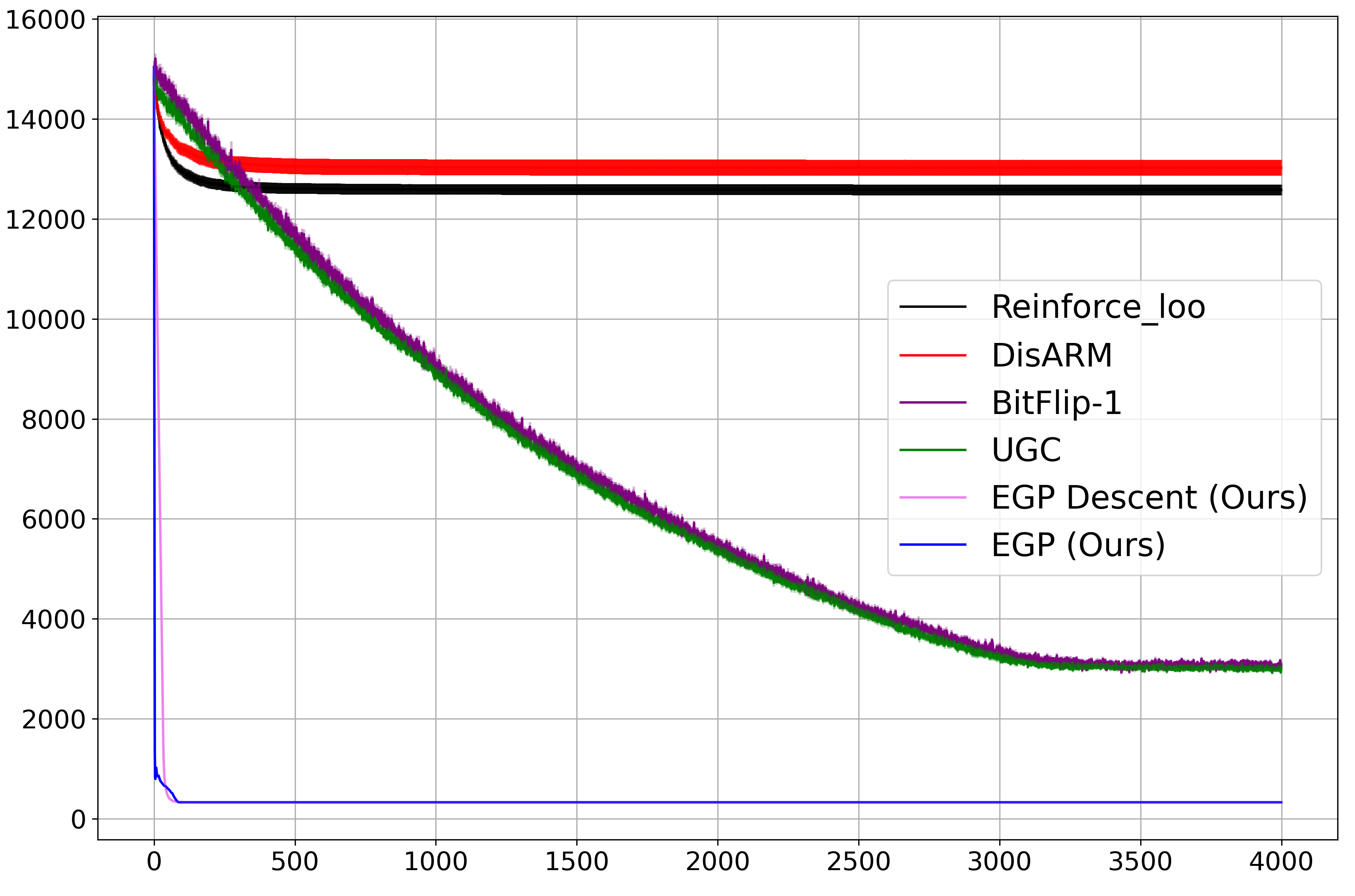}
        \caption{Setting $M_2$}
        \label{fig:MonteCarloMethodComparison2}
    \end{subfigure}
    \caption{Convergence speed comparison of EGP with Monte Carlo methods. Train loss as a function of epochs.}
    \label{fig:MonteCarloMethodComparison}
\end{figure}

We adapted the code of \citep{kunes2023gradient} in which they compared the variance of the gradients and the convergence speed of the different estimators.
In Figure \ref{fig:MonteCarloMethodComparison}, we plot the mean value and standard deviation of the loss as a function of epochs, obtained by running the optimization procedure for the above described settings 10 times until convergence for all 4 Monte Carlo methods, as well as for EGP and an amended version of EGP, where we replaced the default Adam optimizer, cf.~\citep{kingma2014adam}, by plain gradient descent, for a more conservative estimate. We held the dataset fixed to show the variance in the loss introduced by the Monte Carlo estimation procedure. 
Since our gradients are exact, they do not exhibit any variance at all, and hence perform perfectly in this regard, resolving one major concern in the literature about these methods.

When it comes to convergence speed, we see in Figure 
\ref{fig:MonteCarloMethodComparison} and Table \ref{table:MonteCarloComparison}
that the various Monte Carlo estimators either need far longer to converge, or do not converge to a good solution at all. Furthermore, the discrepancy increases for the more challenging setting $M_2$, where the number of optimized dimensions $p$ is much higher than in setting $M_1$. Here, the difference is so big that the loss graph of our methods almost look like the plot axis.

\begin{table}[ht]
    \centering
    \caption{Comparison of EGP and EGP Descent (EGP D.) with Monte Carlo based probabilistic methods. RE = Reconstruction Error, Loss = final $\ell_0$ loss, ASRE = Active set reconstruction error, EUC = Epochs until convergence, TUC = Time until convergence in seconds. All quantities were averaged over 10 runs. Lower values are always better.}
    \label{table:MonteCarloComparison}
    \begin{tabular}{l|ccccc|ccccc}
        & RE & Loss & ASRE & EUC & TUC & RE & Loss & ASRE & EUC & TUC \\
        \midrule
        Reinforce & 6.09 & 49.17 & 24.7 & 600 & 2.23 
        & 87.7 & 12579 & 839.8 & 500 & 17.1\\
        DisARM & 2.07 & 45.53 & 21.3  & 600 & 2.12 
        & 89.4 & 13023 & 873.4 & 500 & 17.5 \\
        Bitflip-1 & 0.08 & 5.56 & 1.62 & 1500 & 4.71 
        & 1130 & 3060 & 229.8 & 3500 & 98.1 \\
        UGC & 0.09 & 5.16 & 1.53 & 1500 & 7.46 
        & 560 & 3006 & 224.6 & 3500 & 164.6\\
        \midrule
        EGP D. & \textbf{0.07} & \textbf{2.17} & \textbf{1.00} & 250 & 0.017 
        & \textbf{3.1} & \textbf{22.9} & \textbf{8.0} & 150 & 0.202\\
        EGP & \textbf{0.07} & \textbf{2.17} & \textbf{1.00} & \textbf{30} & \textbf{0.002} 
        & \textbf{3.1} & \textbf{22.9} & \textbf{8.0} & \textbf{100} & \textbf{0.198} \\
    \end{tabular}
    \end{table}

When comparing actual runtimes, the difference is even more dramatic. In Table \ref{table:MonteCarloComparison}, we show the time until convergence (TUC). EGP is several orders of magnitude faster than the best performing method for setting $M_1$, while for setting $M_2$ none of the Monte Carlo methods converges to a good solution at all.

Furthermore, despite the much faster convergence, the quality of the results of EGP is simultaneously superior to that of the Monte Carlo simulations. In the first 3 columns of Table \ref{table:MonteCarloComparison}, we report reconstruction error (RE: $\sum_j|\beta-\theta|$), final $\ell_0$ loss (Loss: $||y-F\theta||_2^2/n + \lambda \ell_0(\theta)$), and active set reconstruction error (ASRE: $\sum_j |1_{\beta\ne 0}-1_{\theta\ne 0}|$, where $1_{\beta\ne 0} \in \mathbb{R}^p$ is a vector whose entries are equal to $1$ where the entries of $\beta$ are non-zero and $0$ elsewhere). For all these metrics, EGP performs best. In fact, for setting $M_2$, we were not able to find hyperparameter values for which any of the Monte Carlo methods converged to a good solution (and even if one found such values, this would not affect convergence speed).
Finally, we note that there is almost no difference in these metrics between EGP Descent and usual (Adam-based) EGP.

\bigskip

There are three main reasons why the Monte Carlo implementations are so much slower:
\begin{enumerate}
    \item When using multiple samples to increase accuracy, then each gradient computation requires to average $k$ Monte Carlo estimates of the gradient. This takes about $k$ times longer than a usual gradient step or requires additional resources to parallelize these estimates. However, in the experiments presented above, we employed only a single sample per time step, for a conservative estimate (other choices of $k$ lead to slower convergence).
    \item The inexactness and variance of the gradients makes convergence much less efficient. This is especially problematic in the case of $\ell_0$ regularization because the optimal points are at the boundary of the domain (namely where all components of $\gamma$ in \eqref{eq:linearcase} become exactly $0$ or $1$), where the variance of most estimators is especially big, as shown by \citep{kunes2023gradient}. This means that estimators that work well in domains like variational autoencoders, as shown for example for the DisARM estimator in \citep{dong2020disarm}, are very ineffective for $\ell_0$ regularization. Indeed, in Figure \ref{fig:MonteCarloMethodComparison} and Table \ref{table:MonteCarloComparison}, we see that those estimators do not converge at all to a solution that is close to the original $\beta$. This is precisely why \citep{kunes2023gradient} came up with the BitFlip-1 estimator but as the name already suggests, this estimator only allows to flip one bit at a time!\footnote{One can implement flipping of more than 1 bit per Monte Carlo sample but it is likely not an unbiased estimator of \eqref{eq:derivPiAlternative} anymore and in any case just yields information about the improvement with respect to one other particular out of $2^p$ flip combinations, where $p$ is the number of parameters. Intuitively, flipping more than 1 bit also only yields information about improvement in one particular direction in parameter space. 
    In practice, we did not observe any performance improvements when flipping more than one bit per Monte Carlo sample.} Concretely, this means that gradients of this estimator only affect a single component in the vector $\gamma$ at each step. The gradients are hence extremely sparse and massively less effective than exact gradients. Only taking $p$ Bitflip-1 samples per step--one for each component of $\gamma$--would yield the true gradient. However, this requires $2p$ forward passes and is thus $\mathcal{O}(p)$ times more expensive than usual backpropagation costs for differentiable functions (which are usually only a small constant multiple (typically 2-5 times) of the forward pass costs). Therefore the Monte Carlo methods perform worse when $p$ is increased, explaining the difference between Figure \ref{fig:MonteCarloMethodComparison1} and \ref{fig:MonteCarloMethodComparison2}. As a result, no matter which estimator one uses, convergence is very slow in comparison to an exact gradient method. 
    \item In the minimization of \eqref{eq:linearcase}, one has to optimize over both $w$ and $\gamma$. To make the Monte Carlo estimate in \eqref{eq:derivPiAlternative} precise, one first has to evaluate a good estimate of $g(z)=\min_w ||y-F(wz)||_2^2 + \lambda\ell_0(z)$, which includes an entire minimization procedure over $w$. \citep{kunes2023gradient} implement the minimization via a linear regression over $w$, which is quite costly when performed at every single gradient step. One could perhaps speed things up by instead aplying simply a single gradient step with respect to $w$ to $||y-F(wz)||_2^2 + \lambda\ell_0(z)$ and hope for convergence but the price would be that the estimate of the gradient with respect to $\gamma$ would become even more imprecise. Thus, even though approximating the linear regression problem could speed up each gradient step, one would then need more iterations overall to converge or might not achieve convergence to a good solution at all.
\end{enumerate}

Before concluding this section, we want to emphasize that this comparison is already quite conservative because we deactivated several default features of EGP for a fairer comparison, as we explain in more detail in Appendix \ref{app:MonteCarlo}. 

As a conclusion, even if someone further improved the efficiency of the Monte Carlo implementations of \citep{kunes2023gradient}, it is very unlikely that one would get anywhere near the efficiency of the EGP approach. Furthermore, the effectiveness and simplicity of EGP make it very scalable. 
In Subsection \ref{sec:runtimeBenchmarks}, we show that the time complexity of EGP increases linearly with $n$ and $p$.
Even very large matrices $F$ can be optimized efficiently in smaller batches and all matrix operations can be carried out on the GPU. In fact our implementation provides those features by employing the Lux library \citep{pal2023lux}, while switching to the CPU for small problem sizes (like those in the above comparison).

\subsection{Systematic compressive sensing experiments}
\label{sec:compressiveSensingExperiments}

In this subsection, we compare EGP very systematically with the most established methods employed in compressive sensing. Those include 
\textit{Lasso} by 
\citep{tibshirani1996regression},
\textit{Relaxed Lasso} by 
\citep{meinshausen2007relaxed,hastie2017extended},
\textit{Forward Stepwise} by 
\citep{efroymson1966stepwise},
and
\textit{Iterative Hard Thresholding (IHT)} by 
\citep{blumensath2008iterative}.
We use the highly efficient implementations of Lasso, Relaxed Lasso and Forward Stepwise from \citep{hastie2017extended}\footnote{Their interface was implemented in R, while EGP was implemented in julia. Hence we call their R code from julia during the tests. Their R code in turn often calls (compiled) C code, which is very fast.} and of IHT\footnote{implemented in julia under the name \textit{Mendel IHT}, cf.~\href{https://github.com/OpenMendel/MendelIHT.jl}{github.com/OpenMendel/MendelIHT.jl}} from \citep{chu2020IHT}.

These methods all exhibit very competetive speed. We do not include comparatively slow methods that attempt exact best subset selection like Mixed Integer Optimization (MIO) as introduced in \citep{bertsimas2016best}. MIO can often take hours for problems for which the methods, that we mentioned above, take only a fraction of a second. In addition to that, it requires proprietary solver software like Gurobi. 
Furthermore, Lasso, Relaxed Lasso and Forward Stepwise
have already been systematically compared to MIO for signal-to-noise ratios (SNRs) between 0.05 and 6 in \citep{hastie2017extended} (however with a limited compute budget for MIO to be able to perform those comparisons at all),
and there, for almost every setting and SNR, MIO either performed very similar to Forward Stepwise (in high SNR regime) or worse than Relaxed Lasso (in low SNR regime). 
Hence, we are quite confident that in most cases in which our method outperforms both Forward Stepwise and Relaxed Lasso, it would also outperform MIO.

For our comparison of compressive sensing methods, we build off of the nice framework developed in \citep{bertsimas2016best} and extended by
\citep{hastie2017extended}. We take over some settings from \citep{hastie2017extended} but test a larger range of SNRs and larger dimensions.

\subsubsection{Signal-to-noise ratios}
\label{sec:SNRs}

In \citep{hastie2017extended}, it is argued that SNRs up to 6 are sufficient in practice. However, some applications operate at medium or high SNRs like magnetic resonance imaging \citep{jaspan2015compressed} or microscopy systems \citep{robinson2024high}, or might even be virtually noise free, like lossless compression or layerwise pruning, cf.~\citep{phan2020pruning, barth2025efficientcompression}, where the input and response are computed deterministically.

Furthermore, in several relevant simulation settings (like ``High-5'' and ``High-10'' in \citep{hastie2017extended}), most methods only start to perform significantly better than the trivial zero-coefficient in terms of relative test error (RTE) for quite high SNRs, which is an indication that longer ranges must be considered to see the asymptotic behavior of the different methods. We thus decided to double the range of SNRs from 10 to 20 and considered a range from 0.05 to 60000, equally spaced on a logscale, i.e. SNRs $:=\text{Logs}(0.05,60000,20)$ (rounded to two digits), where 
\begin{equation}
    \begin{split}
        \text{Logs}(a,b,n) \in \mathbb{R}^n,~\text{Logs}(a,b,n)_i &:= 10^{c_i},~c_i := \log_{10}{a} + \frac{i-1}{n-1}(\log_{10}{b}-\log_{10}{a}).
    \end{split}
    \label{eq:logs}
\end{equation}
These include values very similar to the 10 logspaced SNRs from $0.05$ to $6$ considered in \citep{hastie2017extended}.

\subsubsection{Settings}
\label{sec:settings}

For a given SNR, and a given (auto)correlation level $\rho$, we describe an experimental \textit{setting} for (linear) compressive sensing by a tuple
$(n,~ p,~ s)$.
The tuple denotes the parameters according to which the relevant quantities in the optimization problem \eqref{eq:RegLinReg} are generated as follows:
\begin{enumerate}
    \item True coefficient vector $\beta\in\mathbb{R}^p$ (to which the estimate $\theta$ in \eqref{eq:RegLinReg} should converge),
    \begin{equation}
        \begin{split}
            \beta_{i\le s} = 1,~\beta_{i > s = 0} = 0
        \end{split}
        \label{eq:truecoef}
    \end{equation}
    \item Covariance matrix $\Sigma \in \mathbb{R}^{p\times p}$,
    $\Sigma_{ij} = \rho^{|i-j|}$
    \item Design matrix $F \in \mathbb{R}^{n \times p}$, where each row $F_i$ is sampled from $\mathcal{N}(\mathbf{0}, \boldsymbol{\Sigma})$
    \item  Noise variance $\sigma^2\in\mathbb{R}$,~
    $\sigma^2 = \boldsymbol{\beta}^T \boldsymbol{\Sigma} \boldsymbol{\beta} / \text{SNR}$,
    corresponding to the desired SNR
    \item Response vector $y\in\mathbb{R}^n$,~    $\mathbf{y} \sim \mathcal{N}(\mathbf{X}\boldsymbol{\beta}, \sigma^2 \mathbf{I}_n)$
\end{enumerate}

Here \eqref{eq:truecoef} corresponds to \textit{beta-type 1} 
in \citep{bertsimas2016best} and \citep{hastie2017extended}. 
The rest of the construction was also taken over from them.

The concrete settings that we used for our benchmarks for a given $\text{SNR} \in\text{Logs}(0.05,60000,20)$ and $\rho \in \{0.0,~0.35,~0.7,~0.9\}$ are
\begin{equation}
    \begin{split}
        &S_1:=(n=100,~ p=10,~ s=5),
        \qquad\quad~ S_2:=(n=50,~ p=1000,~ s=5),\\
        &S_3:=(n=100,~ p=1000,~ s=10),\qquad
        S_4:=(n=100,~ p=10000,~ s=10).
    \end{split}
    \label{eq:settings}
\end{equation}

Settings $S_1$, $S_2$ and $S_3$ were taken over from \citep{hastie2017extended}, while $S_4$ is a more challenging setting that we added to investigate very high $p$ performance.

For each of the $4$ settings, we conduct simulations with $5$ methods across $20$ SNRs and $4$ $\rho$ values, which amounts to $1600$ simulations in total.
In every single simulations, we run each method hundreds of times with different basic parameter choices (varying $\lambda$, learning rate or active set size) and select the final coefficient by comparing the results. The details of this model selection procedure are described in Appendix \ref{app:coefficientSelection}. Apart from these basic parameters, all other hyperparameter values were held fixed throughout all settings.

\subsubsection{Performance metrics}

To measure performance, we adopt the idea to use Relative Test Error (RTE) from \citep{hastie2017extended}:
\begin{equation}
    \begin{split}
        \text{RTE}_{F,\beta,y}(\theta) := \frac{1}{\sigma^2}\mathbb{E}\big[||y-F\theta||_2^2\big] = \frac{(\theta-\beta)^T\Sigma(\theta-\beta)+\sigma^2}{\sigma^2}
    \end{split}
    \label{eq:RTE}
\end{equation}
Among the metrics we have seen, this one seems to be particularly sensible. It measures generalization error (which is the most widely used metric in machine learning) very precisely by employing a reformulation in terms of the true coefficient $\beta$. At the same time, it takes the effect of noise, $\sigma^2$, into account. The division by $\sigma^2$ penalizes mistakes of the estimation for high SNR values much more than for lower ones, which is fair given that estimation becomes considerably harder for lower SNRs. Furthermore, RTE values are always between $1$ and  $(\text{SNR} + 1)$. We indicate the latter by a dotted black line in our plots.

Furthermore, we employ the commonly used Active Set Reconstruction Error (ASRE), 
\begin{equation}
    \begin{split}
        \text{ASRE}:= \sum_j |1_{\beta\ne 0}-1_{\theta\ne 0}|,
    \end{split}
    \label{eq:ASRE}
\end{equation}
where $1_{\beta\ne 0} \in \mathbb{R}^p$ is a vector whose entries are equal to $1$ where the entries of $\beta$ are non-zero and $0$ elsewhere. It measures in how far the correct basis elements of the function basis, described in \eqref{eq:compressiveSensingObjective}, are selected.

\subsubsection{Results}
\label{sec:EGPresults}

\begin{figure}[ht]
    \centering
    \begin{subfigure}[t]{0.49\textwidth}
        \includegraphics[width=\linewidth]{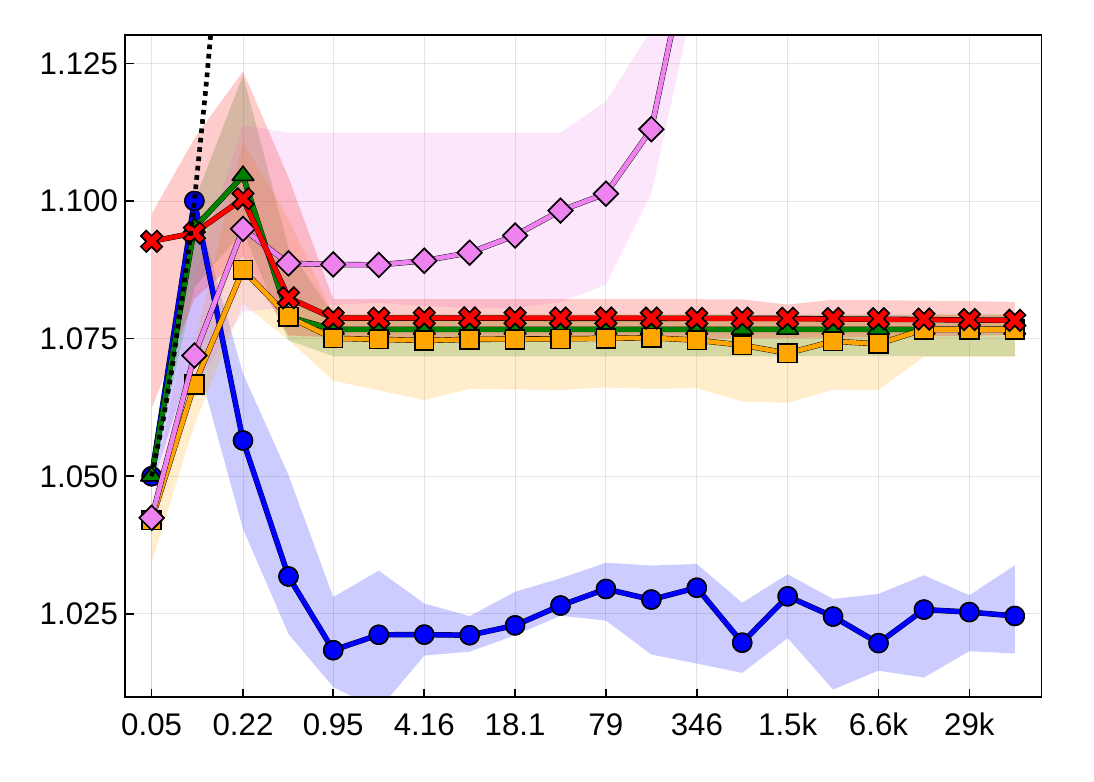}
        \caption{Setting $S_1$}
        \label{fig:RTE:s=1rho=00}
    \end{subfigure}
    \hfill
    \begin{subfigure}[t]{0.49\textwidth}
        \includegraphics[width=\linewidth]{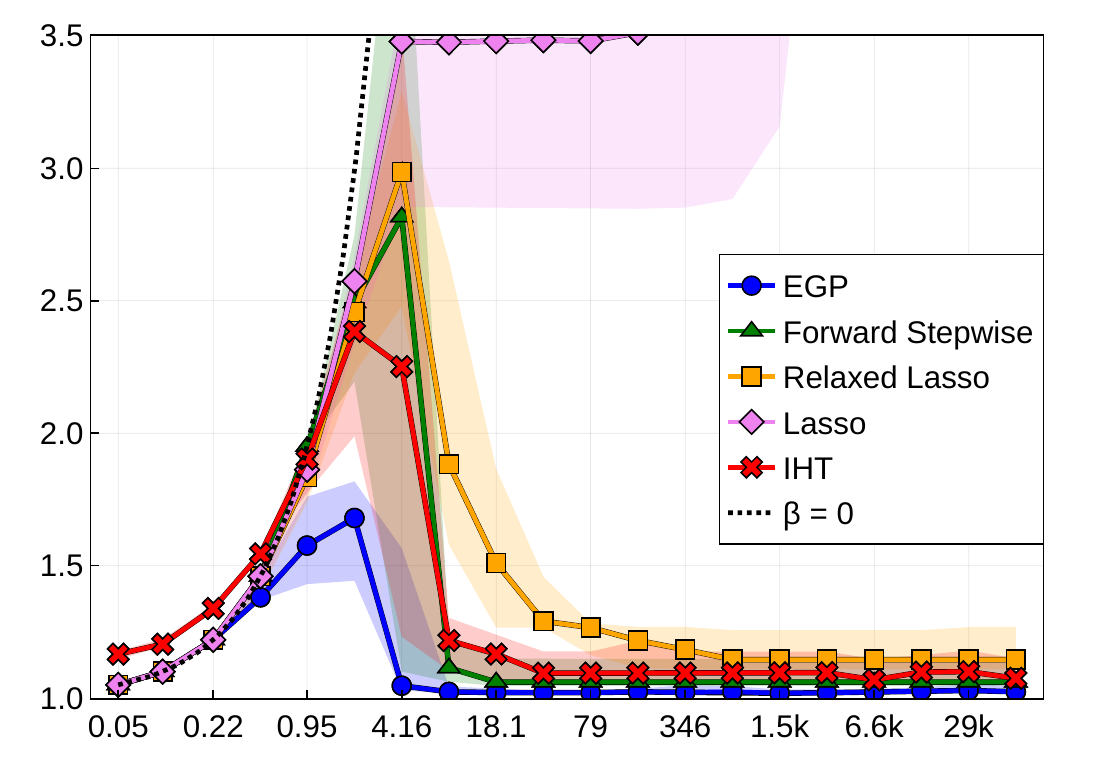}
        \caption{Setting $S_2$}
        \label{fig:RTE:s=2_rho=00}
    \end{subfigure}\\
    \begin{subfigure}[t]{0.49\textwidth}
        \includegraphics[width=\linewidth]{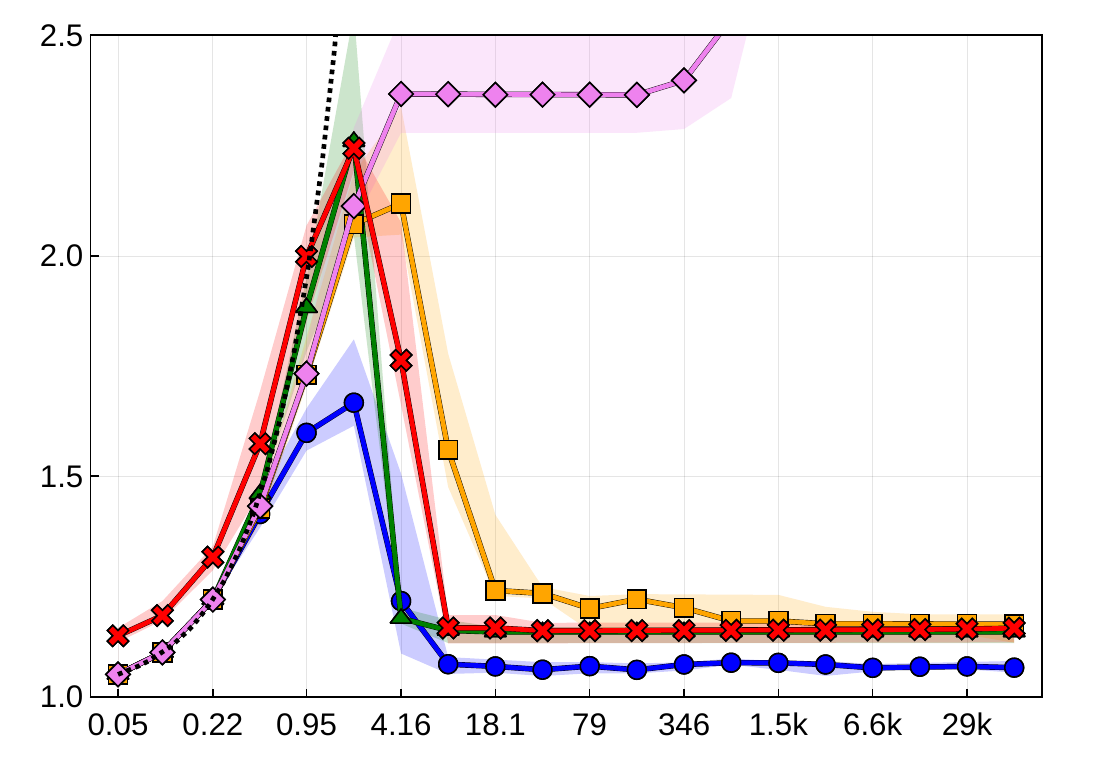}
        \caption{Setting $S_3$}
        \label{fig:RTE:s=3rho=00}
    \end{subfigure}
    \hfill
    \begin{subfigure}[t]{0.49\textwidth}
        \includegraphics[width=\linewidth]{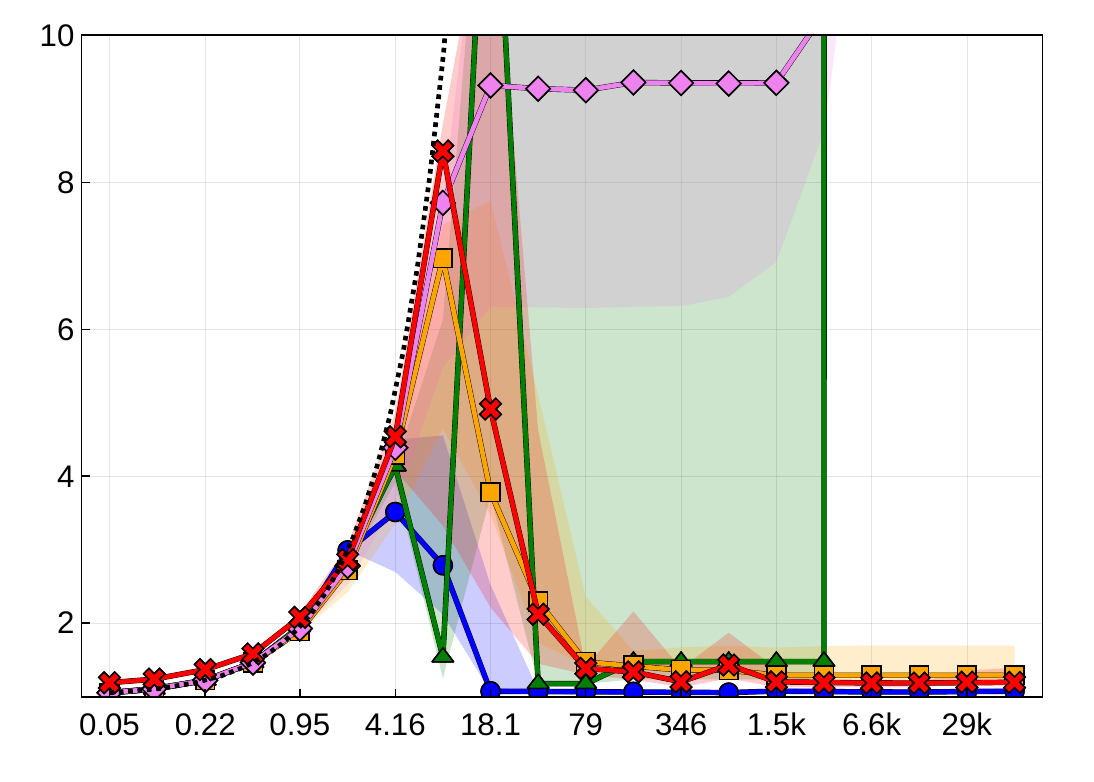}
        \caption{Setting $S_4$}
        \label{fig:RTE:s=4rho=00}
    \end{subfigure}
    \caption{Relative test error (RTE, cf.~\eqref{eq:RTE}) as a function of signal-to-noise ratio (SNR, cf.~Subsection \ref{sec:SNRs}) for the different settings described in \eqref{eq:settings} and (auto)correlation level $\rho=0$. Medians over $10$ runs with quantile range between $0.3$ and $0.7$. Legend holds for all subfigures.}
    \label{fig:RTE:rho0}
\end{figure}

In Figure \ref{fig:RTE:rho0}, we show RTE as a function of SNR for the settings described in \eqref{eq:settings} with correlation level $\rho=0$. In order to stay agnostic to the source of error, we plot the median results together with the quantile range between 0.3 and 0.7 (slightly less wide than interquartile range (IQR) to increase readability), computed from 10 simulations for each method, setting and SNR.

\begin{figure}[ht]
    \centering
    \begin{subfigure}[t]{0.49\textwidth}
        \includegraphics[width=\linewidth]{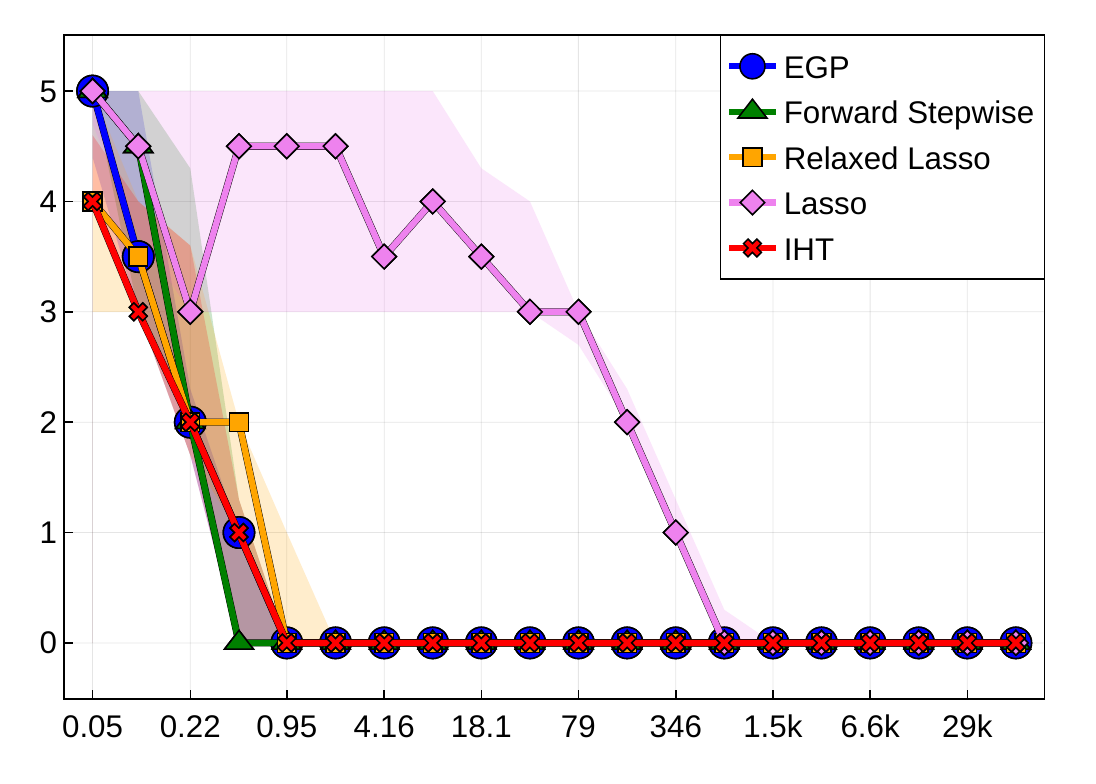}
        \caption{Setting $S_1$}
        \label{fig:ASRE:s=1rho=00}
    \end{subfigure}
    \hfill
    \begin{subfigure}[t]{0.49\textwidth}
        \includegraphics[width=\linewidth]{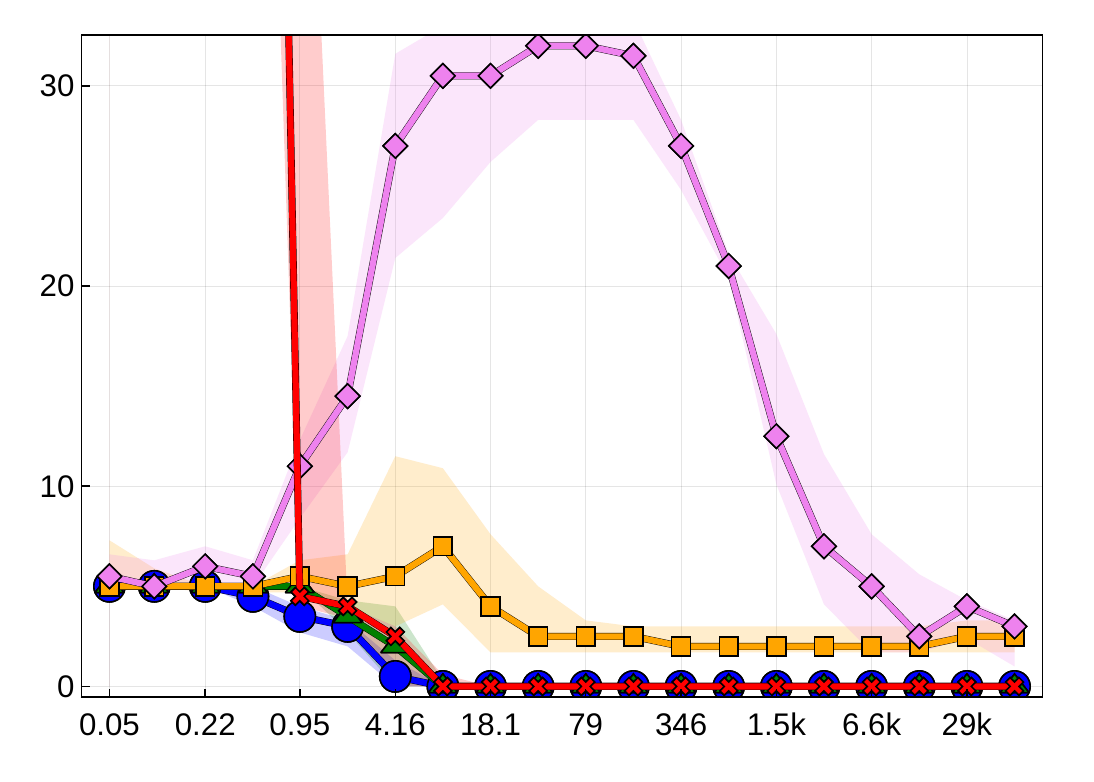}
        \caption{Setting $S_2$}
        \label{fig:ASRE:s=2_rho=00}
    \end{subfigure}\\
    \begin{subfigure}[t]{0.49\textwidth}
        \includegraphics[width=\linewidth]{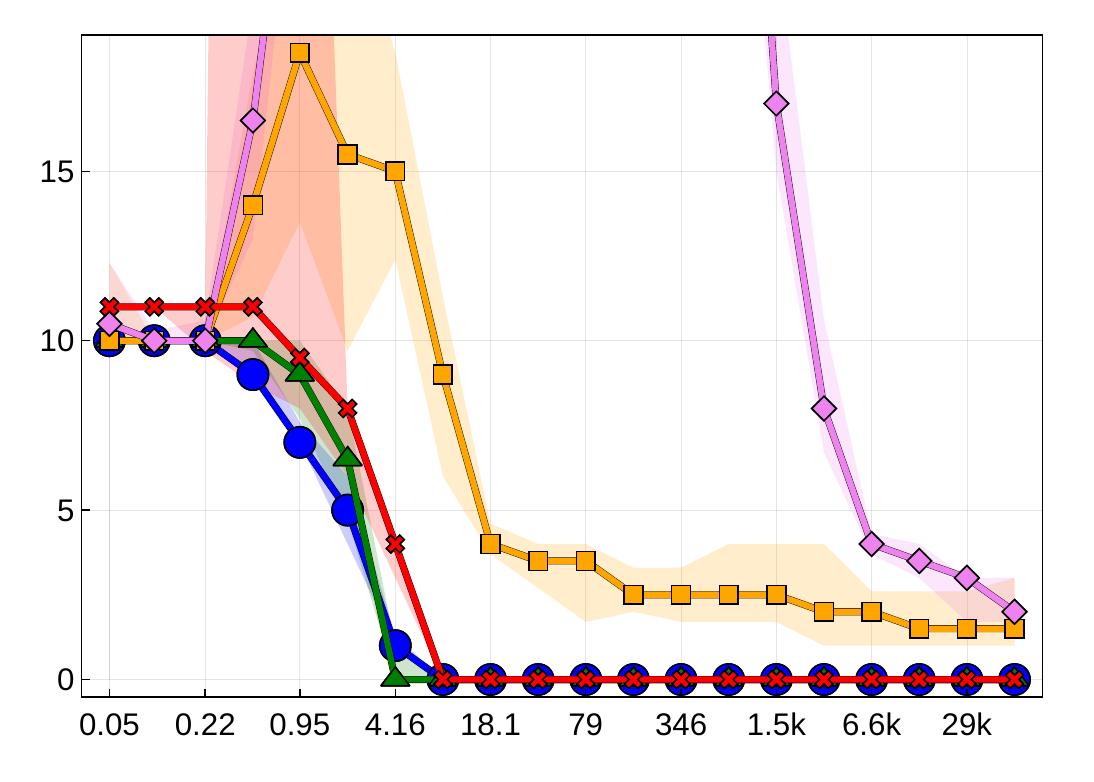}
        \caption{Setting $S_3$}
        \label{fig:ASRE:s=3rho=00}
    \end{subfigure}
    \hfill
    \begin{subfigure}[t]{0.49\textwidth}
        \includegraphics[width=\linewidth]{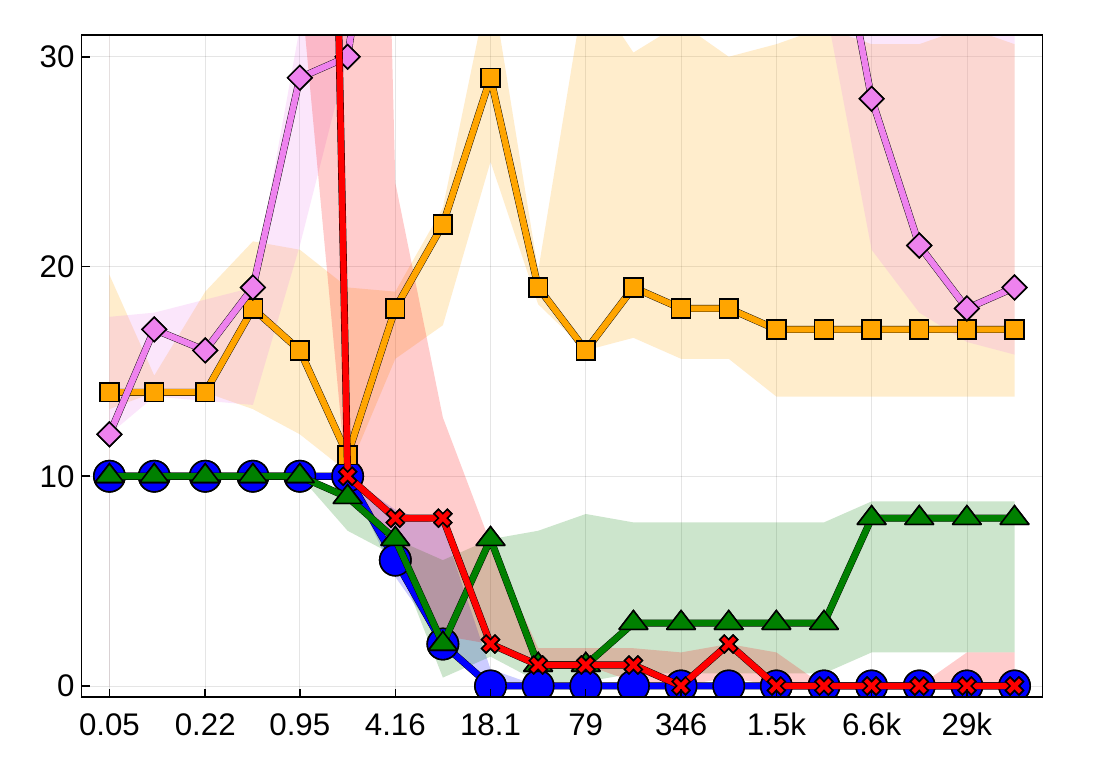}
        \caption{Setting $S_4$}
        \label{fig:ASRE:s=4rho=00}
    \end{subfigure}
    \caption{Active set reconstruction error (ASRE, cf.~\eqref{eq:ASRE}) as a function of signal-to-noise ratio (SNR, cf.~Subsection \ref{sec:SNRs}) for the different settings described in \eqref{eq:settings} and (auto)correlation level $\rho=0$. Medians over $10$ runs with quantile range between $0.3$ and $0.7$. Legend holds for all subfigures.}
    \label{fig:ASRE:rho0}
\end{figure}

\begin{figure}[ht]
    \centering
    \begin{subfigure}[t]{0.49\textwidth}
        \includegraphics[width=\linewidth]{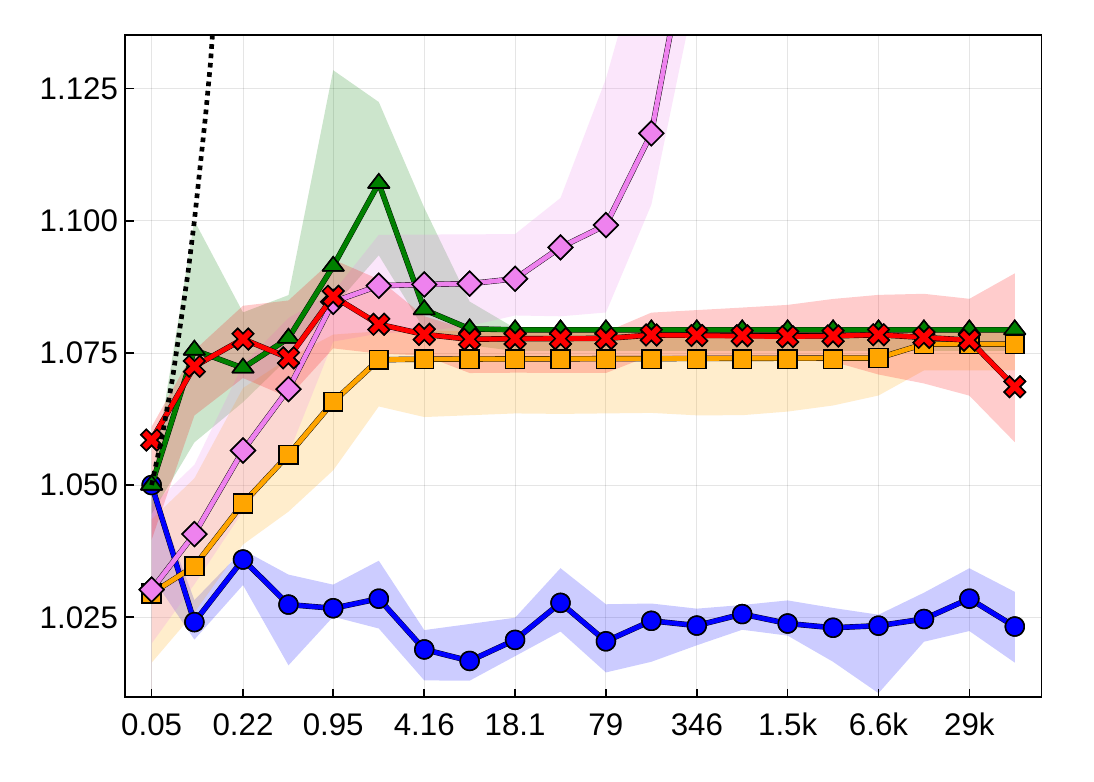}
        \caption{Setting $S_1$}
        \label{fig:RTE:s=1rho=07}
    \end{subfigure}
    \hfill
    \begin{subfigure}[t]{0.49\textwidth}
        \includegraphics[width=\linewidth]{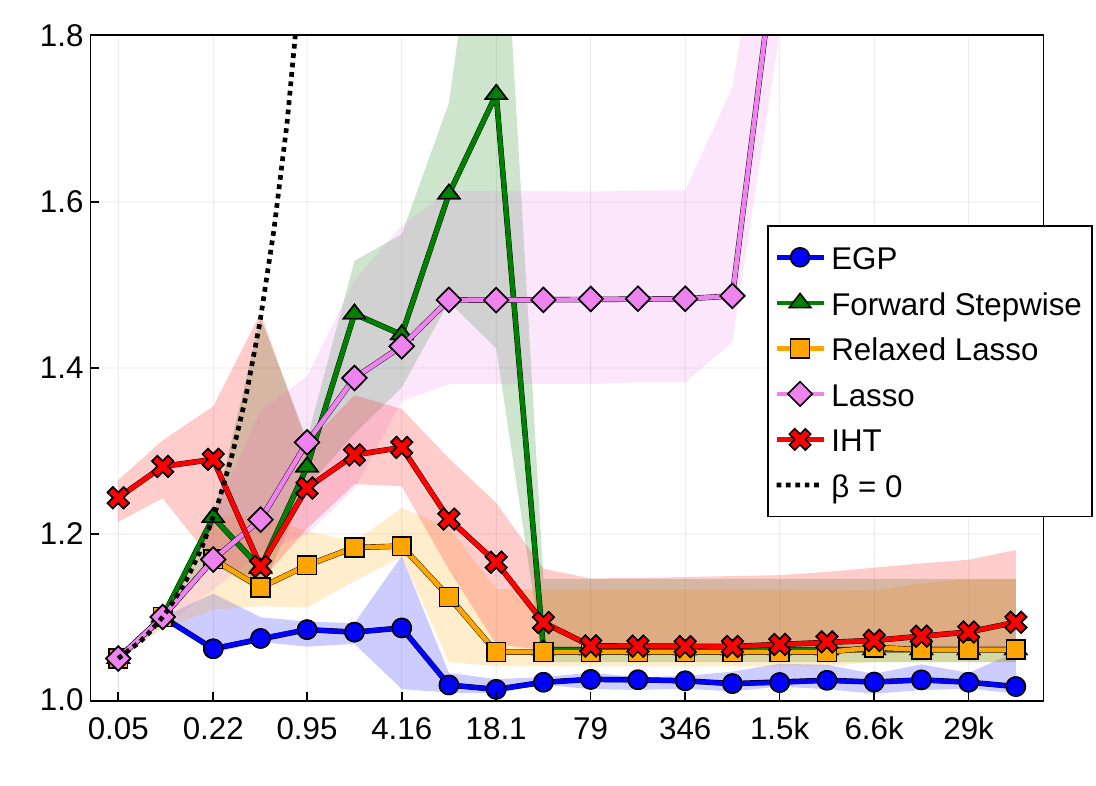}
        \caption{Setting $S_2$}
        \label{fig:RTE:s=2_rho=07}
    \end{subfigure}\\
    \begin{subfigure}[t]{0.49\textwidth}
        \includegraphics[width=\linewidth]{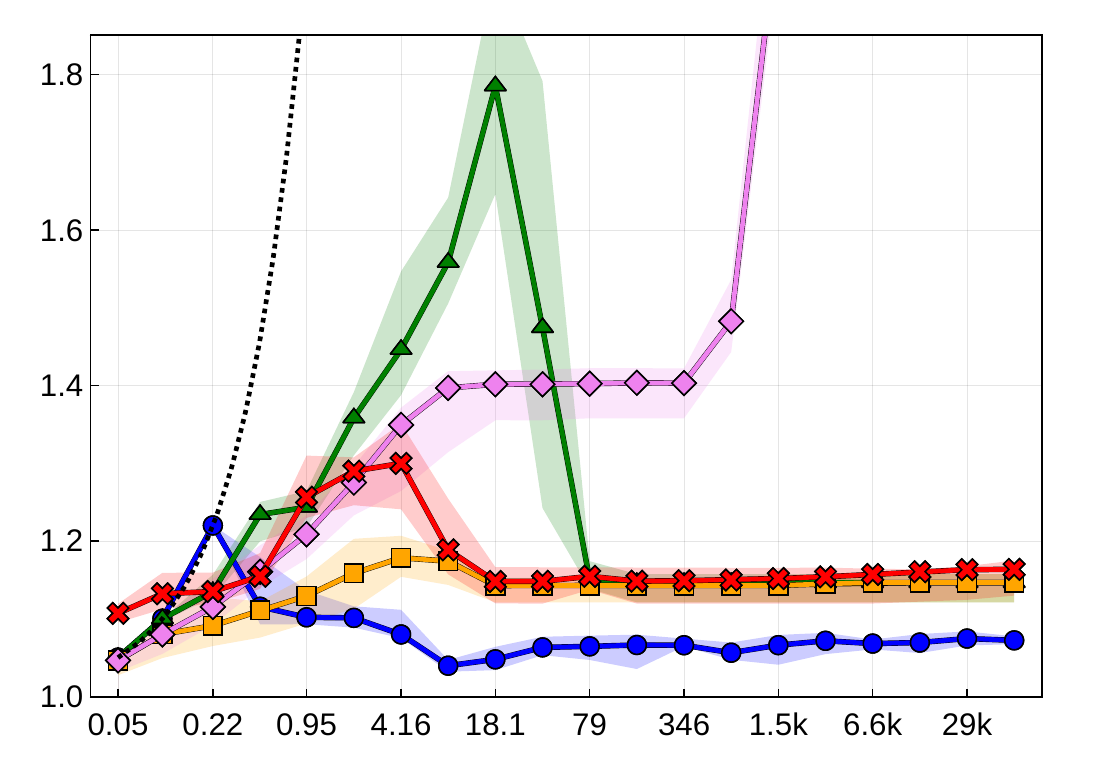}
        \caption{Setting $S_3$}
        \label{fig:RTE:s=3rho=07}
    \end{subfigure}
    \hfill
    \begin{subfigure}[t]{0.49\textwidth}
        \includegraphics[width=\linewidth]{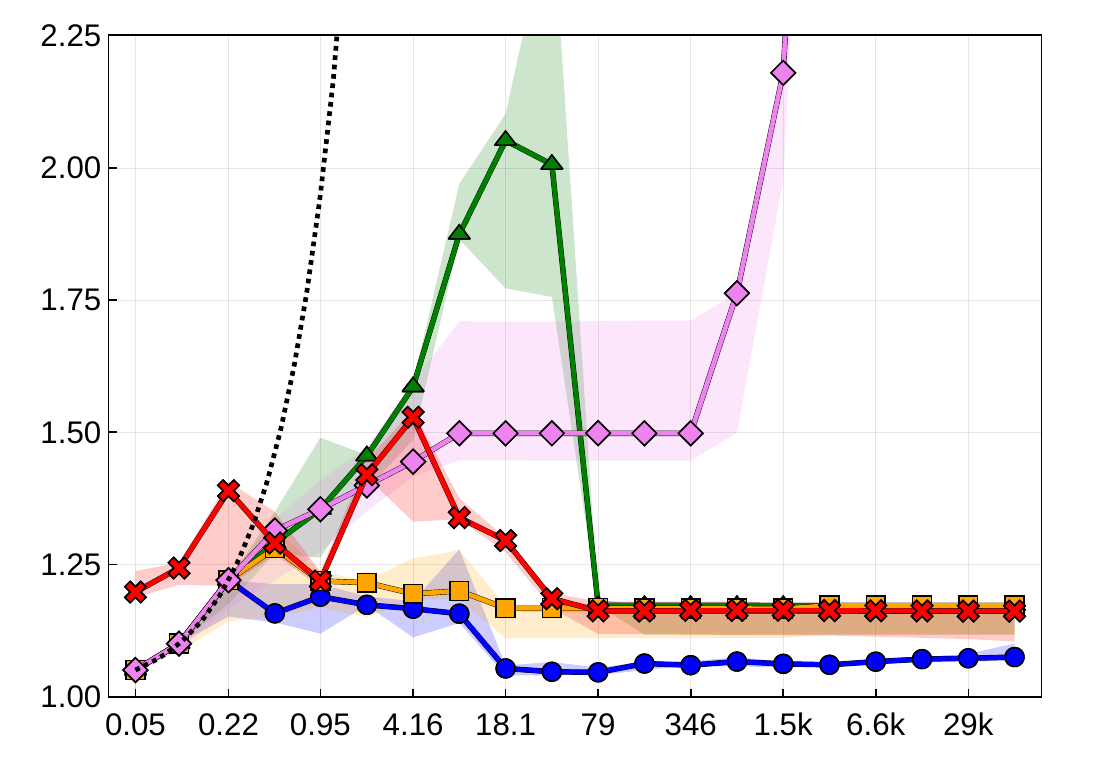}
        \caption{Setting $S_4$}
        \label{fig:RTE:s=4rho=07}
    \end{subfigure}
    \caption{Relative test error (RTE, cf.~\eqref{eq:RTE}) as a function of signal-to-noise ratio (SNR, cf.~Subsection \ref{sec:SNRs}) for the different settings described in \eqref{eq:settings} and (auto)correlation level $\rho=0.7$. Medians over $10$ runs with quantile range between $0.3$ and $0.7$. Legend holds for all subfigures.}
    \label{fig:RTE:rho07}
\end{figure}

Figure \ref{fig:ASRE:rho0} shows the corresponding ASRE.
To check if those results are stable even when varying the correlation among coefficients, we also present RTE results for $\rho=0.7$ in Figure \ref{fig:RTE:rho07} and corresponding ASRE results in Figure \ref{fig:ASRE:rho07}, as well as figures for $\rho=0.35$ and $\rho=0.9$ in Appendix \ref{app:furtherCompressiveSensingExperiments}.

\begin{figure}[ht]
    \centering
    \begin{subfigure}[t]{0.49\textwidth}
        \includegraphics[width=\linewidth]{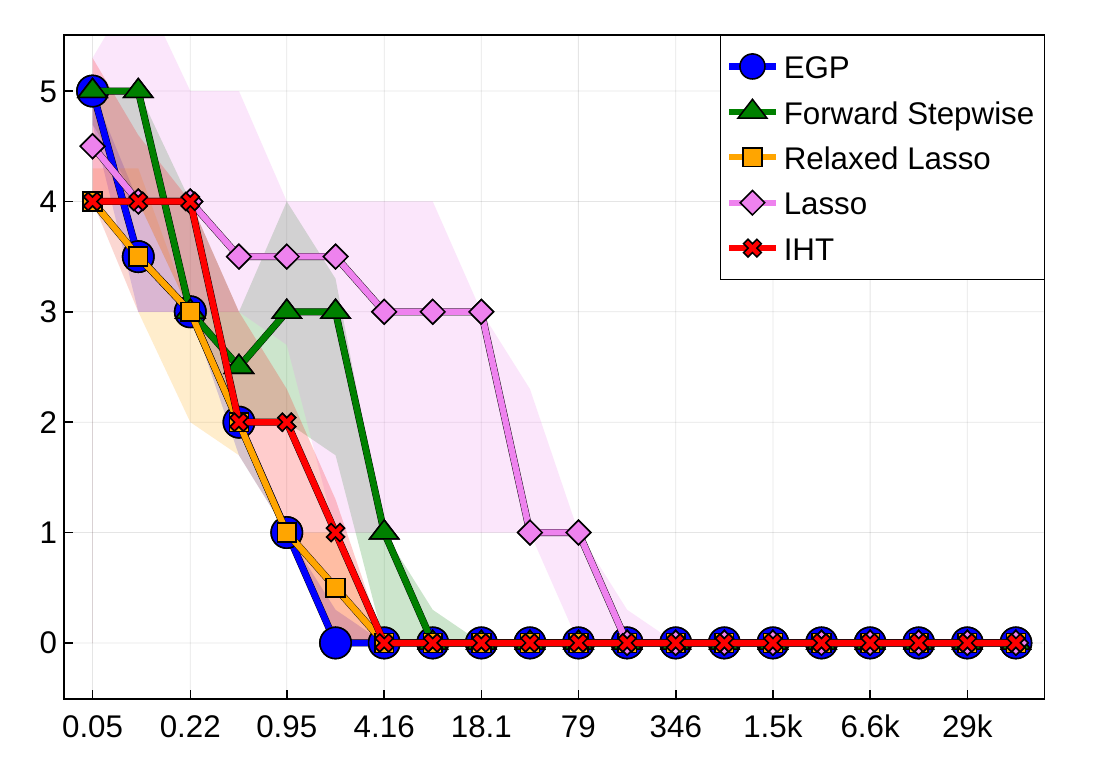}
        \caption{Setting $S_1$}
        \label{fig:ASRE:s=1rho=07}
    \end{subfigure}
    \hfill
    \begin{subfigure}[t]{0.49\textwidth}
        \includegraphics[width=\linewidth]{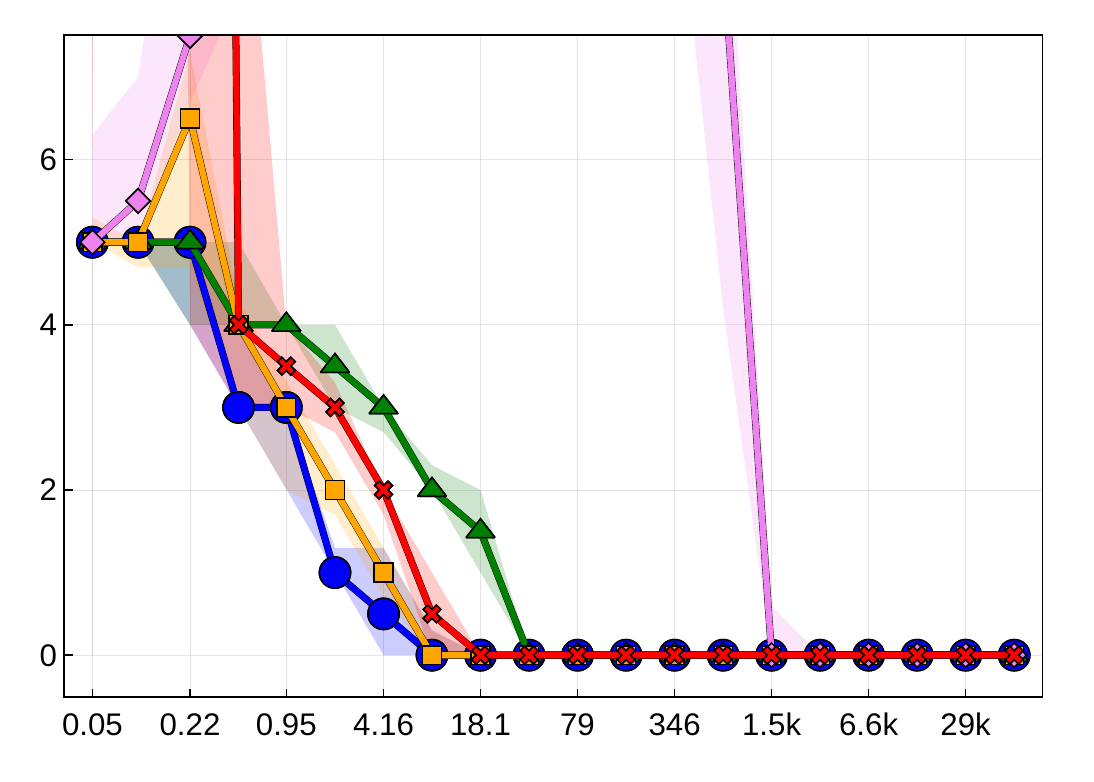}
        \caption{Setting $S_2$}
        \label{fig:ASRE:s=2_rho=07}
    \end{subfigure}\\
    \begin{subfigure}[t]{0.49\textwidth}
        \includegraphics[width=\linewidth]{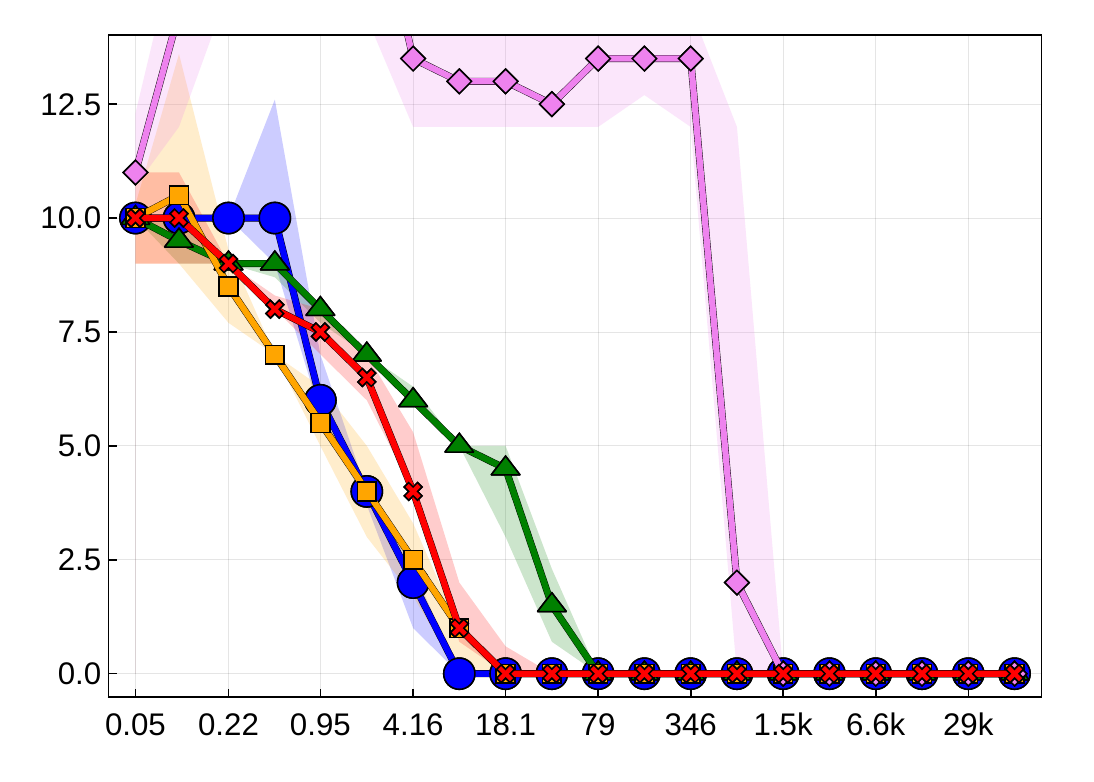}
        \caption{Setting $S_3$}
        \label{fig:ASRE:s=3rho=07}
    \end{subfigure}
    \hfill
    \begin{subfigure}[t]{0.49\textwidth}
        \includegraphics[width=\linewidth]{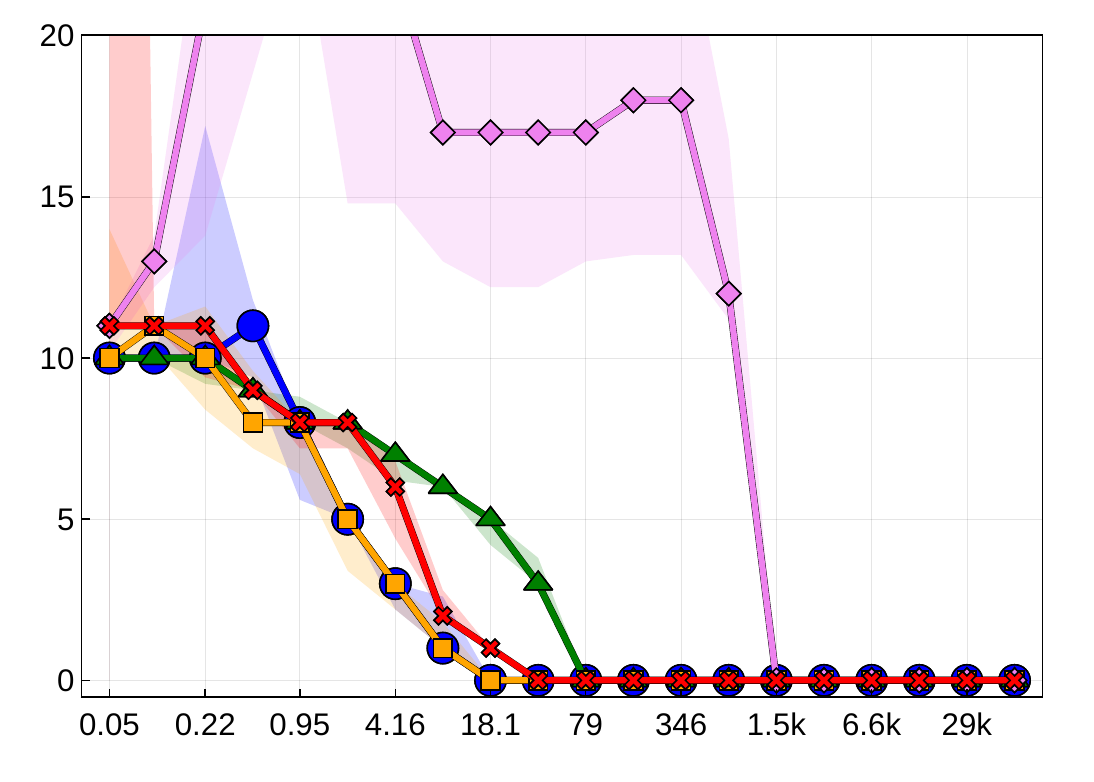}
        \caption{Setting $S_4$}
        \label{fig:ASRE:s=4rho=07}
    \end{subfigure}
    \caption{Active set reconstruction error (ASRE, cf.~\eqref{eq:ASRE}) as a function of signal-to-noise ratio (SNR, cf.~Subsection \ref{sec:SNRs}) for the different settings described in \eqref{eq:settings} and (auto)correlation level $\rho=0.7$. Medians over $10$ runs with quantile range between $0.3$ and $0.7$. Legend holds for all subfigures.}
    \label{fig:ASRE:rho07}
\end{figure}

We can observe that for almost all SNRs, EGP performs better than all other methods in all settings. 
For highly correlated coefficients in the very high noise regime, EGP does not always achieve the lowest relative test error. However, even then EGP often performs surprisingly good and sometimes better than all other methods.

Furthermore, EGP is always among the best performing methods when it comes to active set reconstruction. For example in Figure \ref{fig:ASRE:rho0}, we can observe that, in contrast to EGP, Lasso and Relaxed Lasso are not capable of reducing ASRE to $0$, not even in the very high SNR regime. Setting $S_4$ seems to be particularly challenging because there not even Forward Stepwise achieves perfect active set reconstruction, which usually performs well in high SNR regimes. 

Since EGP by default employs a combination of $\ell_0$  and $\ell_1$ regularization (cf. Subsection \ref{sec:L1L2combinationsEGP}), we perform an ablation study with simulations that only use $\ell_0$ regularization and present the results in Appendix \ref{app:ablationEGP}. One can observe that $\ell_1$ regularization can sometimes be helpful in the very-high-noise regime but overall the results are very similar to the simulations that perform combined ($\ell_0$  and $\ell_1$ ) regularization. EGP's $\ell_2$ regularization is by default deactivated for all simulations in this article.

The plots also show that Lasso performs very poorly for high SNRs. This is due to the shrinkage effect that Lasso has on the optimized coefficients. It is thus absolutely essential to relax Lasso, as explained in the introduction. Unfortunately, Relaxed Lasso is still not as widely known as Lasso and therefore many benchmarks in the literature only compare to Lasso (e.g.~\citep{bertsimas2016best, yin2020probabilistic,chu2020IHT}), which is very easily outperformed at high SNRs. Future benchmarks should therefore always consider Relaxed Lasso instead.

Overall, these results show that EGP is extremely competetive. The simplicity of integrating it into deep learning frameworks furthermore makes it an excellent candidate for solving compressive sensing tasks on large distributed computing and data infrastructure.

\subsubsection{Runtime benchmarks}
\label{sec:runtimeBenchmarks}

\begin{figure}[ht]
    \begin{subfigure}[t]{0.49\textwidth}
        \includegraphics[width=\linewidth]{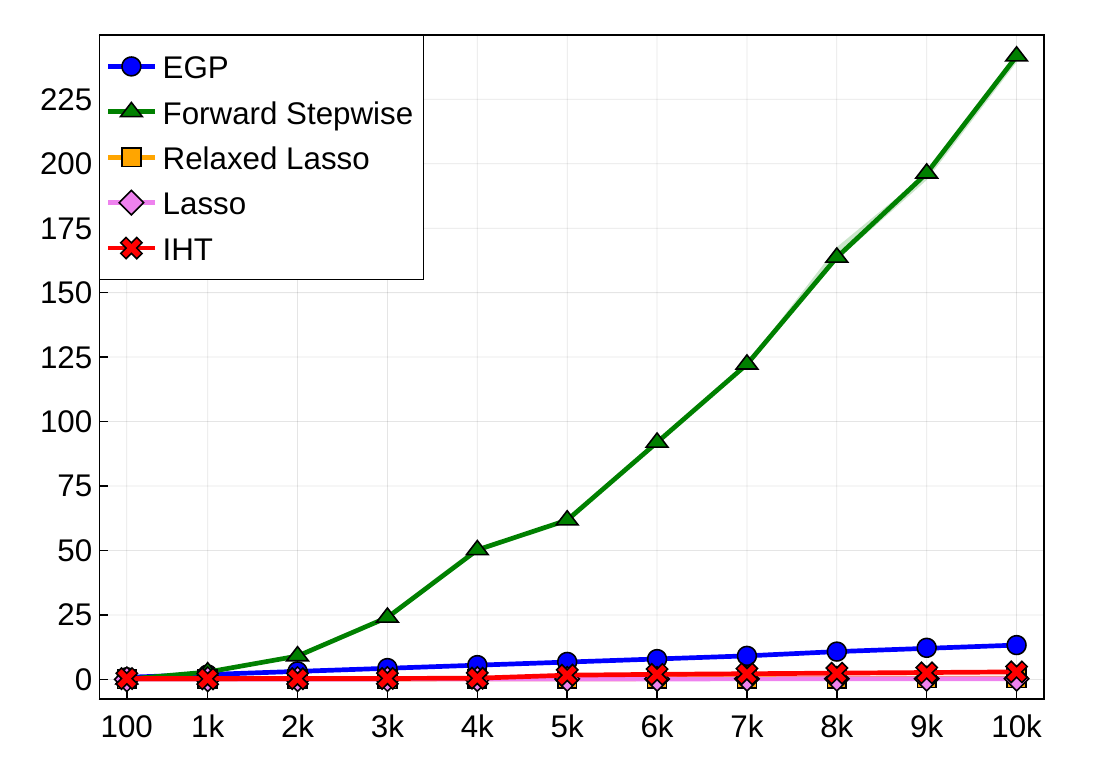}
        \caption{Varying $n$ (on $x$-axis), constant $p=100$}
        \label{fig:runtime_n}
    \end{subfigure}
    \hfill
    \begin{subfigure}[t]{0.49\textwidth}
        \includegraphics[width=\linewidth]{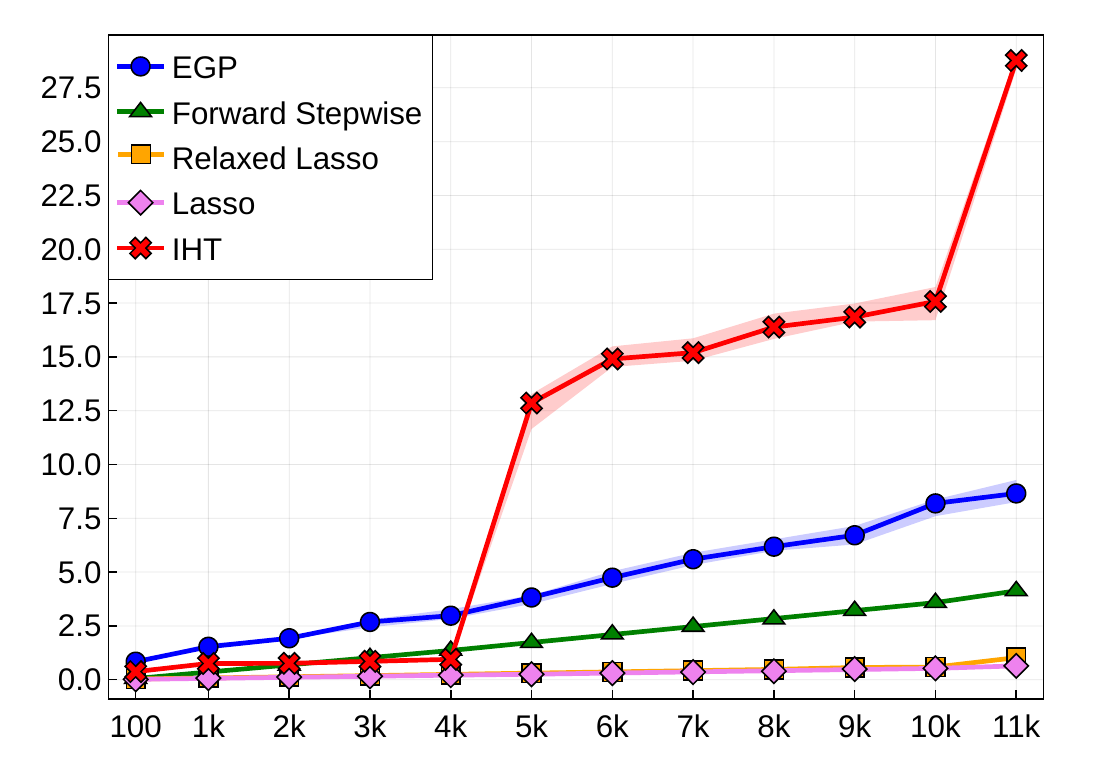}
        \caption{Constant $n=100$, varying $p$ (on $x$-axis)}
        \label{fig:runtime_p}
    \end{subfigure}
    \caption{Runtime (in seconds) as a function of either $n$ or $p$. Plot shows medians over $10$ runs with quantile range between $0.3$ and $0.7$.}
    \label{fig:runtimeComparison}
\end{figure}

To conclude this Subsection, we present some comparison of the runtimes of the five tested methods, each run with the same hyperparameters values that were used to run the main experiments.
The two fundamental parameters that determine the runtime are $n$ and $p$, that occur in the definition of a setting, cf.~\eqref{eq:settings}, while there is no or only weak dependence on $s$. 
In Figure \ref{fig:runtimeComparison}
we therefore show how the runtime (in seconds) varies for the different methods, as we vary either $n$ or $p$, while holding the other variables fixed. Data is generated as described in Subsection \ref{sec:settings}, with $\rho$ and $\text{SNR}$ fixed.

We see that Forward Stepwise performs worse than all other methods because its runtime increases quadratically with $n$. This is due to the (guided) series of QR decompositions employed by the algorithm. 

The runtime of the remaining methods seems to depend approximately linearly on both $n$ and $p$. However, for IHT we observe some abrupt jumps of the runtime as a function of $p$. (We are not sure why that happens.) 

EGP performs better than IHT (as Figure \ref{fig:MonteCarloMethodComparison2} shows) but is slower than Lasso and Relaxed Lasso, whose runtime is impressively low. However, since the runtime of all three methods depends linearly on $n$ and $p$, this factor at least does not change much with the dataset size. 
We computed that EGP is on average $26 \pm 3$ times slower than Relaxed Lasso for varying $n$ and $14 \pm 4$ times slower for varying $p$.
Despite this, the absolute runtime values for EGP are quite small,
and computations finished after about $14$ seconds for $n=11,000,~p=100,~np=1,100,000$ and about $9$ seconds for $n = 100,~p = 11,000,~np = 1,100,000$, on a single machine, whose hardware is detailed in Appendix \ref{app:computeresources}.

Since EGP runtime depends linearly on dataset and model coefficient size, EGP is very scalable, even if Relaxed Lasso is faster by an approximately constant factor. 
Finally, since EGP is gradient descent based, it is very easy to integrate it into sophisticated deep learning frameworks like Pytorch \citep{paszke2019pytorchimperativestylehighperformance} or Lux \citep{pal2023lux}, that can make use of large distributed compute resources.

\subsection{Teacher-Student Parameter Recovery Experiments}
\label{sec:TeacherStudentExperiments}

Building on the theoretical results from Section \ref{sec:FeffermanMarkelTheory}, we now empirically test 
in how far the convergence results in the infinite data limit are applicable to the finite data regime.
We conduct experiments focusing on the question of whether regularized or unregularized optimization can recover the ground truth parameters in a teacher-student setup--up to permutation of neurons and sign flips.

Using Fefferman-Markel networks ($\tanh$-MLPs satisfying the 3 generic conditions defined in Section \ref{sec:FeffermanMarkelTheory}), we generate teacher-student pairs and sample training and test datasets from the teacher. 
Both teacher and student have the same input-output dimensions but potentially different hidden layer sizes. The teacher network is ``sparse'' in the sense that its hidden layers are smaller than those of the student, creating an over-parameterized student that must learn to compress its representation.

\subsubsection{Compression improves convergence of functions but not of parameters}
\label{sec:functionalVsParameterConvergence}

\begin{figure}[ht]
    \centering
    \begin{subfigure}[t]{0.35\textwidth}
        \includegraphics[width=\linewidth]{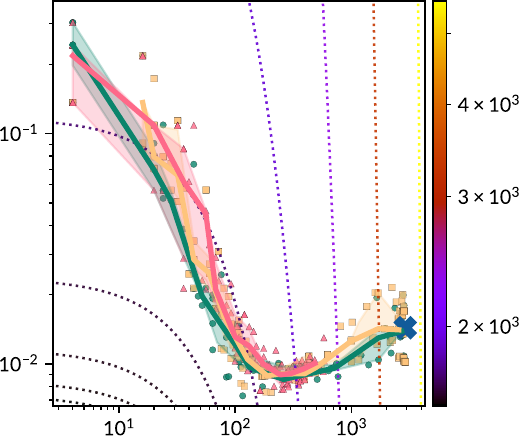}
    \end{subfigure}
    $\qquad$
    \begin{subfigure}[t]{0.35\textwidth}
        \includegraphics[width=\linewidth]{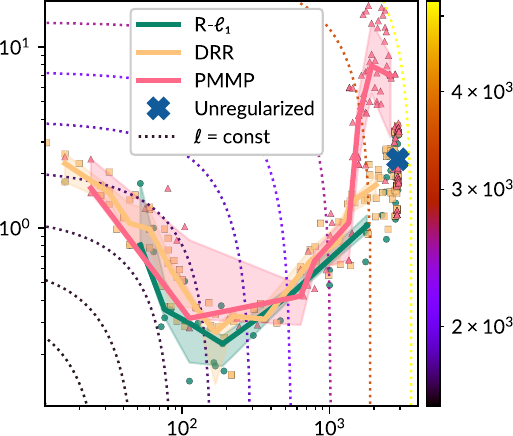}
    \end{subfigure}
    \caption{`Test Loss` vs `Model Byte Size` for teacher-student setup. Left: ($\ell_0$ regularized) MSE loss. Right: Gauss loss (see eq.~\eqref{eq:GaussianLoss}). Description length (in bytes) isolines are color coded. Curves and error bars were computed as described in \citep[Appendix I]{barth2025efficientcompression}. Teacher and student architectures have layer sizes [2, 5, 8, 1] and [2, 25, 25, 1], respectively. The student was trained on $300$ data samples generated by the teacher network with standard deviation $\sigma=0.08$.}
    \label{fig:student-teacher-losses}
\end{figure}

One strength of (linear) compressive sensing is that regularization can enable more sample efficient convergence to the true distribution, thus lowering test error in comparison to unregularized regression. This effect is especially strong when the amount of data is small in comparison to the number of variables.
In \citep{barth2025efficientcompression}, we showed that this effect carries over to the nonlinear setting and that Solomonoff induction theory \citep{solomonoff1978complexity,poland2004convergence} can be used to explain this. To illustrate this, we display two plots taken from \citep{barth2025efficientcompression} in Figure \ref{fig:student-teacher-losses}, where we tested DRR, R-L1 and PMMP (summarized in Section \ref{sec:nonlinearL0Regularization}).

One can observe that the lowest test error is achieved by the networks which are slightly larger than those that are the best data compressors (which in turn are those that achieve to touch the left-most isolines). Hence, to the extent to which compressive sensing is about using compression for sample efficient convergence, it works in the nonlinear setting.

However, compressive sensing is usually also about exact parameter recovery. Here we thus subsequently investigate in how far this is possible.

As a first step, we visualize the pruned student networks that achieve lowest test losses and compare their final parameter configurations with the teacher configuration. As in Figure \ref{fig:student-teacher-losses}, the student architecture is set to [2,25,25,1] (2 hidden layers with 25 neurons each) and the teacher architecture to [2,5,8,1]. The student is trained with DRR described in Subsection \ref{sec:nonlinearL0Regularization}. To compare the parameters up to symmetries, we apply the normal form algorithm (that we described before Theorem \ref{thm:fefferman-markel}) to the students after training. The pruning is carried out with a binary search procedure described in \citep{barth2025efficientcompression}, which deletes all parameters whose deletion does not decrease the validation loss by more than a small percentage.

\begin{figure}[ht]
    \centering
    \begin{subfigure}[t]{0.28\textwidth}
        \includegraphics[width=\linewidth]{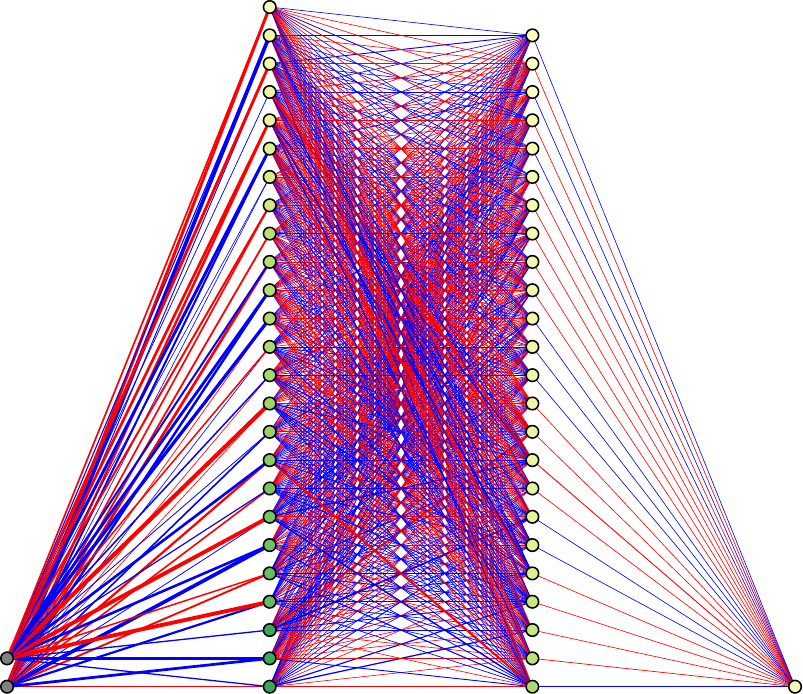}
        \caption{$\lambda = 0.0$}
        \label{fig:student-teacher-losses:0.0}
    \end{subfigure}
    \hfill
    \begin{subfigure}[t]{0.28\textwidth}
        \includegraphics[width=\linewidth]{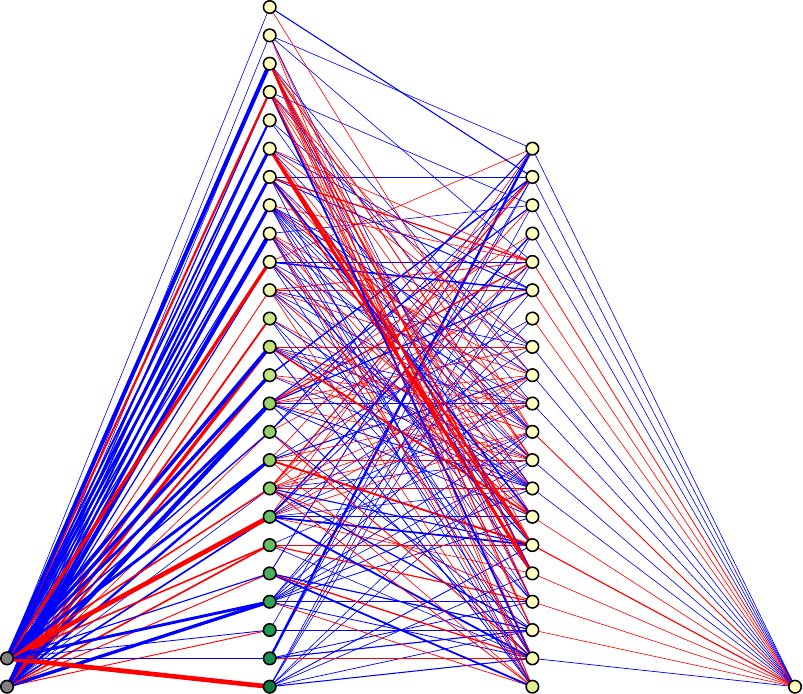}
        \caption{$\lambda = 0.2$}
        \label{fig:student-teacher-losses:0.2}
    \end{subfigure}
    \hfill
    \begin{subfigure}[t]{0.28\textwidth}
        \includegraphics[width=\linewidth]{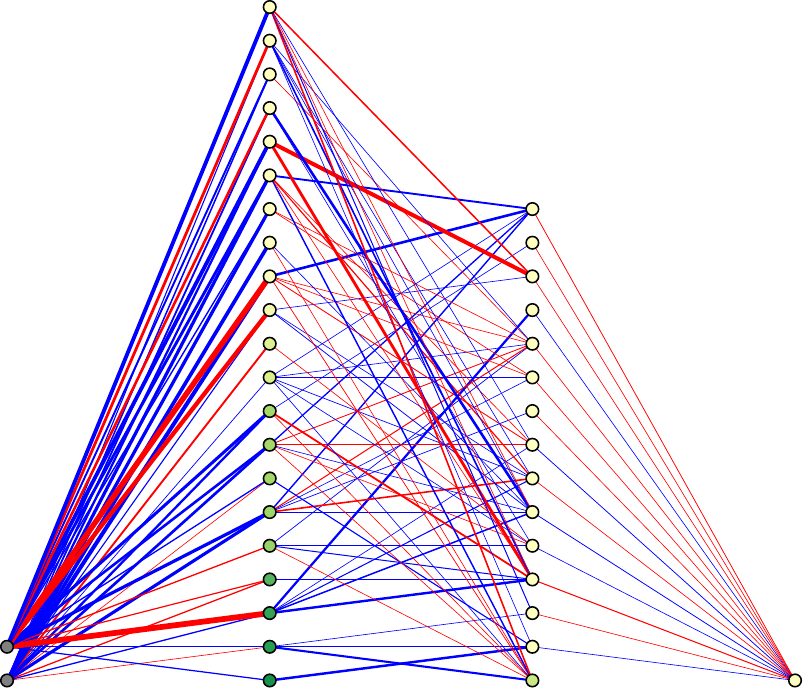}
        \caption{$\lambda = 0.4$}
        \label{fig:student-teacher-losses:0.4}
    \end{subfigure}
    \\
    \begin{subfigure}[t]{0.28\textwidth}
        \includegraphics[width=\linewidth]{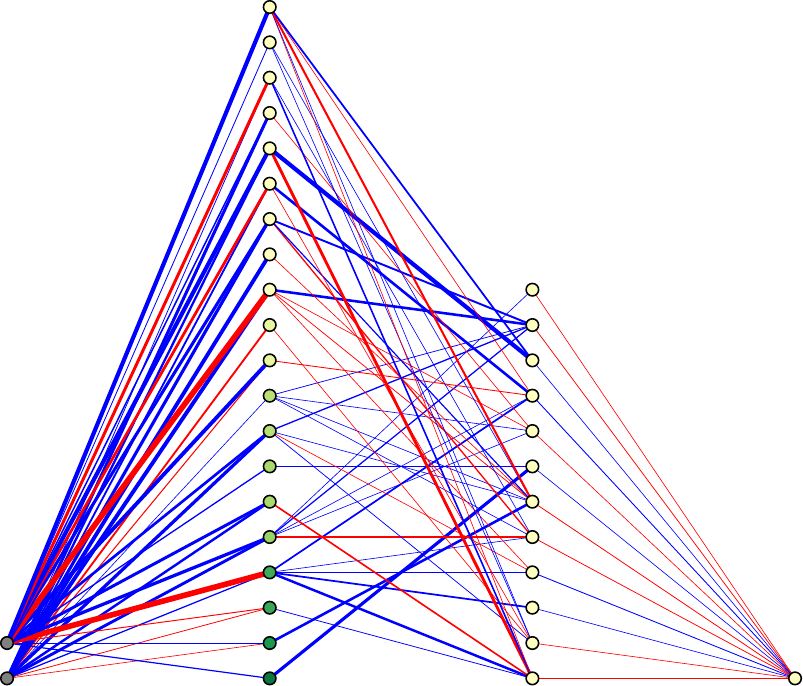}
        \caption{$\lambda = 0.6$}
        \label{fig:student-teacher-losses:0.6}
    \end{subfigure}
    \hfill
    \begin{subfigure}[t]{0.28\textwidth}
        \includegraphics[width=\linewidth]{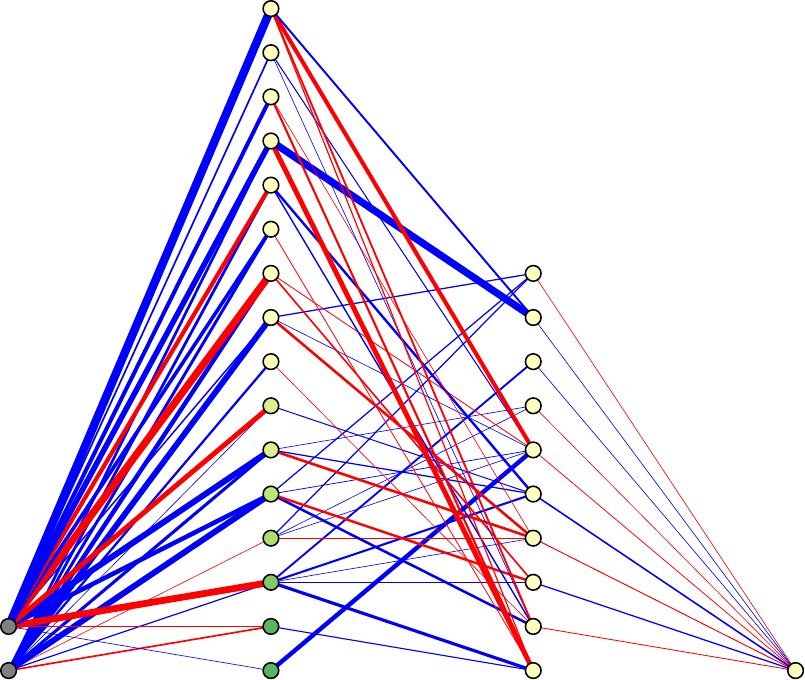}
        \caption{$\lambda = 0.8$}
        \label{fig:student-teacher-losses:0.8}
    \end{subfigure}
    \hfill
    \begin{subfigure}[t]{0.28\textwidth}
        \includegraphics[width=\linewidth]{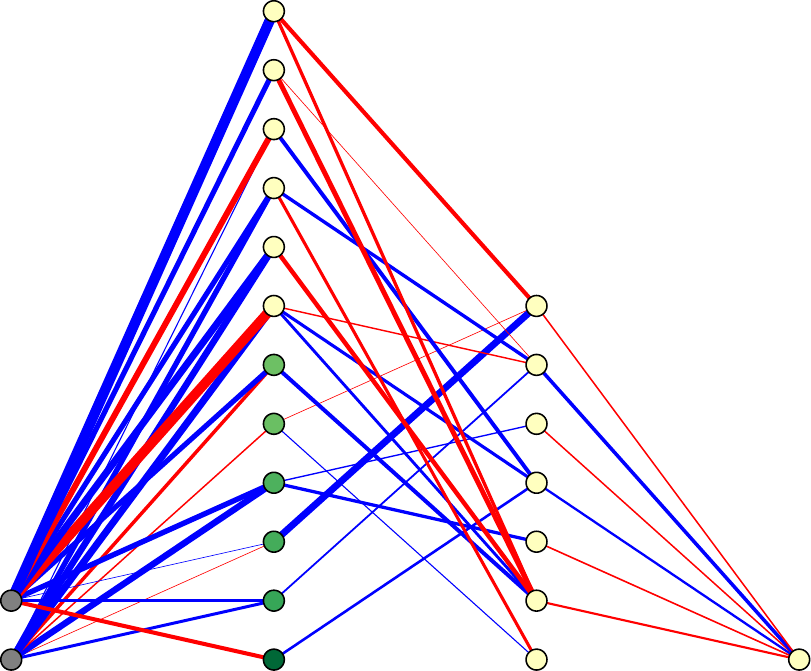}
        \caption{$\lambda = 1.0$}
        \label{fig:student-teacher-losses:1.0}
    \end{subfigure}
    \caption{Student network normal form parameter configurations after training across different $\lambda$ values (cf.~eq.~\eqref{eq:optimizationObjectiveTeacherStudent}) indicated below the subfigures. Teacher and student architectures have layer sizes [2, 5, 8, 1] and [2, 25, 25, 1], respectively. The student was trained on $300$ data samples generated by the teacher network with standard deviation $\sigma=0.08$.}
    \label{fig:student-teacher-parameter-visualization}
\end{figure}

\begin{figure}[ht]
    \centering 
    \begin{subfigure}[t]{0.5\textwidth}
        \includegraphics[width=\linewidth]{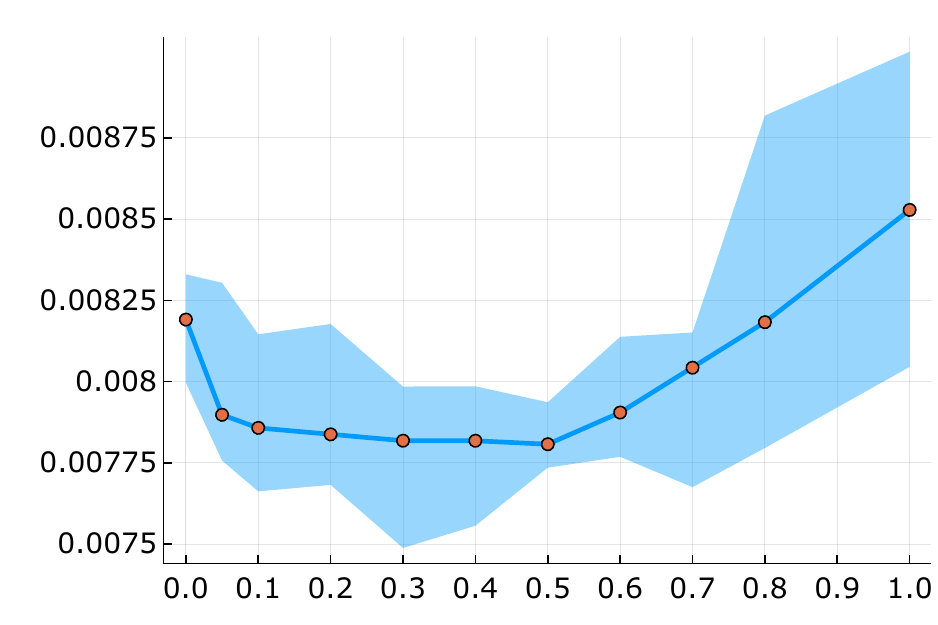}
        \caption{Test loss vs $\lambda$}
        \label{fig:student-teacher-testloss-vs-lambda:a}
    \end{subfigure}
    $\qquad$
    \begin{subfigure}[t]{0.35\textwidth}
        \includegraphics[width=\linewidth]{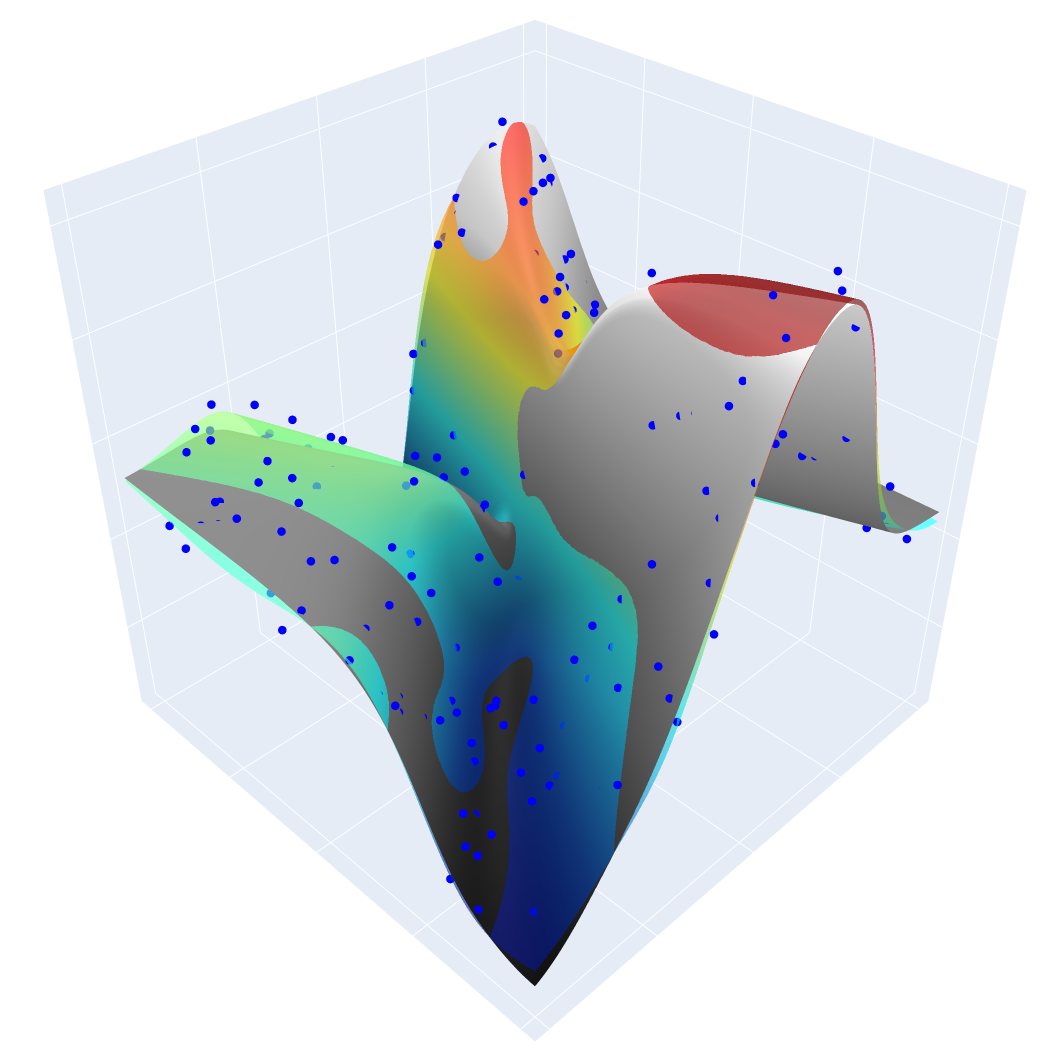}
        \caption{Function overlap at $\lambda = 0.3$}
        \label{fig:student-teacher-lambda:0.2}
    \end{subfigure}
    \caption{(a) Test loss as a function of $\lambda$ values (cf.~eq.~\eqref{eq:optimizationObjectiveTeacherStudent}). Plotpoints correspond to medians across 10 runs and the shaded regions to the interquartile range (IQR). The students were trained with DRR on $300$ data samples generated by the teacher network with standard deviation $\sigma=0.08$. (b) Overlap of the teacher and student functions is shown to illustrate the convergence quality. Teacher function (grey), student function (rainbow color, after training) and teacher-generated training dataset (blue dots).}
    \label{fig:student-teacher-testloss-vs-lambda}
\end{figure}

In Figure \ref{fig:student-teacher-parameter-visualization}, we can see how the normalized student networks become progressively smaller as the regularization strength $\lambda$ increases. Meanwhile Figure \ref{fig:student-teacher-testloss-vs-lambda:a} shows how the test loss depends on $\lambda$. It exhibits a U-shape, similar to those visible in Figure \ref{fig:student-teacher-losses}, further supporting the claim that mild regularization can help to improve the test loss. The optimal $\lambda$ value seems to be between $\lambda=0.3$ and $\lambda=0.5$ in this case. 

However, Figure \ref{fig:student-teacher-parameter-visualization} shows that for such $\lambda$ values (at which lowest test loss is obtained), the normal form parameter configurations of the students are not as sparse as that of the original teacher, which is visualized on the left side of Figure \ref{fig:TeacherStudentSetup}. Since higher $\lambda$ values would be necessary to achieve the sparsity of the teacher but higher $\lambda$ values also increase the test loss, we can conclude that there are no $\lambda$ values for which our optimization procedure achieves exact parameter recovery up to symmetries (for the tested initializations). Furthermore, the final normal form configurations of the students also look rather different from that of the teacher. This indicates that the students converge to local optima that are very different from the optimum that corresponds to the teacher network, even though Figure \ref{fig:student-teacher-testloss-vs-lambda} shows that students achieve very low test error (at the order of the standard deviation of the data) and that the overlap of teacher and student functions is very strong. Since we compare normal forms, we can conclude that even up to symmetry the optimization landscape is highly non-convex and that (regularized) SGD of randomly initialized students typically converges to optima that are not just different but even in different equivalence classes.

\subsubsection{\texorpdfstring{$\ell_2$}{L2} distance analysis reveals limitations of parameter recovery}
\label{sec:optimizationLandscapeRecovery}

In order to investigate further if convergence of student network parameter configurations to teacher configurations is possible at all, we subsequently introduce further experiments, in which we initialize the student very close to the teacher network and keep the noise of the teacher-generated data very small. Subsequently, we assume that the teacher and student both have architecture $[2,25,25,1]$ (2 hidden layers with 25 neurons each). Since pruning is not required in this case, we train the student without regularization.

To measure convergence in parameter space, we use the $\ell_2$-distance of the teacher and student networks in normal form, $\theta_T, \theta_S \in \mathbb{R}^p$, that is
\begin{equation}
    \|\theta_T - \theta_S\|_2 = \sqrt{\sum_i (\theta_{T,i} - \theta_{S,i})^2}.
\end{equation}
Proposition \ref{prop:nn_isomorphim_loss_minimization} states that the $\ell_2$-distance should converge to zero in the limit of large training sets.
To isolate different effects, we designed three focused experiments:
\begin{enumerate}
    \item \textit{Near-optimal initialization:} Students initialized in an $\epsilon$-neighborhood of teacher parameters $\theta_T$ and trained on noise-free data.
    \item \textit{Exact initialization with noise:} Students initialized exactly at $\theta_T$ and trained on data with small Gaussian noise.
    \item \textit{Noise sensitivity analysis:} Students initialized at $\theta_T$ and trained for a fixed number of iterations on datasets with varying noise levels.
\end{enumerate}

\begin{figure}[ht]
    \centering
    \begin{subfigure}[t]{0.33\textwidth}
        \includegraphics[width=\linewidth]{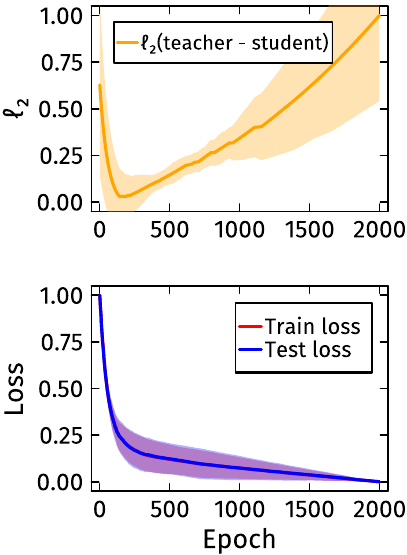}$~$
        \caption*{(a)}
    \end{subfigure}
    \hfill
    \begin{subfigure}[t]{0.33\textwidth}
        \includegraphics[width=\linewidth]{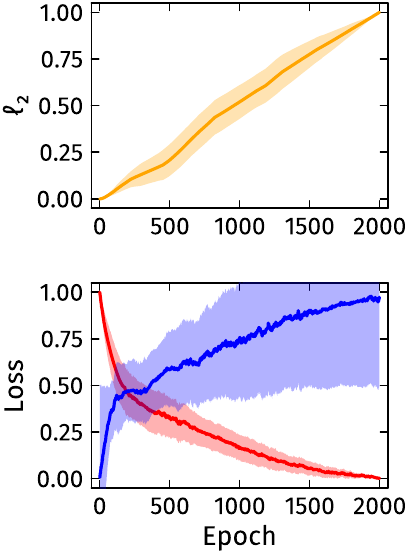}$~$
        \caption*{(b)}
    \end{subfigure}
    \hfill
    \begin{subfigure}[t]{0.29\textwidth}
        \includegraphics[width=\linewidth]{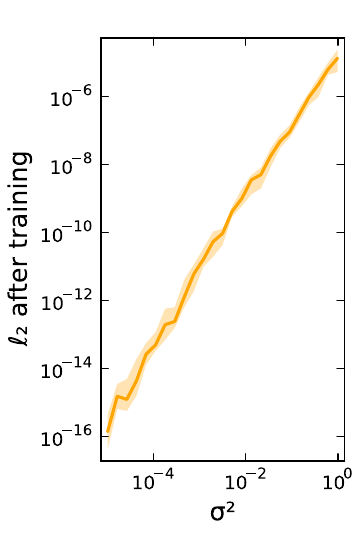}$~$
        \caption*{(c)}
    \end{subfigure}
    \caption{The $\ell_2$ rebound phenomenon in teacher-student parameter recovery: (a) Small initial perturbations ($\theta_i \leftarrow \theta_i +10^{-6} x_i$ where $x_i \sim \mathcal{U}([-1,1])$) from teacher parameters lead to initial convergence followed by divergence despite decreasing loss (training loss overlaps with test loss and is not visible); (b) Training from exact teacher parameters on noisy data ($\sigma^2=10^{-3}$) shows monotonic parameter divergence while loss decreases; (c) Log-log linear relationship between dataset noise level and equilibrium parameter distance after 100 epochs. All experiments use MLP architecture [2,25,25,1] with SGD optimization, showing median over 20 trials with interquartile range shading. Parameters are compared in normal form to account for network symmetries. Training set size: 10000, learning rate: $10^{-3}$.}
    \label{fig:rebound-phenomenon}
\end{figure}

In the first experiment, student parameters are initialized near teacher parameters,
\begin{equation}
\theta_i \leftarrow \theta_i + p x_i,
\end{equation} 
where $x_i \sim \mathcal{U}([-1,1])$ and $p=10^{-6}$ is the perturbation strength. As Figure \ref{fig:rebound-phenomenon}a, when students are trained on clean data, the $\ell_2$-distance in normal form initially decreases as expected. However, after reaching a minimum, it begins to increase significantly while both training and test losses continue to improve. 
We term this the \textit{$\ell_2$ rebound phenomenon}. It might be related to the fact that minima in loss landscapes of overparameterized networks are essentially without barrier as described in \citep{draxler2019essentiallybarriersneuralnetwork}.
This result was consistent across different learning rates, dataset sizes, and perturbation strengths. 
To ensure that the graphs in Figure \ref{fig:rebound-phenomenon} are statistically representative, we perform the experiment over $N=20$ runs, normalize the curves such that their range is within $[0,1]$, and display the median curve with interquartile range shading.

For the second experiment, the student parameters are initialized to exactly match those of the teacher. 
We then train the student on a dataset sampled from the teacher with small Gaussian noise added to the data. 
We find that the student parameters consistently diverge from the teacher as the training loss decreases. Since test loss simultaneously increases, this indicates overfitting, even though the noise is very small and the training set size of 10,000 is relatively large for this problem.
Moreover, the $\ell_2$ distance increases linearly, while the test loss only seems to exhibit logarithmic growth.

Finally, for the third experiment, we fix the number of training steps and study the final parameter distance as a function of the dataset noise. We observe an almost linear relationship in the log-log plot.

These findings--in particular Figure \ref{fig:rebound-phenomenon}a--suggest fundamental limitations in extending compressive sensing theory to the nonlinear setting of MLP-based machine learning. 
While Propositions \ref{prop:nn_isomorphim_loss_minimization} and \ref{prop:nn_isomorphim_loss_minimization_l0} guarantee parameter recovery in the infinite data limit, practical gradient-based optimization discovers functionally equivalent but structurally distant parameter configurations.
This confirms the conclusion of the previous subsection that even up to symmetry the optimization landscape is highly non-convex. Furthermore, this finding even holds in a very small neighborhood of the global optimum, indicating that even on small scales, the optimization landscape can exhibit a complex structure.
This suggests that the theoretical guarantees may not be practically relevant for finite datasets and gradient-based optimization.
Unlike linear compressive sensing, where parameter recovery and function approximation are closely linked, in our nonlinear setting, there appears to be a decoupling between these objectives.

These results show that parameter identifiability in neural networks remains a significant challenge even under the controlled conditions of Fefferman-Markel theory and that developing interpretable neural network compression methods may require fundamentally different approaches than those successful in linear settings.

\section{Conclusion, limitations and outlook}
\label{sec:conclusion}

We presented a probabilistic reformulation of $\ell_0$ regularized regression called EGP that is orders of magnitude faster than Monte Carlo based methods and outperforms other compressive sensing methods like Lasso, Relaxed Lasso, IHT and Forward Stepwise in a wide range of settings and signal-to-noise ratios.
Furthermore, we showed that exact parameter recovery of a wide class of nonlinear multilayer perceptrons is theoretically guaranteed up to certain symmetries in the infinite-data limit but empirically verified that there are fundamental limits for parameter recovery in the finite data regime. Nevertheless, we provided additional evidence that nonlinear compressive sensing and $\ell_0$ regularization can help to improve the test loss and make convergence more sample-efficient.

EGP is broadly applicable to linear regularized regression problems. It can exhibit slight sensitivity to hyperparameter choices such as learning rate or initial conditions. Nevertheless, as remarked in Appendix \ref{app:coefficientSelection} and \ref{app:EGPhyperparameters}, apart from the parameters that are varied for model selection, all experiments presented in Subsection \ref{sec:compressiveSensingExperiments} and in the appendix (4 settings, 4 correlation levels and 20 SNRs, totalling 320 different scenarios) were performed with only a single hyperparameter choice. This is possible because our method allows to conduct simulations for all remaining selection parameter values in parallel at a comparatively low cost as explained in Subsection \ref{sec:convToGoodOptima}.
The Fefferman-Markel based parameter recovery theory that we developed in Section \ref{sec:FeffermanMarkelTheory} is restricted to multilayer perceptrons with tanh activation functions. Nevertheless, since these are universal function approximators, their class is at least more general than other classes that we have seen in the nonlinear compressive sensing literature.

We would be excited to see EGP being applied in future work, in particular in genome association studies similar to those presented in \citep{chu2020IHT} and research that explores pruning of neural ODEs and PDEs, whose solutions can exhibit highly nonlinear behavior, as explained around eq.~\eqref{eq:ODE}. We would also be very interested in theory that helps to understand the $\ell_2$ rebound phenomenon that we discussed in Section \ref{sec:TeacherStudentExperiments}. 


\newpage
\acks{  
  We thank Jürgen Jost and Guido Montúfar for helpful discussions related to the topic of this article. Furthermore, we thank Janis Keck for proofreading the article and for providing the alternative proof of Theorem 1 described in Appendix \ref{app:alternativeProof} after that.
  
  We gratefully acknowledge support from the Max Planck Institute for Mathematics in the Sciences and the German Academic Scholarship Foundation.
  
  The authors declare that they have no competing interests (related financial activities outside the submitted work).
}

\section*{Contributions}

Lukas Barth developed the probabilistic reformulation of $\ell_0$ regularized regression (Section \ref{sec:probReformulation}), implemented the Exact Gradient Pruning (EGP) method, conducted the experimental comparison with Monte Carlo methods (Section \ref{sec:ComparisonWithMonteCarloMethods}), the systematic compressive sensing experiments (Section \ref{sec:compressiveSensingExperiments}), the nonlinear pruning experiments (Section \ref{sec:functionalVsParameterConvergence}), and wrote the majority of the manuscript. 

Paulo von Petersenn developed the theoretical framework for nonlinear compressive sensing based on Fefferman-Markel networks (Section \ref{sec:FeffermanMarkelTheory}), conducted the teacher-student $\ell_2$ distance experiments (Section \ref{sec:optimizationLandscapeRecovery}), and contributed to the theoretical analysis of neural network parameter identifiability.

Both authors equally contributed to the implementation of the general features of the nonlinear pruning teacher-student setup.


\vskip 0.2in
\bibliography{bib}

\begin{thebibliography}{74}
\providecommand{\natexlab}[1]{#1}
\providecommand{\url}[1]{\texttt{#1}}
\expandafter\ifx\csname urlstyle\endcsname\relax
  \providecommand{\doi}[1]{doi: #1}\else
  \providecommand{\doi}{doi: \begingroup \urlstyle{rm}\Url}\fi

\bibitem[Abrol et~al.(2015)Abrol, Sharma, and Sao]{abrol2015voiced}
V.~Abrol, P.~Sharma, and A.~K. Sao.
\newblock Voiced/nonvoiced detection in compressively sensed speech signals.
\newblock \emph{Speech Communication}, 72:\penalty0 194--207, 2015.

\bibitem[Barth and von Petersenn(2025)]{barth2025efficientcompression}
L.~S. Barth and P.~von Petersenn.
\newblock Efficient compression of neural networks and datasets, 2025.
\newblock URL \url{https://arxiv.org/abs/2505.17469}.

\bibitem[Beale et~al.(1967)Beale, Kendall, and Mann]{Beale1967}
E.~M.~L. Beale, M.~G. Kendall, and D.~W. Mann.
\newblock The discarding of variables in multivariate analysis.
\newblock \emph{Biometrika}, 54\penalty0 (3-4):\penalty0 357--366, 12 1967.
\newblock ISSN 0006-3444.
\newblock \doi{10.1093/biomet/54.3-4.357}.
\newblock URL \url{https://doi.org/10.1093/biomet/54.3-4.357}.

\bibitem[Bertsimas et~al.(2016)Bertsimas, King, and Mazumder]{bertsimas2016best}
D.~Bertsimas, A.~King, and R.~Mazumder.
\newblock Best subset selection via a modern optimization lens.
\newblock \emph{arXiv preprint arXiv:1507.03133}, 2016.

\bibitem[Bezanson et~al.(2017)Bezanson, Edelman, Karpinski, and Shah]{bezanson2017juliaProgrammingLanguage}
J.~Bezanson, A.~Edelman, S.~Karpinski, and V.~B. Shah.
\newblock Julia: A fresh approach to numerical computing.
\newblock \emph{SIAM review}, 59\penalty0 (1):\penalty0 65--98, 2017.
\newblock URL \url{https://doi.org/10.1137/141000671}.

\bibitem[Blumensath(2013)]{blumensath2013nonlinear}
T.~Blumensath.
\newblock Compressed sensing with nonlinear observations and related nonlinear optimization problems.
\newblock \emph{IEEE Transactions on Information Theory}, 59\penalty0 (6):\penalty0 3466--3474, 2013.

\bibitem[Blumensath and Davies(2008)]{blumensath2008iterative}
T.~Blumensath and M.~E. Davies.
\newblock Iterative thresholding for sparse approximations.
\newblock \emph{Journal of Fourier analysis and Applications}, 14:\penalty0 629--654, 2008.

\bibitem[Brunton et~al.(2014)Brunton, Tu, Bright, and Kutz]{brunton2014compressive}
S.~L. Brunton, J.~H. Tu, I.~Bright, and J.~N. Kutz.
\newblock Compressive sensing and low-rank libraries for classification of bifurcation regimes in nonlinear dynamical systems.
\newblock \emph{SIAM Journal on Applied Dynamical Systems}, 13\penalty0 (4):\penalty0 1716--1732, 2014.

\bibitem[B{\"u}hlmann and Van De~Geer(2011)]{buhlmann2011statistics}
P.~B{\"u}hlmann and S.~Van De~Geer.
\newblock \emph{Statistics for high-dimensional data: methods, theory and applications}.
\newblock Springer Science \& Business Media, 2011.

\bibitem[Candes et~al.(2006)Candes, Romberg, and Tao]{CandesTao}
E.~Candes, J.~Romberg, and T.~Tao.
\newblock Robust uncertainty principles: exact signal reconstruction from highly incomplete frequency information.
\newblock \emph{IEEE Transactions on Information Theory}, 52\penalty0 (2):\penalty0 489--509, 2006.
\newblock \doi{10.1109/TIT.2005.862083}.

\bibitem[Cartis and Thompson(2014)]{cartis2014newimprovedquantitativerecovery}
C.~Cartis and A.~Thompson.
\newblock A new and improved quantitative recovery analysis for iterative hard thresholding algorithms in compressed sensing, 2014.
\newblock URL \url{https://arxiv.org/abs/1309.5406}.

\bibitem[Chen et~al.(2017)Chen, Kung, and Comiter]{chen2017nonlinear}
H.-C. Chen, H.~Kung, and M.~Comiter.
\newblock Nonlinear compressive sensing for distorted measurements and application to improving efficiency of power amplifiers.
\newblock In \emph{2017 IEEE international conference on communications (ICC)}, pages 1--7. IEEE, 2017.

\bibitem[Chu et~al.(2020)Chu, Keys, German, Zhou, Zhou, Sobel, Sinsheimer, and Lange]{chu2020IHT}
B.~B. Chu, K.~L. Keys, C.~A. German, H.~Zhou, J.~J. Zhou, E.~M. Sobel, J.~S. Sinsheimer, and K.~Lange.
\newblock Iterative hard thresholding in genome-wide association studies: Generalized linear models, prior weights, and double sparsity.
\newblock \emph{GigaScience}, 9\penalty0 (6):\penalty0 giaa044, 2020.

\bibitem[Dai et~al.(2008)Dai, Sheikh, Milenkovic, and Baraniuk]{dai2008compressive}
W.~Dai, M.~A. Sheikh, O.~Milenkovic, and R.~G. Baraniuk.
\newblock Compressive sensing dna microarrays.
\newblock \emph{EURASIP journal on bioinformatics and systems biology}, 2009\penalty0 (1):\penalty0 162824, 2008.

\bibitem[Davenport et~al.(2010)Davenport, Hegde, Duarte, and Baraniuk]{davenport2010joint}
M.~A. Davenport, C.~Hegde, M.~F. Duarte, and R.~G. Baraniuk.
\newblock Joint manifolds for data fusion.
\newblock \emph{IEEE Transactions on Image Processing}, 19\penalty0 (10):\penalty0 2580--2594, 2010.

\bibitem[de~Resende~Oliveira et~al.(2024)de~Resende~Oliveira, Batista, and Seara]{oliveira2024compression}
F.~D. de~Resende~Oliveira, E.~L.~O. Batista, and R.~Seara.
\newblock On the compression of neural networks using l0-norm regularization and weight pruning.
\newblock \emph{Neural Networks}, 171:\penalty0 343--352, 2024.

\bibitem[Dong et~al.(2020)Dong, Mnih, and Tucker]{dong2020disarm}
Z.~Dong, A.~Mnih, and G.~Tucker.
\newblock Disarm: An antithetic gradient estimator for binary latent variables.
\newblock \emph{Advances in neural information processing systems}, 33:\penalty0 18637--18647, 2020.

\bibitem[Donoho(2006)]{Donoho}
D.~Donoho.
\newblock Compressed sensing.
\newblock \emph{IEEE Transactions on Information Theory}, 52\penalty0 (4):\penalty0 1289--1306, 2006.
\newblock \doi{10.1109/TIT.2006.871582}.

\bibitem[Draper(1998)]{draper1998applied}
N.~Draper.
\newblock \emph{Applied regression analysis}.
\newblock McGraw-Hill. Inc, 1998.

\bibitem[Draxler et~al.(2019)Draxler, Veschgini, Salmhofer, and Hamprecht]{draxler2019essentiallybarriersneuralnetwork}
F.~Draxler, K.~Veschgini, M.~Salmhofer, and F.~A. Hamprecht.
\newblock Essentially no barriers in neural network energy landscape, 2019.
\newblock URL \url{https://arxiv.org/abs/1803.00885}.

\bibitem[Efroymson(1966)]{efroymson1966stepwise}
M.~Efroymson.
\newblock Stepwise regression--a backward and forward look.
\newblock In \emph{Eastern Regional Meetings of the Institute of Mathematical Statistics}, pages 27--9, 1966.

\bibitem[Fefferman and Markel(1993)]{fefferman1993recoveringNNfromOutputs}
C.~Fefferman and S.~Markel.
\newblock Recovering a feed-forward net from its output.
\newblock In J.~Cowan, G.~Tesauro, and J.~Alspector, editors, \emph{Advances in Neural Information Processing Systems}, volume~6. Morgan-Kaufmann, 1993.
\newblock URL \url{https://proceedings.neurips.cc/paper_files/paper/1993/file/e49b8b4053df9505e1f48c3a701c0682-Paper.pdf}.

\bibitem[Foster and George(1994)]{foster1994risk}
D.~P. Foster and E.~I. George.
\newblock The risk inflation criterion for multiple regression.
\newblock \emph{The Annals of Statistics}, 22\penalty0 (4):\penalty0 1947--1975, 1994.

\bibitem[Foucart and Rauhut(2013)]{foucart_mathematical_2013}
S.~Foucart and H.~Rauhut.
\newblock \emph{A {Mathematical} {Introduction} to {Compressive} {Sensing}}.
\newblock Applied and {Numerical} {Harmonic} {Analysis}. Springer New York, New York, NY, 2013.
\newblock ISBN 978-0-8176-4947-0 978-0-8176-4948-7.
\newblock \doi{10.1007/978-0-8176-4948-7}.
\newblock URL \url{https://link.springer.com/10.1007/978-0-8176-4948-7}.

\bibitem[George et~al.(2015)George, Augustine, and Pattathil]{george2015audio}
S.~N. George, N.~Augustine, and D.~P. Pattathil.
\newblock Audio security through compressive sampling and cellular automata.
\newblock \emph{Multimedia Tools and Applications}, 74\penalty0 (23):\penalty0 10393--10417, 2015.

\bibitem[Giacobello et~al.(2009)Giacobello, Christensen, Murthi, Jensen, and Moonen]{giacobello2009retrieving}
D.~Giacobello, M.~G. Christensen, M.~N. Murthi, S.~H. Jensen, and M.~Moonen.
\newblock Retrieving sparse patterns using a compressed sensing framework: applications to speech coding based on sparse linear prediction.
\newblock \emph{IEEE Signal processing letters}, 17\penalty0 (1):\penalty0 103--106, 2009.

\bibitem[Greenshtein(2006)]{greenshtein2006best}
E.~Greenshtein.
\newblock Best subset selection, persistence in high-dimensional statistical learning and optimization under l1 constraint.
\newblock \emph{The Annals of Statistics}, 34\penalty0 (5):\penalty0 2367 -- 2386, 2006.
\newblock \doi{10.1214/009053606000000768}.
\newblock URL \url{https://doi.org/10.1214/009053606000000768}.

\bibitem[Gu et~al.(2009)Gu, Jin, and Mei]{gu2009l0}
Y.~Gu, J.~Jin, and S.~Mei.
\newblock The l0 norm constraint lms algorithm for sparse system identification.
\newblock \emph{IEEE Signal Processing Letters}, 16\penalty0 (9):\penalty0 774--777, 2009.

\bibitem[Hastie et~al.(2017)Hastie, Tibshirani, and Tibshirani]{hastie2017extended}
T.~Hastie, R.~Tibshirani, and R.~J. Tibshirani.
\newblock Extended comparisons of best subset selection, forward stepwise selection, and the lasso.
\newblock \emph{arXiv preprint arXiv:1707.08692}, 2017.

\bibitem[Hocking and Leslie(1967)]{hocking1967selection}
R.~R. Hocking and R.~Leslie.
\newblock Selection of the best subset in regression analysis.
\newblock \emph{Technometrics}, 9\penalty0 (4):\penalty0 531--540, 1967.

\bibitem[Howland et~al.(2011)Howland, Dixon, and Howell]{Howland2011Photon}
G.~A. Howland, P.~B. Dixon, and J.~C. Howell.
\newblock Photon-counting compressive sensing laser radar for 3d imaging.
\newblock \emph{Applied Optics}, 50\penalty0 (31):\penalty0 5917--5920, Nov. 2011.
\newblock \doi{10.1364/AO.50.005917}.

\bibitem[Hutter(2005)]{hutter2005universal}
M.~Hutter.
\newblock \emph{Universal artificial intelligence: Sequential decisions based on algorithmic probability}.
\newblock Springer Science \& Business Media, 2005.

\bibitem[Jaspan et~al.(2015)Jaspan, Fleysher, and Lipton]{jaspan2015compressed}
O.~N. Jaspan, R.~Fleysher, and M.~L. Lipton.
\newblock Compressed sensing mri: a review of the clinical literature.
\newblock \emph{The British journal of radiology}, 88\penalty0 (1056):\penalty0 20150487, 2015.

\bibitem[Johnson et~al.(2015)Johnson, Lin, Ungar, Foster, and Stine]{johnson2015risk}
K.~D. Johnson, D.~Lin, L.~H. Ungar, D.~P. Foster, and R.~A. Stine.
\newblock A risk ratio comparison of $ l\_0 $ and $ l\_1 $ penalized regression.
\newblock \emph{arXiv preprint arXiv:1510.06319}, 2015.

\bibitem[Kingma and Ba(2014)]{kingma2014adam}
D.~P. Kingma and J.~Ba.
\newblock Adam: A method for stochastic optimization.
\newblock \emph{arXiv preprint arXiv:1412.6980}, 2014.

\bibitem[Kool et~al.(2019)Kool, van Hoof, and Welling]{kool2019buy}
W.~Kool, H.~van Hoof, and M.~Welling.
\newblock Buy 4 reinforce samples, get a baseline for free!
\newblock \emph{ICLR}, 2019.

\bibitem[Kunes et~al.(2023)Kunes, Yin, Land, Haviv, Pe'er, and Tavar{\'e}]{kunes2023gradient}
R.~Z. Kunes, M.~Yin, M.~Land, D.~Haviv, D.~Pe'er, and S.~Tavar{\'e}.
\newblock Gradient estimation for binary latent variables via gradient variance clipping.
\newblock In \emph{Proceedings of the AAAI Conference on Artificial Intelligence}, volume 37-7, pages 8405--8412, 2023.

\bibitem[LeCun et~al.(1989)LeCun, Denker, and Solla]{lecun1989optimal}
Y.~LeCun, J.~Denker, and S.~Solla.
\newblock Optimal brain damage.
\newblock \emph{Advances in neural information processing systems}, 2, 1989.

\bibitem[Liu et~al.(2014)Liu, Zhu, Kong, Liu, Gu, Vasilakos, and Wu]{liu2014cdc}
X.-Y. Liu, Y.~Zhu, L.~Kong, C.~Liu, Y.~Gu, A.~V. Vasilakos, and M.-Y. Wu.
\newblock Cdc: Compressive data collection for wireless sensor networks.
\newblock \emph{IEEE Transactions on Parallel and Distributed Systems}, 26\penalty0 (8):\penalty0 2188--2197, 2014.

\bibitem[Louizos et~al.(2017)Louizos, Welling, and Kingma]{louizos2017learning}
C.~Louizos, M.~Welling, and D.~P. Kingma.
\newblock Learning sparse neural networks through {L}0 regularization.
\newblock \emph{arXiv preprint arXiv:1712.01312}, 2017.

\bibitem[Maddu et~al.(2022)Maddu, Cheeseman, Sbalzarini, and M{\"u}ller]{maddu2022stability}
S.~Maddu, B.~L. Cheeseman, I.~F. Sbalzarini, and C.~L. M{\"u}ller.
\newblock Stability selection enables robust learning of differential equations from limited noisy data.
\newblock \emph{Proceedings of the Royal Society A}, 478\penalty0 (2262):\penalty0 20210916, 2022.

\bibitem[Martensen et~al.(2021)Martensen, Rackauckas, et~al.]{datadrivendiffeq2021}
J.~Martensen, C.~Rackauckas, et~al.
\newblock Datadrivendiffeq.jl, July 2021.
\newblock URL \url{https://doi.org/10.5281/zenodo.5083412}.

\bibitem[Meinshausen(2007)]{meinshausen2007relaxed}
N.~Meinshausen.
\newblock Relaxed lasso.
\newblock \emph{Computational Statistics \& Data Analysis}, 52\penalty0 (1):\penalty0 374--393, 2007.

\bibitem[Meinshausen and B{\"u}hlmann(2010)]{meinshausen2010stability}
N.~Meinshausen and P.~B{\"u}hlmann.
\newblock Stability selection.
\newblock \emph{Journal of the Royal Statistical Society Series B: Statistical Methodology}, 72\penalty0 (4):\penalty0 417--473, 2010.

\bibitem[Metzler et~al.(2016)Metzler, Maleki, and Baraniuk]{metzler2016denoising}
C.~A. Metzler, A.~Maleki, and R.~G. Baraniuk.
\newblock From denoising to compressed sensing.
\newblock \emph{IEEE Transactions on Information Theory}, 62\penalty0 (9):\penalty0 5117--5144, 2016.

\bibitem[Mummadi et~al.(2019)Mummadi, Genewein, Zhang, Brox, and Fischer]{mummadi2019group}
C.~K. Mummadi, T.~Genewein, D.~Zhang, T.~Brox, and V.~Fischer.
\newblock Group pruning using a bounded-lp norm for group gating and regularization.
\newblock In \emph{German Conference on Pattern Recognition}, pages 139--155. Springer, 2019.

\bibitem[Nagesh and Li(2009)]{nagesh2009compressive}
P.~Nagesh and B.~Li.
\newblock A compressive sensing approach for expression-invariant face recognition.
\newblock In \emph{2009 IEEE conference on computer vision and pattern recognition}, pages 1518--1525. IEEE, 2009.

\bibitem[Natarajan(1995)]{natarajan1995sparse}
B.~K. Natarajan.
\newblock Sparse approximate solutions to linear systems.
\newblock \emph{SIAM journal on computing}, 24\penalty0 (2):\penalty0 227--234, 1995.

\bibitem[Otazo et~al.(2015)Otazo, Candes, and Sodickson]{otazo2015low}
R.~Otazo, E.~Candes, and D.~K. Sodickson.
\newblock Low-rank plus sparse matrix decomposition for accelerated dynamic mri with separation of background and dynamic components.
\newblock \emph{Magnetic resonance in medicine}, 73\penalty0 (3):\penalty0 1125--1136, 2015.

\bibitem[Pal(2023)]{pal2023lux}
A.~Pal.
\newblock Lux: Explicit parameterization of deep neural networks in julia, April 2023.
\newblock URL \url{https://doi.org/10.5281/zenodo.7808904}.
\newblock If you use this software, please cite it as below.

\bibitem[Pasco(1976)]{pasco1976source}
R.~C. Pasco.
\newblock \emph{Source coding algorithms for fast data compression}.
\newblock PhD thesis, Stanford University CA, 1976.

\bibitem[Paszke et~al.(2019)Paszke, Gross, Massa, Lerer, Bradbury, Chanan, Killeen, Lin, Gimelshein, Antiga, Desmaison, Köpf, Yang, DeVito, Raison, Tejani, Chilamkurthy, Steiner, Fang, Bai, and Chintala]{paszke2019pytorchimperativestylehighperformance}
A.~Paszke, S.~Gross, F.~Massa, A.~Lerer, J.~Bradbury, G.~Chanan, T.~Killeen, Z.~Lin, N.~Gimelshein, L.~Antiga, A.~Desmaison, A.~Köpf, E.~Yang, Z.~DeVito, M.~Raison, A.~Tejani, S.~Chilamkurthy, B.~Steiner, L.~Fang, J.~Bai, and S.~Chintala.
\newblock Pytorch: An imperative style, high-performance deep learning library, 2019.
\newblock URL \url{https://arxiv.org/abs/1912.01703}.

\bibitem[Pati et~al.(1993)Pati, Rezaiifar, and Krishnaprasad]{OrthogonalMatchinPursuit1993}
Y.~Pati, R.~Rezaiifar, and P.~Krishnaprasad.
\newblock Orthogonal matching pursuit: recursive function approximation with applications to wavelet decomposition.
\newblock In \emph{Proceedings of 27th Asilomar Conference on Signals, Systems and Computers}, pages 40--44 vol.1, 1993.
\newblock \doi{10.1109/ACSSC.1993.342465}.

\bibitem[Phan et~al.(2020)Phan, Nguyen, Nguyen, and Kalagnanam]{phan2020pruning}
D.~T. Phan, L.~M. Nguyen, N.~H. Nguyen, and J.~R. Kalagnanam.
\newblock Pruning deep neural networks with l\_0-constrained optimization.
\newblock In \emph{2020 IEEE International Conference on Data Mining (ICDM)}, pages 1214--1219. IEEE, 2020.

\bibitem[Poland and Hutter(2004)]{poland2004convergence}
J.~Poland and M.~Hutter.
\newblock Convergence of discrete mdl for sequential prediction.
\newblock In \emph{Learning Theory: 17th Annual Conference on Learning Theory, COLT 2004, Banff, Canada, July 1-4, 2004. Proceedings 17}, pages 300--314. Springer, 2004.

\bibitem[Qiao et~al.(2010)Qiao, Chen, and Tan]{qiao2010sparsity}
L.~Qiao, S.~Chen, and X.~Tan.
\newblock Sparsity preserving projections with applications to face recognition.
\newblock \emph{Pattern recognition}, 43\penalty0 (1):\penalty0 331--341, 2010.

\bibitem[Rani et~al.(2018)Rani, Dhok, and Deshmukh]{rani2018systematic}
M.~Rani, S.~B. Dhok, and R.~B. Deshmukh.
\newblock A systematic review of compressive sensing: Concepts, implementations and applications.
\newblock \emph{IEEE access}, 6:\penalty0 4875--4894, 2018.

\bibitem[Rissanen(1978)]{rissanen1978modeling}
J.~Rissanen.
\newblock Modeling by shortest data description.
\newblock \emph{Automatica}, 14\penalty0 (5):\penalty0 465--471, 1978.

\bibitem[Rissanen(1976)]{RissanenArithmeticCoding}
J.~J. Rissanen.
\newblock Generalized kraft inequality and arithmetic coding.
\newblock \emph{IBM Journal of Research and Development}, 20\penalty0 (3):\penalty0 198--203, 1976.
\newblock \doi{10.1147/rd.203.0198}.

\bibitem[Robinson et~al.(2024)Robinson, Moshtaghpour, Wells, Nicholls, Chi, MacLaren, Kirkland, and Browning]{robinson2024high}
A.~W. Robinson, A.~Moshtaghpour, J.~Wells, D.~Nicholls, M.~Chi, I.~MacLaren, A.~I. Kirkland, and N.~D. Browning.
\newblock High-speed 4-dimensional scanning transmission electron microscopy using compressive sensing techniques.
\newblock \emph{Journal of Microscopy}, 295\penalty0 (3):\penalty0 278--286, 2024.

\bibitem[Roesch et~al.(2021)Roesch, Rackauckas, and Stumpf]{RoeschRackauckasStumpf}
E.~Roesch, C.~Rackauckas, and M.~P.~H. Stumpf.
\newblock Collocation based training of neural ordinary differential equations.
\newblock \emph{Statistical Applications in Genetics and Molecular Biology}, 20\penalty0 (2):\penalty0 37--49, 2021.
\newblock \doi{doi:10.1515/sagmb-2020-0025}.
\newblock URL \url{https://doi.org/10.1515/sagmb-2020-0025}.

\bibitem[Sahoo et~al.(2018)Sahoo, Lampert, and Martius]{sahoo2018learning}
S.~Sahoo, C.~Lampert, and G.~Martius.
\newblock Learning equations for extrapolation and control.
\newblock In \emph{International Conference on Machine Learning}, pages 4442--4450. Pmlr, 2018.

\bibitem[Shannon(1949)]{ShannonNyquist}
C.~Shannon.
\newblock Communication in the presence of noise.
\newblock \emph{Proceedings of the IRE}, 37\penalty0 (1):\penalty0 10--21, 1949.
\newblock \doi{10.1109/JRPROC.1949.232969}.

\bibitem[Solomonoff(1978)]{solomonoff1978complexity}
R.~Solomonoff.
\newblock Complexity-based induction systems: comparisons and convergence theorems.
\newblock \emph{IEEE transactions on Information Theory}, 24\penalty0 (4):\penalty0 422--432, 1978.

\bibitem[Solomonoff(1964)]{solomonoff1964formal}
R.~J. Solomonoff.
\newblock A formal theory of inductive inference. part i.
\newblock \emph{Information and control}, 7\penalty0 (1):\penalty0 1--22, 1964.

\bibitem[Tang et~al.(2011)Tang, Cao, Duan, and Wang]{tang2011compressed}
W.~Tang, H.~Cao, J.~Duan, and Y.-P. Wang.
\newblock A compressed sensing based approach for subtyping of leukemia from gene expression data.
\newblock \emph{Journal of bioinformatics and computational biology}, 9\penalty0 (05):\penalty0 631--645, 2011.

\bibitem[Tibshirani(1996)]{tibshirani1996regression}
R.~Tibshirani.
\newblock Regression shrinkage and selection via the lasso.
\newblock \emph{Journal of the Royal Statistical Society Series B: Statistical Methodology}, 58\penalty0 (1):\penalty0 267--288, 1996.

\bibitem[Turinici(2023)]{SGDconvergence}
G.~Turinici.
\newblock The convergence of the stochastic gradient descent (sgd) : a self-contained proof, 2023.
\newblock URL \url{https://zenodo.org/doi/10.5281/zenodo.4638694}.

\bibitem[Williams(1992)]{williams1992simple}
R.~J. Williams.
\newblock Simple statistical gradient-following algorithms for connectionist reinforcement learning.
\newblock \emph{Machine learning}, 8:\penalty0 229--256, 1992.

\bibitem[Yin and Zhou(2018)]{yin2018arm}
M.~Yin and M.~Zhou.
\newblock Arm: Augment-reinforce-merge gradient for stochastic binary networks.
\newblock \emph{arXiv preprint arXiv:1807.11143}, 2018.

\bibitem[Yin et~al.(2020)Yin, Ho, Yan, Qian, and Zhou]{yin2020probabilistic}
M.~Yin, N.~Ho, B.~Yan, X.~Qian, and M.~Zhou.
\newblock Probabilistic best subset selection via gradient-based optimization.
\newblock \emph{arXiv preprint arXiv:2006.06448}, 2020.

\bibitem[Zhang et~al.(2011)Zhang, Li, Liu, Acar, Rutenbar, and Blanton]{zhang2011virtual}
W.~Zhang, X.~Li, F.~Liu, E.~Acar, R.~A. Rutenbar, and R.~D. Blanton.
\newblock Virtual probe: A statistical framework for low-cost silicon characterization of nanoscale integrated circuits.
\newblock \emph{IEEE Transactions on Computer-Aided Design of Integrated Circuits and Systems}, 30\penalty0 (12):\penalty0 1814--1827, 2011.

\bibitem[Zhang et~al.(2014)Zhang, Wainwright, and Jordan]{zhang2014lowerboundsperformancepolynomialtime}
Y.~Zhang, M.~J. Wainwright, and M.~I. Jordan.
\newblock Lower bounds on the performance of polynomial-time algorithms for sparse linear regression, 2014.
\newblock URL \url{https://arxiv.org/abs/1402.1918}.

\bibitem[Zheng et~al.(2019)Zheng, Askham, Brunton, Kutz, and Aravkin]{brunton2019}
P.~Zheng, T.~Askham, S.~L. Brunton, J.~N. Kutz, and A.~Y. Aravkin.
\newblock A unified framework for sparse relaxed regularized regression: Sr3.
\newblock \emph{IEEE Access}, 7:\penalty0 1404--1423, 2019.
\newblock \doi{10.1109/ACCESS.2018.2886528}.

\end{thebibliography}

\newpage

\appendix



\section{Alternative proof of Theorem 1}
\label{app:alternativeProof}

The following constitutes an alternative proof of Theorem \ref{lem:quadratic}. The proof was provided by \href{https://scholar.google.com/citations?user=FLO79kMAAAAJ}{Janis Keck} after proofreading our article. It has the advantage that the special properties of the product Bernoulli distribution are only used at the very end.

First, a general well known formula is derived.
Let $y$ be a fixed vector, $A$ be a fixed matrix, and $u$ be a random vector. 
Then 
\begin{equation}
    \begin{split}
        \mathbb{E}[ \vert \vert y - A u \vert \vert^2] = \mathbb{E}[y^Ty - 2 y^T A u + u^T A^T A u ].
    \end{split}
\end{equation}
Now, $u^T A^T A u= \text{Trace}(Auu^TA^T)$ and hence 
\begin{equation}
    \begin{split}
        \mathbb{E}[u^T A^T A u] &= \text{Trace}(A~ \mathbb{E}[uu^T]~ A^T) = \text{Trace}(A (\mathrm{Cov}[u] + \mathbb{E}[u] \mathbb{E}[u]^T) A^T) \\
        &= \text{Trace}(A ~\mathrm{Cov}[u]~A^T) 
        + \text{Trace}(A~\mathbb{E}[u] \mathbb{E}[u]^T~A^T).
    \end{split}
\end{equation}
Now the second term can be rewritten again in the original form to be
$ \mathbb{E}[u]^TA^TA \mathbb{E}[u]$.
Thus, plugging this back in, one obtains the general formula
\begin{equation}
    \begin{split}
        \mathbb{E}[ \vert \vert y - A u \vert \vert^2] &= y^Ty - 2 y^T A\mathbb{E}[u] + \mathbb{E}[u]^T A^T A \mathbb{E}[u] + \text{Trace}(A ~\mathrm{Cov}(u) ~A^T) \\
        &= \vert \vert y - A \mathbb{E}[u] \vert \vert^2 + \text{Trace}(A ~\mathrm{Cov}(u) ~A^T).
    \end{split}
\end{equation}
This holds for any choice of random vector $u$. Now, in the special case that the components $u_i$ are independent random variables taking value $w_i$ with probability $\gamma_i$ and zero else, we have $\mathbb{E}[u_i] = \gamma_i w_i$ , and $\mathrm{Cov}[u]_{ii} = \mathbb{E}[u_i^2]- \mathbb{E}[u_i]^2 = \gamma_i w_i^2 - \gamma_i^2 w_i^2 = \gamma_i(1-\gamma_i)w_i^2$.
The other terms of the covariance are zero due to independence. 
Then, writing the trace term out explicitly in indices yields the result. \hfill$\blacksquare$

\section{Refined model selection}
\label{app:refinedModelSelection}

Here we explain how the validation loss selection criterion discussed in Subsection \ref{sec:convToGoodOptima} can be further improved for EGP by emphasizing sparsity in models in the following way. 

Let $(\overline{y},\overline{F})$ be a validation set. Then $\mathcal{L}_k = \sum_i(\overline{y}_i-\sum_j \overline{F}_{ij}w_{jk}\gamma_{jk})^2$ is (an approximation of) the validation loss corresponding to the $k$-th column/model. Furthermore, $s_k = \sum_i \gamma_{ik}$ is the expected value of the $\ell_0$ norm of the $k$-th model. We could evaluate the performance of the $k$-th model by adding these terms. However, we can emphasize the sparsity requirement by instead considering
\begin{equation}
    \begin{split}
        r_k = \alpha s_k + \ln (\mathcal{L}_k),\quad \alpha\in \mathbb{R}.
    \end{split}
    \label{eq:rk}
\end{equation}
The logarithmic term ensures that sparsity is exponentially more important for the evaluation. In practice, we add a small numerical value to $\mathcal{L}_k$ to prevent divergence of $\ln$ and normalize $r_k$ to make subsequent steps less dataset-dependent. Next, we use the $r_k$ scores (lower is better) to combine the various columns/models into an estimate of the true model. To do so, we use an exponential family to transform the $r_k$ scores into normalized weights $c_k$ that we can then use to linearly combine the columns of $w_{ik}$ and $\gamma_{ik}$:
\begin{equation}
    \begin{split}
        c_k := \frac{\exp(-\beta r_k)}{\sum_m \exp(-\beta r_m)},\quad \beta\in \mathbb{R}.
    \end{split}
    \label{eq:ck}
\end{equation}
The $\exp$ in \eqref{eq:ck} pairs well with the $\ln$ in \eqref{eq:rk}.
One nice feature of the probabilistic model is that it can be used to derive a mask that zeros out the components whose $c_k$ coefficients are small:
\begin{equation}
    \begin{split}
        m_i := \text{round}\bigg(\sum_k c_k \gamma_{ik} \bigg).
    \end{split}
    \label{eq:mask}
\end{equation}
We can now use $c_k$ and $m_i$ to estimate the true coefficient/model as follows:
\begin{equation}
    \begin{split}
        \theta_i &= \begin{cases}
            \sum_k c_k \gamma_{ik}w_{ik},&\text{if }m_i=1,\\
            0,&\text{if }m_i=0.
        \end{cases}
    \end{split}
\end{equation}
We found that these estimates were usually superior to the alternative $\theta_i = w_{iK}\gamma_{iK}$, where $K=\text{argmin}_k \mathcal{L}_k$.
Note that this is not possible without a probabilistic mask because for $w$, the rounding operator in \eqref{eq:mask} cannot be expected to yield reasonable results.

\section{Employed hardware}
\label{app:computeresources}

The simulations in Subsection \ref{sec:ComparisonWithMonteCarloMethods} and \ref{sec:TeacherStudentExperiments} were performed on a Linux machine with:
\begin{itemize}
    \item CPU: 2 $\times$ 8-Core Intel Xeon Gold 6144 at 3.5 GHz
    \item RAM: $>$ 100GB (only a fraction of which was used)
\end{itemize}
The simulations in Subsection \ref{sec:compressiveSensingExperiments} were performed on a Linux machine with:
\begin{itemize}
    \item CPU: 2 $\times$ 16-Core Intel Xeon Gold 6226R at 2.9 GHz
    \item GPU: NVidia Tesla A100 80GB 
    \item RAM: $>$ 100GB (only a fraction of which was used)
\end{itemize} 

\section{Comparison with Monte Carlo methods}
\label{app:MonteCarlo}

Here we briefly describe why the comparison in Subsection \ref{sec:ComparisonWithMonteCarloMethods} is already quite conservative.
One reason is that we added EGP Descent to the table to account for the fact that the other Monte Carlo estimators were not implemented with a more advanced optimizer. Another reason is that we removed unnecessary gradient variance and loss estimations in the original Monte Carlo implementations to increase their speed. Furthermore, we restricted each epoch to a single Monte Carlo sample instead of using multiple samples per epoch, which resulted in fastest convergence.

Moreover, there are two ways used in practice to ensure that $\gamma$ always remains an element of $[0,1]^p$ and never leaves this constraint set. The first one is to use a reparameterization, where one sets $\gamma=\sigma(\phi)$, where $\sigma$ is the sigmoid function, and the second is to simply clamp the values of $\gamma$ to $[0,1]^p$ after each gradient step. We compared both methods and found that for the Monte Carlo estimators the clipping variant leads to about ten times faster convergence. We then compared our method with the faster of those variants.

Finally, we did not use many of the additional features that we implemented for EGP in the comparison like parallel hyperparameter tuning, finetuning, validation set convergence, $\ell_1$ regularization, multiple initializations etc. In fact, running EGP with default settings on the $M_2$ problem leads to even lower reconstruction error and lower ASRE, while TUC is only slightly increased.

\section{Systematic compressive sensing experiments}

\begin{figure}[ht]
    \centering
    \begin{subfigure}[t]{0.49\textwidth}
        \includegraphics[width=\linewidth]{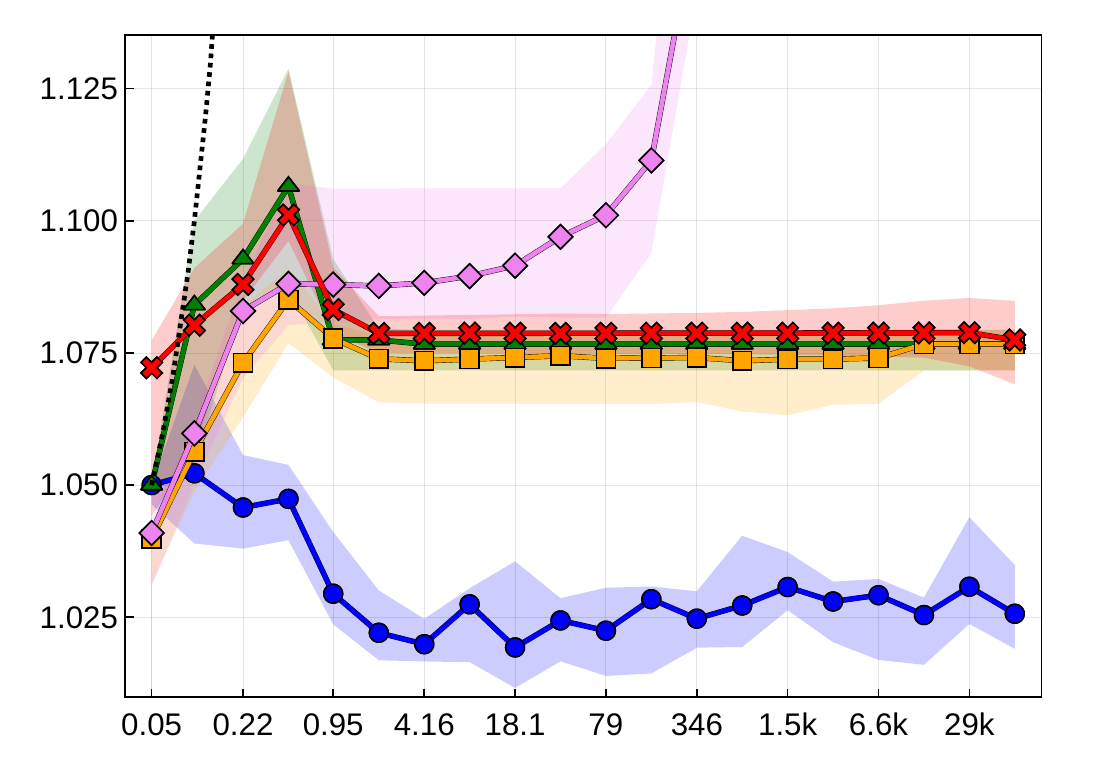}
        \caption{Setting 1}
        \label{fig:RTE:s=1rho=035}
    \end{subfigure}
    \hfill
    \begin{subfigure}[t]{0.49\textwidth}
        \includegraphics[width=\linewidth]{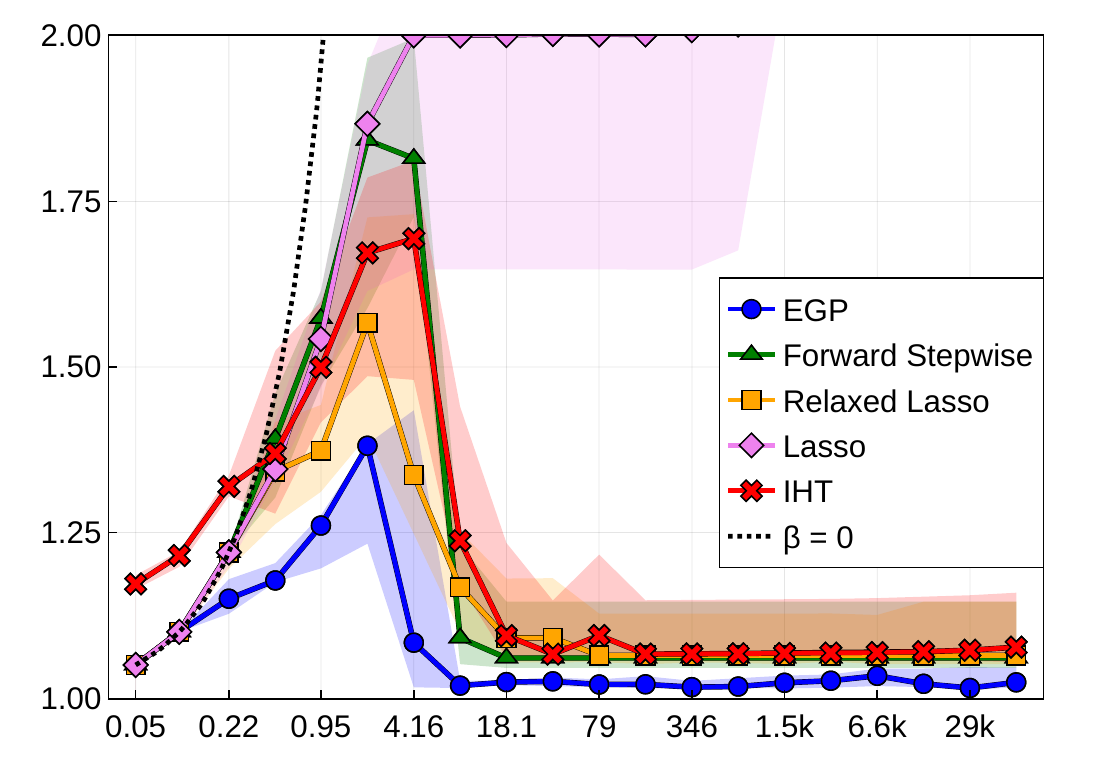}
        \caption{Setting 2}
        \label{fig:RTE:s=2_rho=035}
    \end{subfigure}\\
    \begin{subfigure}[t]{0.49\textwidth}
        \includegraphics[width=\linewidth]{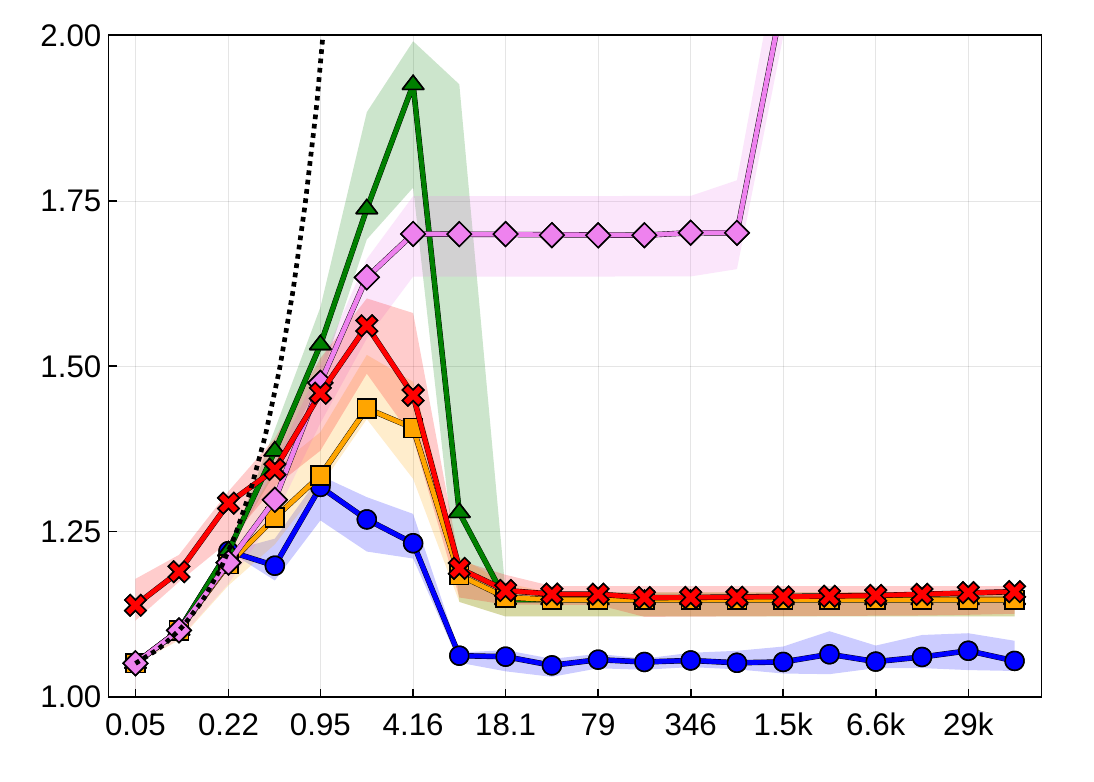}
        \caption{Setting 3}
        \label{fig:RTE:s=3rho=035}
    \end{subfigure}
    \hfill
    \begin{subfigure}[t]{0.49\textwidth}
        \includegraphics[width=\linewidth]{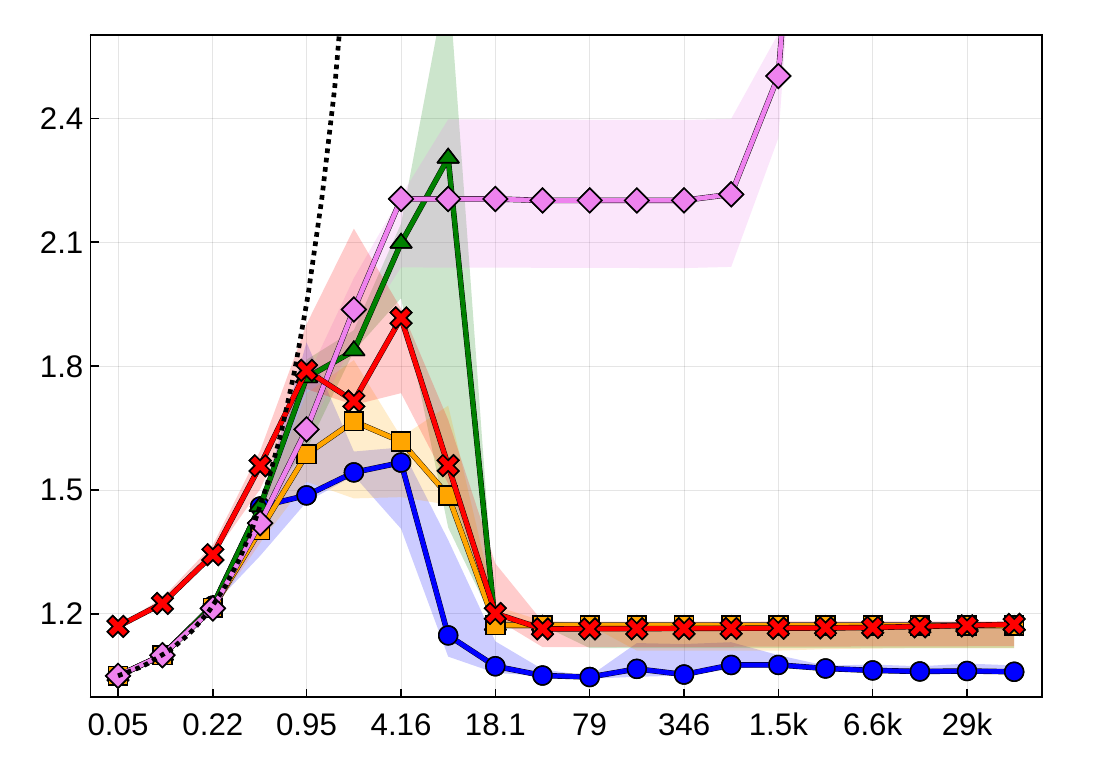}
        \caption{Setting 4}
        \label{fig:RTE:s=4rho=035}
    \end{subfigure}
    \caption{Relative test error (RTE, cf.~\eqref{eq:RTE}) as a function of signal-to-noise ratio (SNR, cf.~Subsection \ref{sec:SNRs}) for the different settings described in \eqref{eq:settings} and (auto)correlation level $\rho=0.35$. Medians over $10$ runs with quantile range between $0.3$ and $0.7$. Legend holds for all subfigures.}
    \label{fig:RTE:rho035}
\end{figure}
\begin{figure}[ht]
    \centering
    \begin{subfigure}[t]{0.49\textwidth}
        \includegraphics[width=\linewidth]{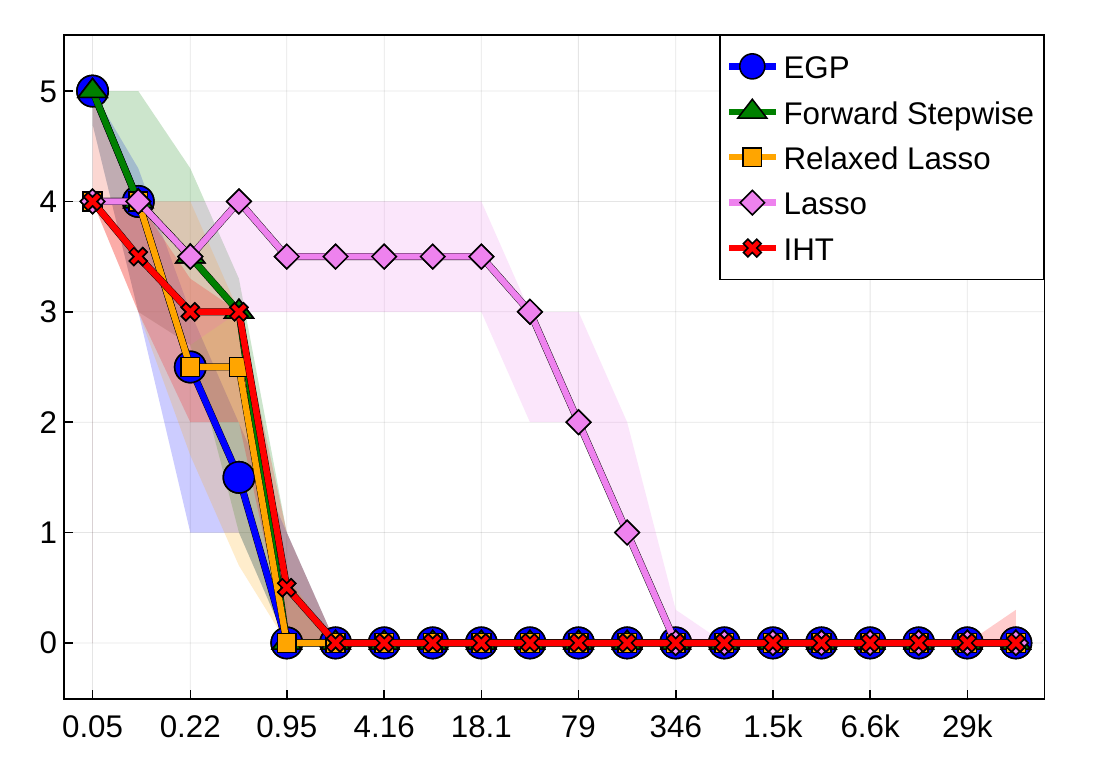}
        \caption{Setting 1}
        \label{fig:ASRE:s=1rho=035}
    \end{subfigure}
    \hfill
    \begin{subfigure}[t]{0.49\textwidth}
        \includegraphics[width=\linewidth]{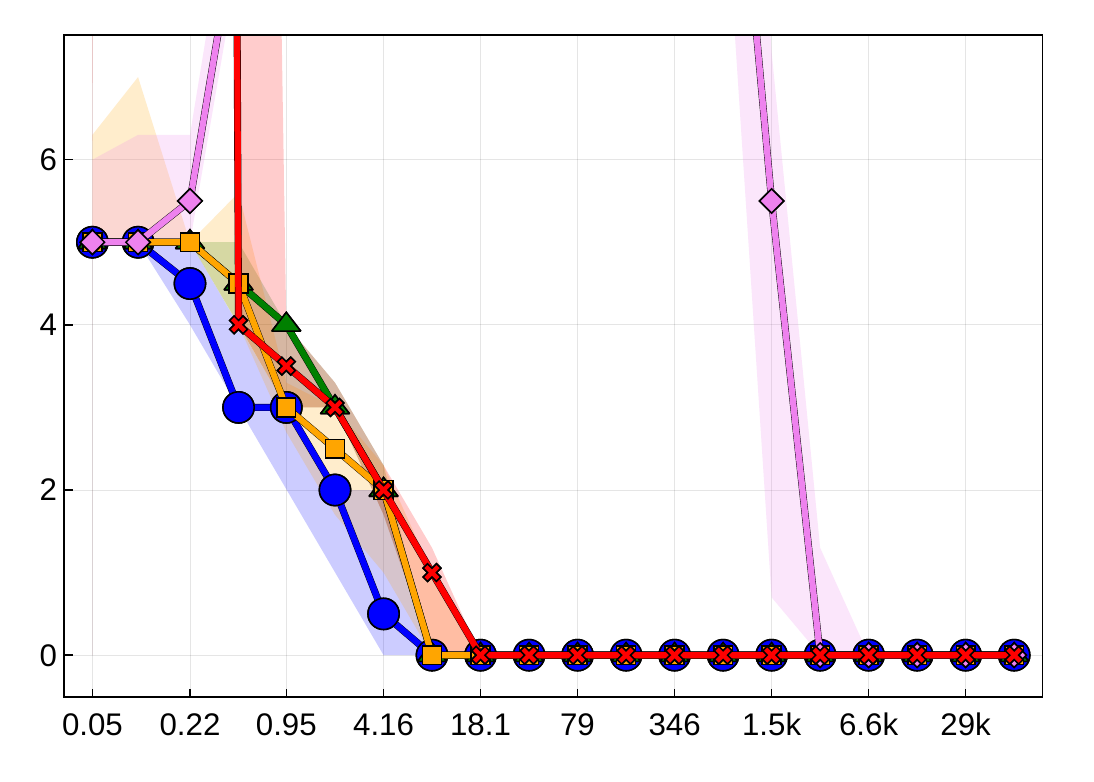}
        \caption{Setting 2}
        \label{fig:ASRE:s=2_rho=035}
    \end{subfigure}\\
    \begin{subfigure}[t]{0.49\textwidth}
        \includegraphics[width=\linewidth]{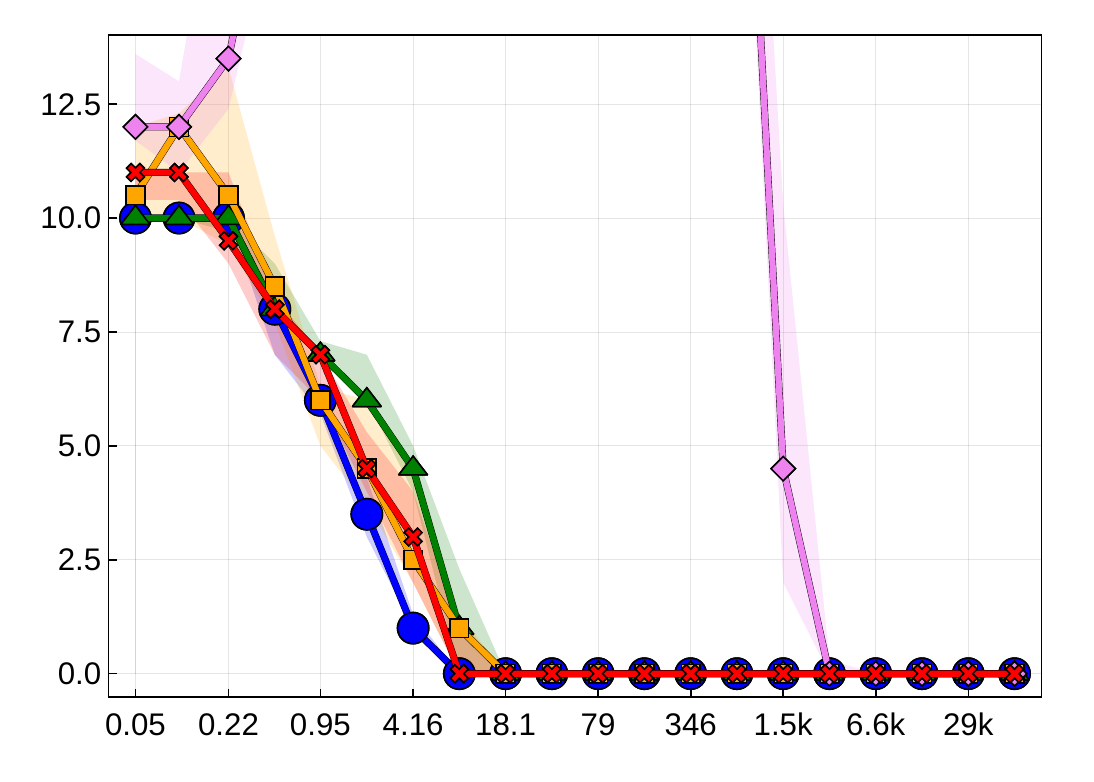}
        \caption{Setting 3}
        \label{fig:ASRE:s=3rho=035}
    \end{subfigure}
    \hfill
    \begin{subfigure}[t]{0.49\textwidth}
        \includegraphics[width=\linewidth]{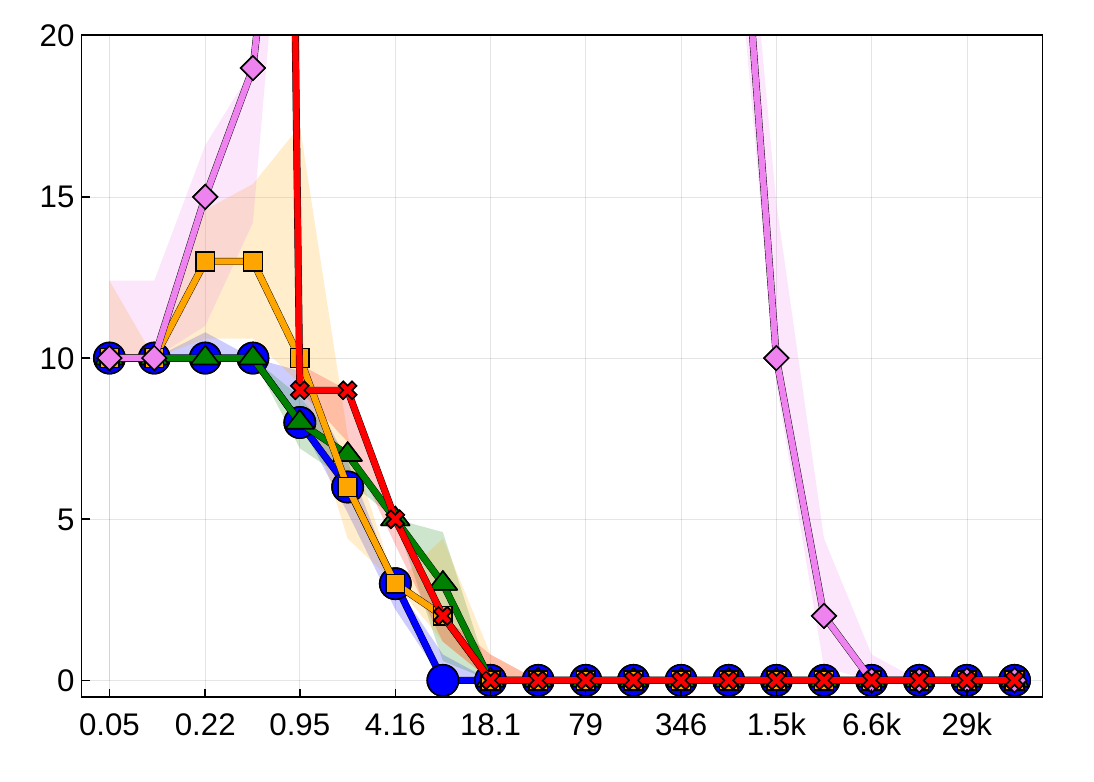}
        \caption{Setting 4}
        \label{fig:ASRE:s=4rho=035}
    \end{subfigure}
    \caption{Active set reconstruction error (ASRE, cf.~\eqref{eq:ASRE}) as a function of signal-to-noise ratio (SNR, cf.~Subsection \ref{sec:SNRs}) for the different settings described in \eqref{eq:settings} and (auto)correlation level $\rho=0.35$. Medians over $10$ runs with quantile range between $0.3$ and $0.7$. Legend holds for all subfigures.}
    \label{fig:ASRE:rho035}
\end{figure}

\begin{figure}[ht]
    \centering
    \begin{subfigure}[t]{0.49\textwidth}
        \includegraphics[width=\linewidth]{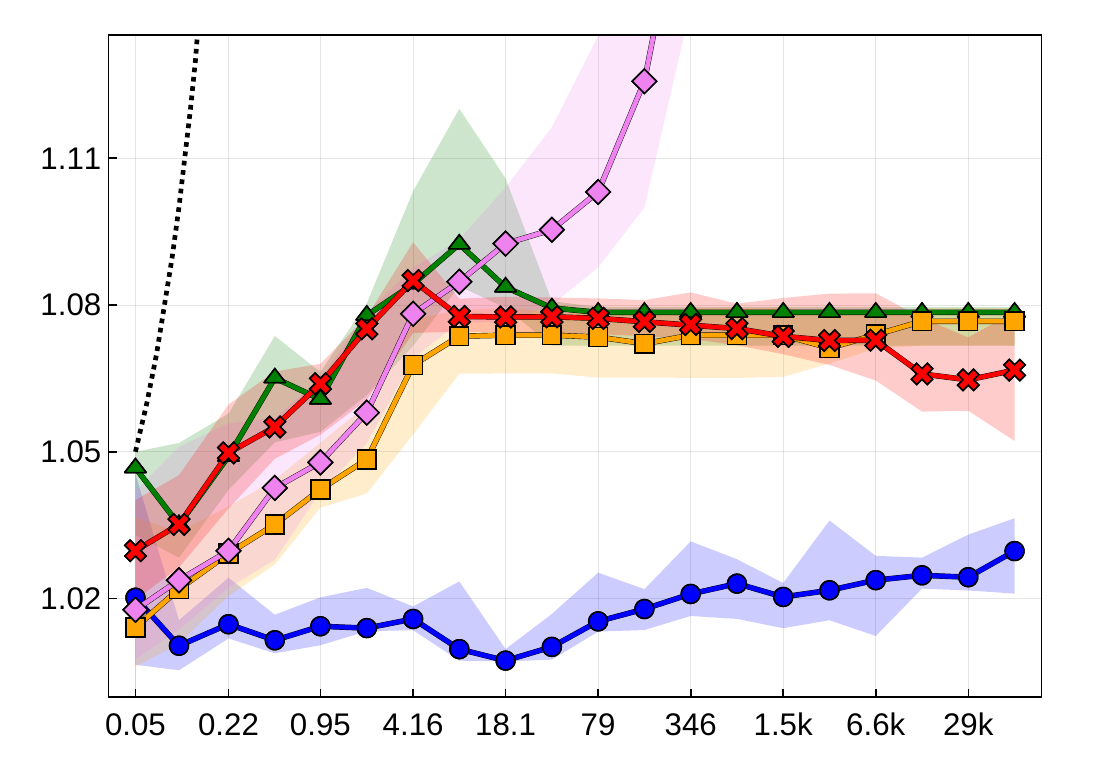}
        \caption{Setting 1}
        \label{fig:RTE:s=1rho=09}
    \end{subfigure}
    \hfill
    \begin{subfigure}[t]{0.49\textwidth}
        \includegraphics[width=\linewidth]{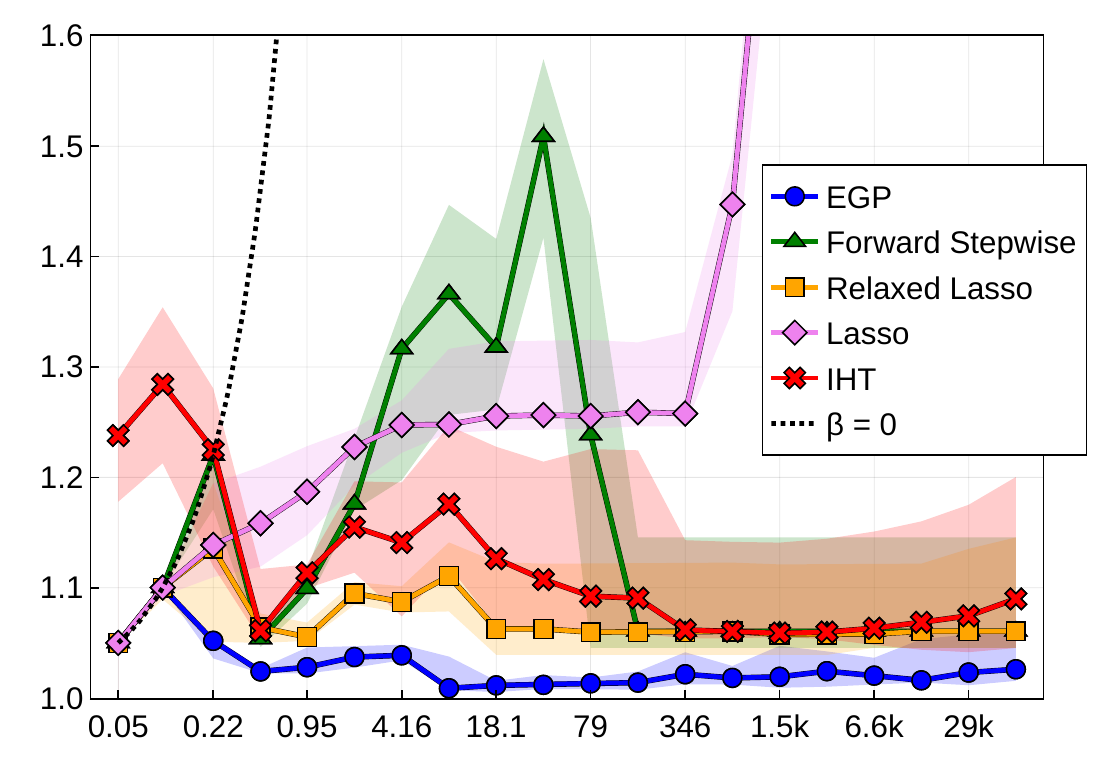}
        \caption{Setting 2}
        \label{fig:RTE:s=2_rho=09}
    \end{subfigure}\\
    \begin{subfigure}[t]{0.49\textwidth}
        \includegraphics[width=\linewidth]{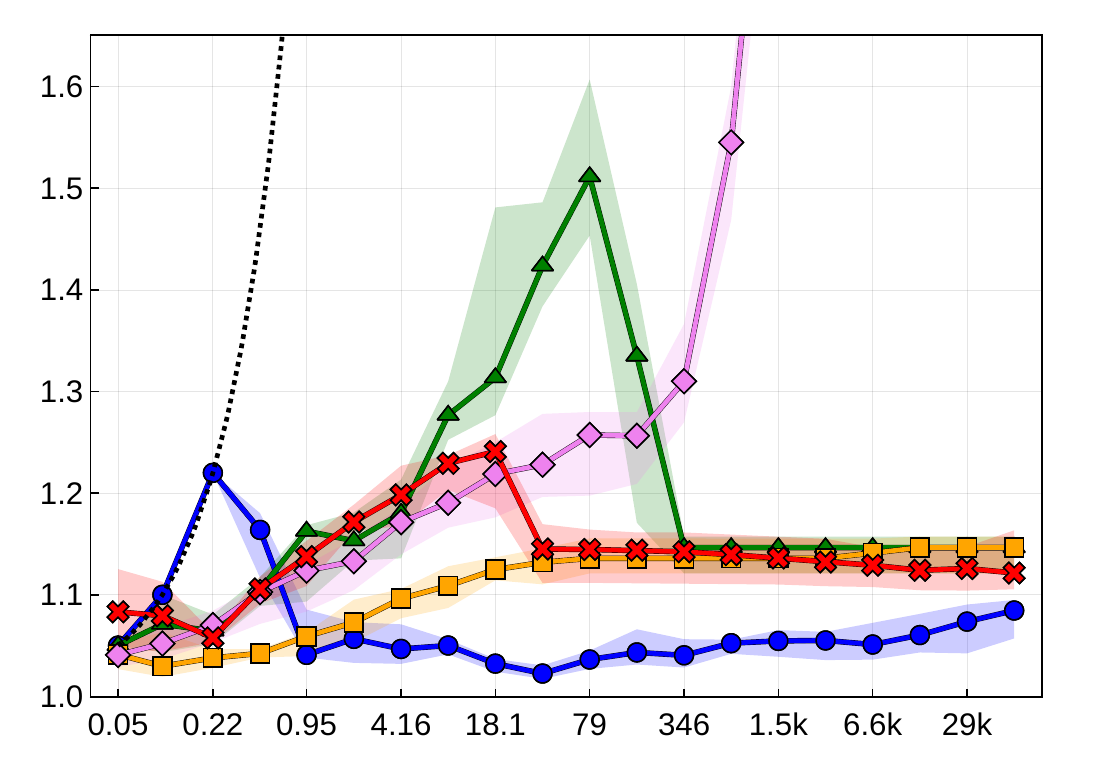}
        \caption{Setting 3}
        \label{fig:RTE:s=3rho=09}
    \end{subfigure}
    \hfill
    \begin{subfigure}[t]{0.49\textwidth}
        \includegraphics[width=\linewidth]{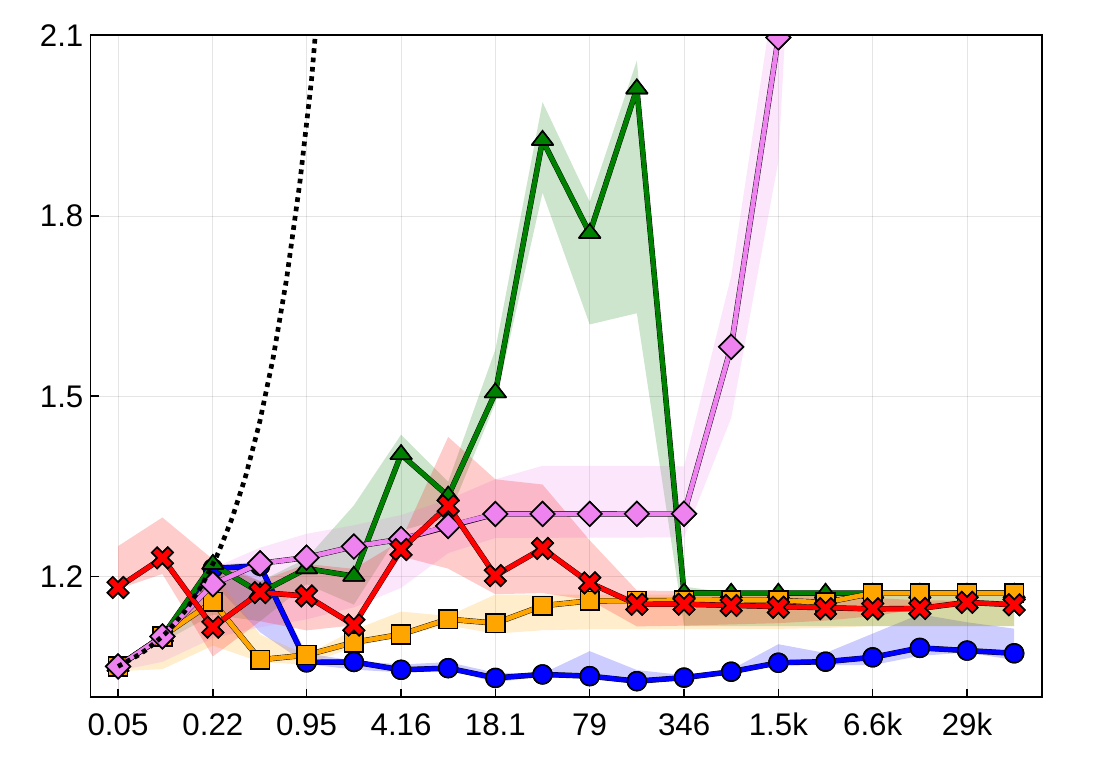}
        \caption{Setting 4}
        \label{fig:RTE:s=4rho=09}
    \end{subfigure}
    \caption{Relative test error (RTE, cf.~\eqref{eq:RTE}) as a function of signal-to-noise ratio (SNR, cf.~Subsection \ref{sec:SNRs}) for the different settings described in \eqref{eq:settings} and (auto)correlation level $\rho=0.9$. Medians over $10$ runs with quantile range between $0.3$ and $0.7$. Legend holds for all subfigures.}
    \label{fig:RTE:rho09}
\end{figure}
\begin{figure}[ht]
    \centering
    \begin{subfigure}[t]{0.49\textwidth}
        \includegraphics[width=\linewidth]{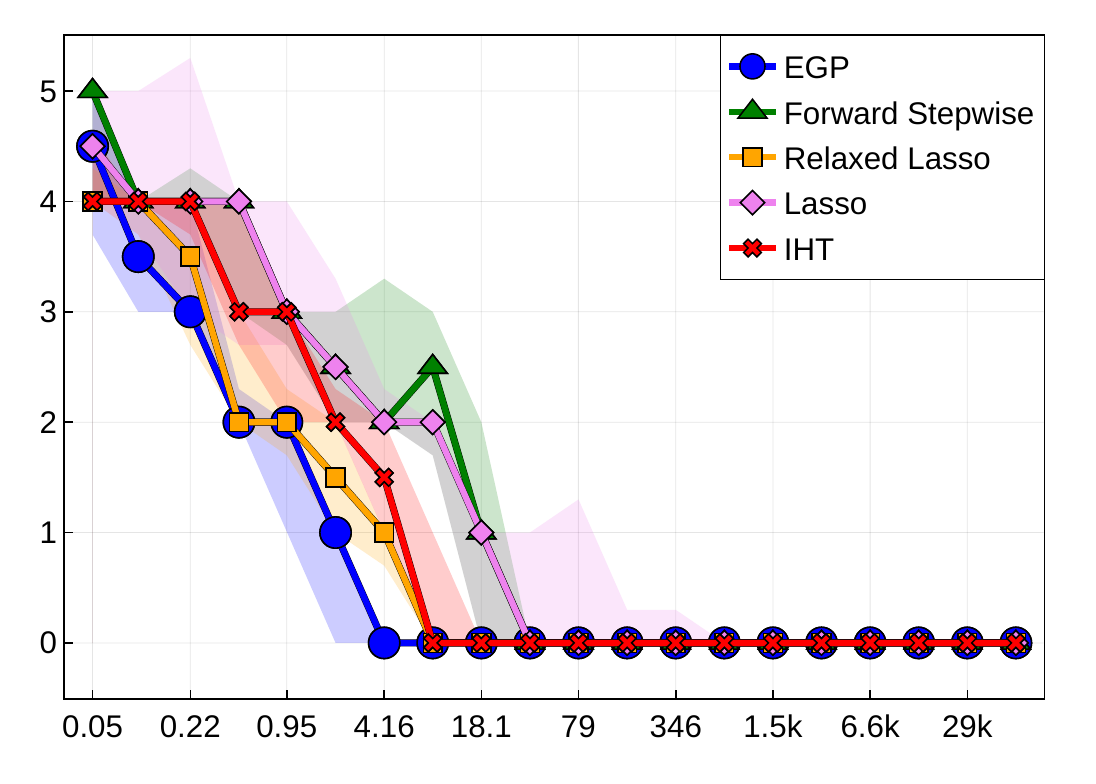}
        \caption{Setting 1}
        \label{fig:ASRE:s=1rho=09}
    \end{subfigure}
    \hfill
    \begin{subfigure}[t]{0.49\textwidth}
        \includegraphics[width=\linewidth]{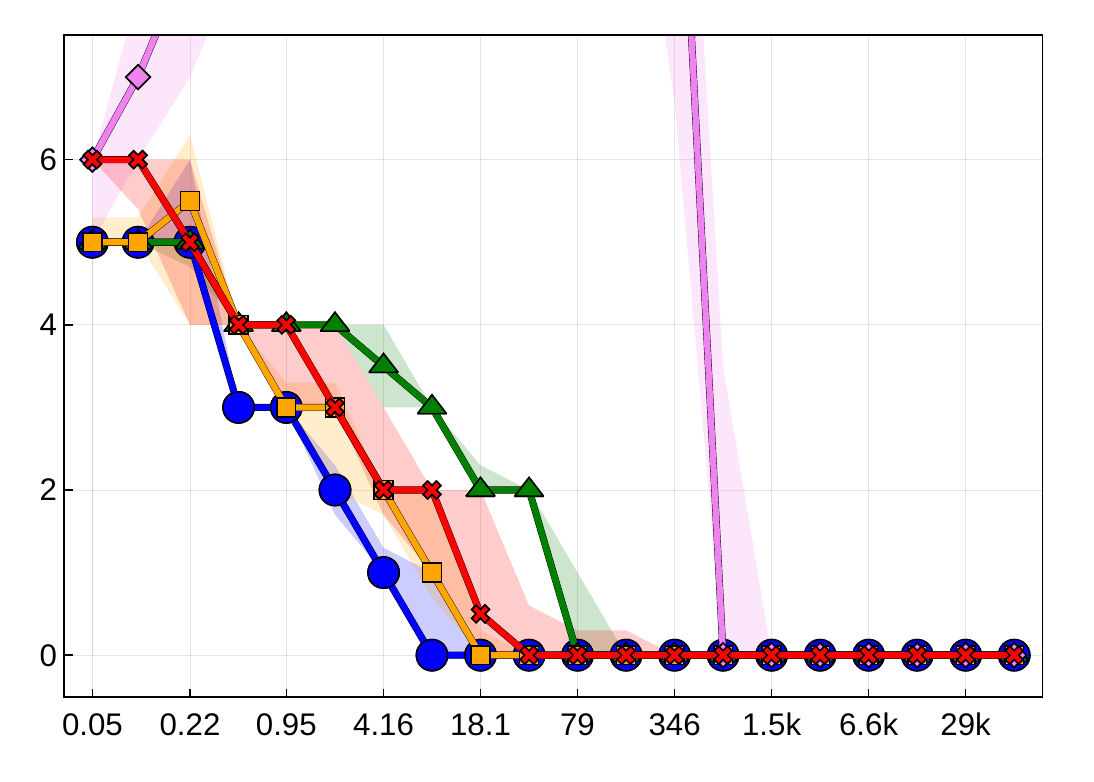}
        \caption{Setting 2}
        \label{fig:ASRE:s=2_rho=09}
    \end{subfigure}\\
    \begin{subfigure}[t]{0.49\textwidth}
        \includegraphics[width=\linewidth]{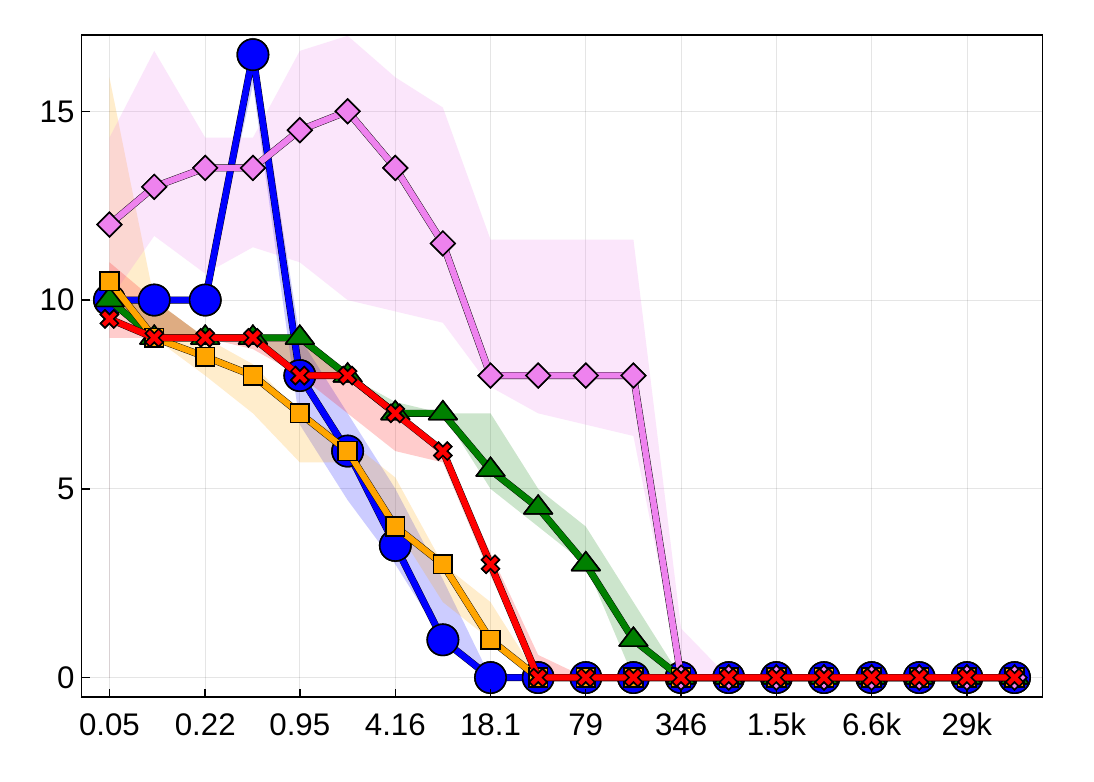}
        \caption{Setting 3}
        \label{fig:ASRE:s=3rho=09}
    \end{subfigure}
    \hfill
    \begin{subfigure}[t]{0.49\textwidth}
        \includegraphics[width=\linewidth]{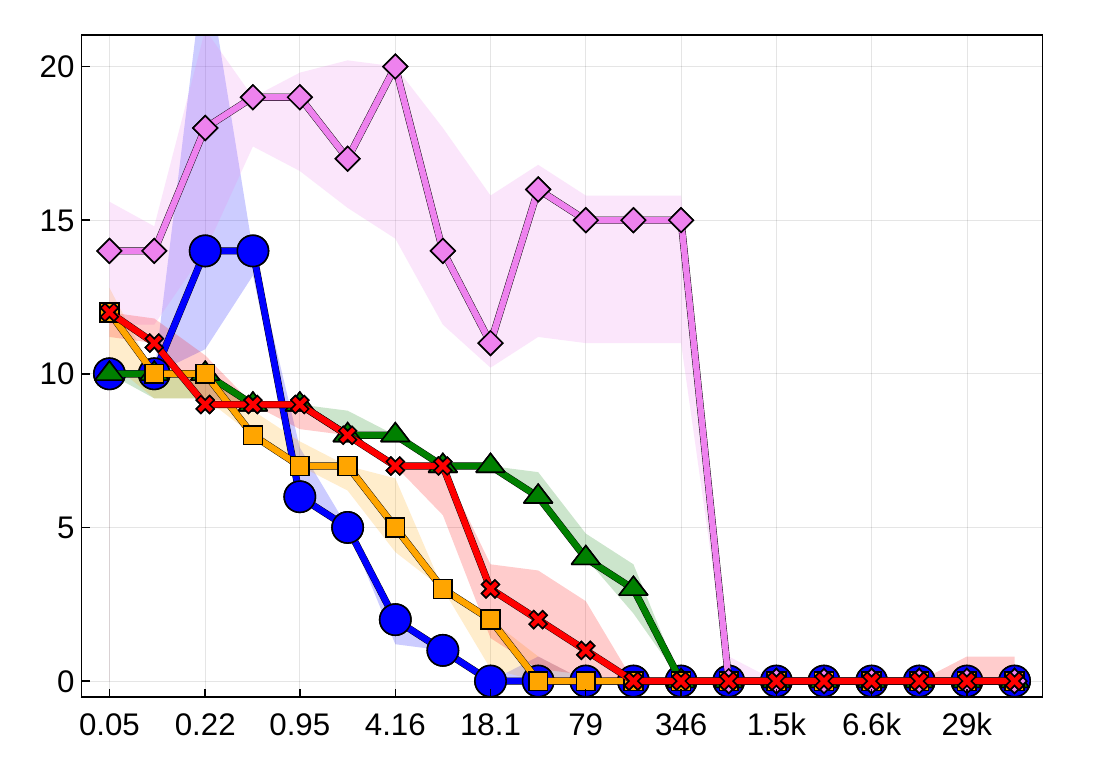}
        \caption{Setting 4}
        \label{fig:ASRE:s=4rho=09}
    \end{subfigure}
    \caption{Active set reconstruction error (ASRE, cf.~\eqref{eq:ASRE}) as a function of signal-to-noise ratio (SNR, cf.~Subsection \ref{sec:SNRs}) for the different settings described in \eqref{eq:settings} and (auto)correlation level $\rho=0.9$. Medians over $10$ runs with quantile range between $0.3$ and $0.7$. Legend holds for all subfigures.}
    \label{fig:ASRE:rho09}
\end{figure}

\subsection{Coefficient selection}
\label{app:coefficientSelection}

Here we describe the details of model selection for the simulations presented in Subsection \ref{sec:compressiveSensingExperiments}.
In every simulations, we run each method many times with the different hyperparameter values specified below and select the final coefficient by comparing the validation losses or employing the procedure described in Subsection \ref{sec:convToGoodOptima}. Our validation set size is equal to $500$ across all experiments.
\begin{enumerate}
    \item We run Relaxed Lasso with $10$ $\lambda$ values and $50$ relax values, amounting to $500$ runs per simulation (meaning that across all simulations, Relaxed Lasso is in total run 160,000 times). We set the intercept to false (no addition of bias vector since the training data has zero mean).
    \item Since plain Lasso does not have a relax value, we correspondingly run it with $500$ different $\lambda$ values per simulations. Also here, we set the intercept to false.
    \item Forward Stepwise is initialized with the zero coefficient and then, at each step, adds the next best non-zero component to the coefficient. At each of the $p$ iterations, the corresponding coefficient is stored. We then determine the coefficient with lowest validation loss among all $p$ coefficients.
    \item We run IHT $150$ times, one time for each $k$ value (that means, finding the best coefficient with the constraint that at most $k$ values are non-zero for all $0\le k\le 150$). Since none of our experiments have more than $150$ non-zero coefficients, this is in fact very generous to IHT and gives more than enough opportunities to find the best possible configuration.
    \item EGP runs several hyperparameter optimizations in parallel by promoting $w,\gamma$ to matrices as described in Subsection \ref{sec:convToGoodOptima}. It varies over $5$ different $\lambda_{1,k}$ values, $3$ different learning rates and $33$ $\lambda_{0,k}$ values, while the initial coefficients before training are all randomly initialized, giving rise to $5\cdot 3\cdot 33=495$ parallel runs for each simulation.
\end{enumerate}

Several other hyperparameters are internal to the methods above but we only discuss those that we changed. For EGP, 
we provide a more detailed list of further hyperparameters in Appendix \ref{app:EGPhyperparameters}. We remark that all these hyperparameter values were held fixed for all the results described in Subsection \ref{sec:EGPresults}, demonstrating some robustness to hyperparameter choice.

\subsection{EGP hyperparameters}
\label{app:EGPhyperparameters}

Next to learning rate and $\lambda$ values, EGP has several other parameters, the most important ones of which we briefly mention below.
All of them were held fixed throughout all simulations described in Section \ref{sec:compressiveSensingExperiments}.

\begin{itemize}
    \item \texttt{lambda\_range}: Range of $\lambda$'s for which problem \eqref{eq:EGP} is solved in parallel.
    \item \texttt{lr\_range}: Range of learning rates for which problem \eqref{eq:EGP} is solved in parallel.
    \item \texttt{batch\_size}: Batch size for SGD solver.
    \item \texttt{optimizer}: Optimizer for SGD solver. Default is Adam.
    \item \texttt{initial\_p\_value}: Initial value array of what we call $\gamma$ in this article. Default is simply equal to $1$ for all entries.
    \item \texttt{min\_epochs, max\_epochs}: Min / max epochs for SGD solver. 
    \item \texttt{weight\_factor\_trank}: Factor that corresponds to $\alpha$ in \eqref{eq:rk}.
    \item \texttt{weight\_factor}: Factor that corresponds to $\beta$ in \eqref{eq:ck}.
    \item \texttt{convergence\_check\_interval}: Interval (in epochs) at which convergence is checked.
    \item \texttt{min\_possible\_deviation}: Corresponds to the deviation parameter $\Delta$ used for convergence as described in Subsection \ref{sec:convergence}.
    \item \texttt{mask\_stability\_factor}: Corresponds to $\epsilon$ used for convergence as described in Subsection \ref{sec:convergence}.
    \item \texttt{l1\_regularization}: Whether $\ell_1$ regularization is employed in addition to $\ell_0$ regularization or not.
    \item \texttt{l1\_range}: If the above parameter is set to \texttt{true}, then this specifies the range of $\lambda_{1,k}$ values in \eqref{eq:L0andL1}.
    \item \texttt{finetuning}: If set to \texttt{true}, then finetuning is performed (which simply optimizes the least squares loss without penalization, amounting to a form of relaxed regularization). Separate hyperparameters (like learning rate range and min / max epochs) can be adjusted for the finetuning optimization in this case.
\end{itemize}

\subsection{Further compressive sensing comparison results}
\label{app:furtherCompressiveSensingExperiments}

Here, we present further numerical results. In addition to the $\rho$ values presented in Section \ref{sec:compressiveSensingExperiments}, we show results for all settings for $\rho=0.35$ in Figure \ref{fig:RTE:rho035} (RTE) and Figure \ref{fig:ASRE:rho035} (ASRE) and for $\rho=0.9$ in Figure \ref{fig:RTE:rho09} (RTE) and Figure \ref{fig:ASRE:rho09} (ASRE). The results largely confirm the trends described in Section \ref{sec:compressiveSensingExperiments}.

\subsection{Ablation study - EGP without \texorpdfstring{$\ell_1$}{L1}-regularization}
\label{app:ablationEGP}

Since EGP by default employs a combination of $\ell_0$  and $\ell_1$ regularization (cf. Subsection \ref{sec:L1L2combinationsEGP}), we perform an ablation study with simulations that only use $\ell_0$ regularization. To this end, we set the hyperparameter \texttt{l1\_regularization} to \texttt{false} (cf.~Appendix \ref{app:EGPhyperparameters}) and instead perform parallel optimization over more $\lambda_{0,k}$ values, while keeping the total number of parallel runs $\le 500$ (in accordance with the coefficient selection strategy described in Appendix \ref{app:coefficientSelection}).

\begin{figure}[ht]
    \centering
    \begin{subfigure}[t]{0.45\textwidth}
        \includegraphics[width=\linewidth]{figures/EGP/RTE/RTE_s=1_rho=035_noLegend.pdf}
    \end{subfigure}
    \begin{subfigure}[t]{0.45\textwidth}
        \includegraphics[width=\linewidth]{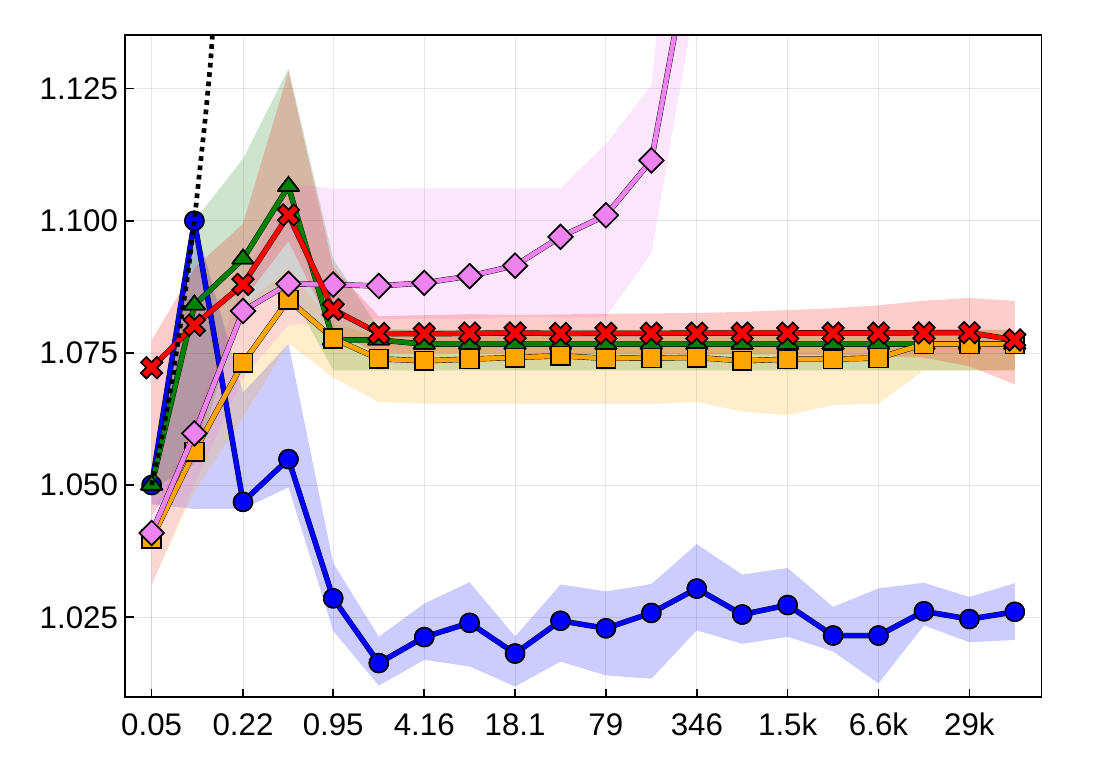}
    \end{subfigure}\\
    \begin{subfigure}[t]{0.45\textwidth}
        \includegraphics[width=\linewidth]{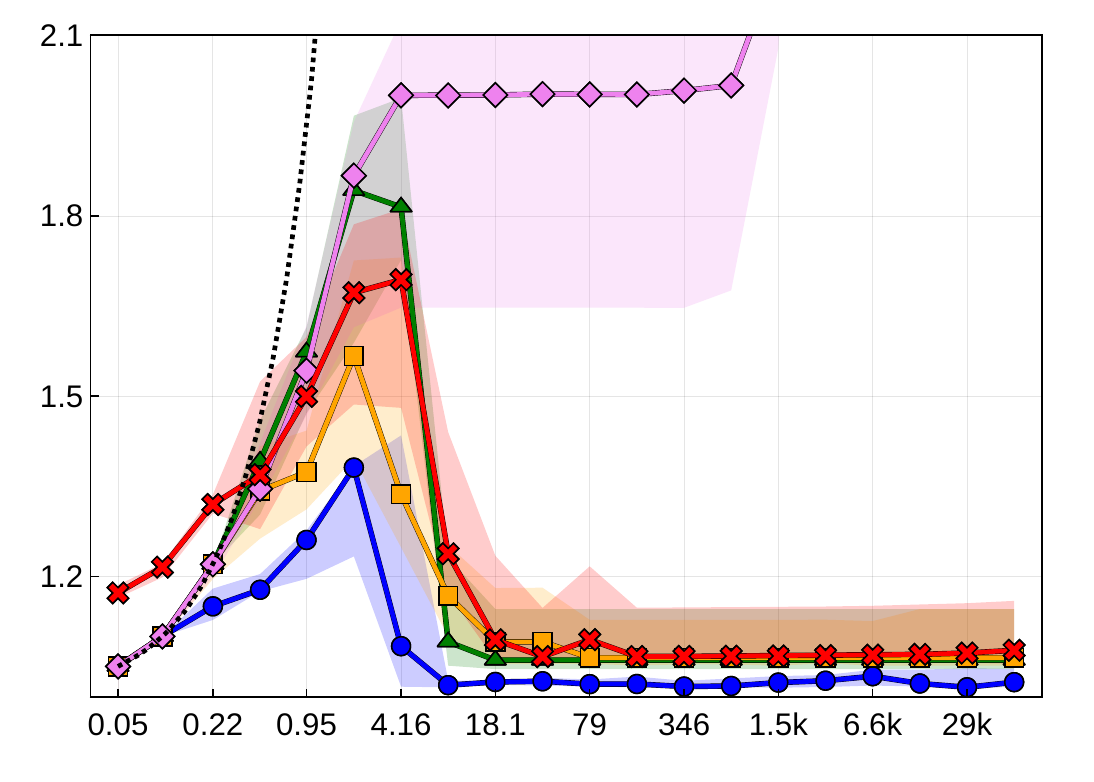}
    \end{subfigure}
    \begin{subfigure}[t]{0.45\textwidth}
        \includegraphics[width=\linewidth]{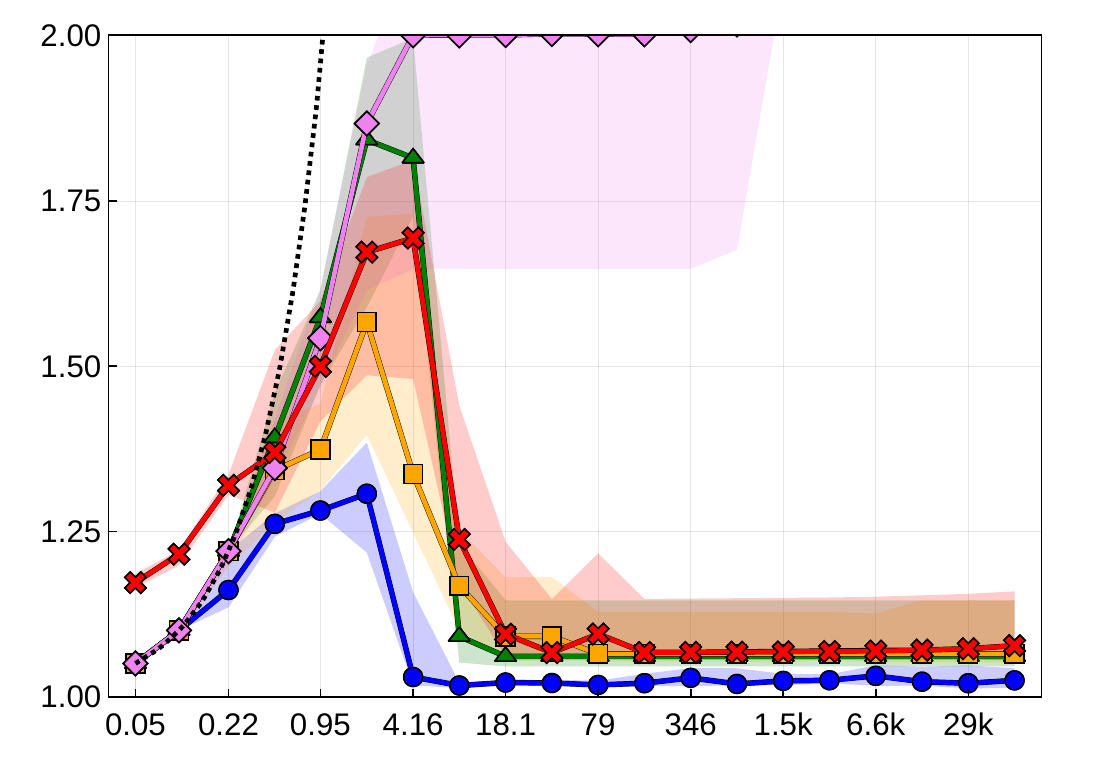}
    \end{subfigure}\\
    \begin{subfigure}[t]{0.45\textwidth}
        \includegraphics[width=\linewidth]{figures/EGP/RTE/RTE_s=3_rho=035_noLegend.pdf}
    \end{subfigure}
    \begin{subfigure}[t]{0.45\textwidth}
        \includegraphics[width=\linewidth]{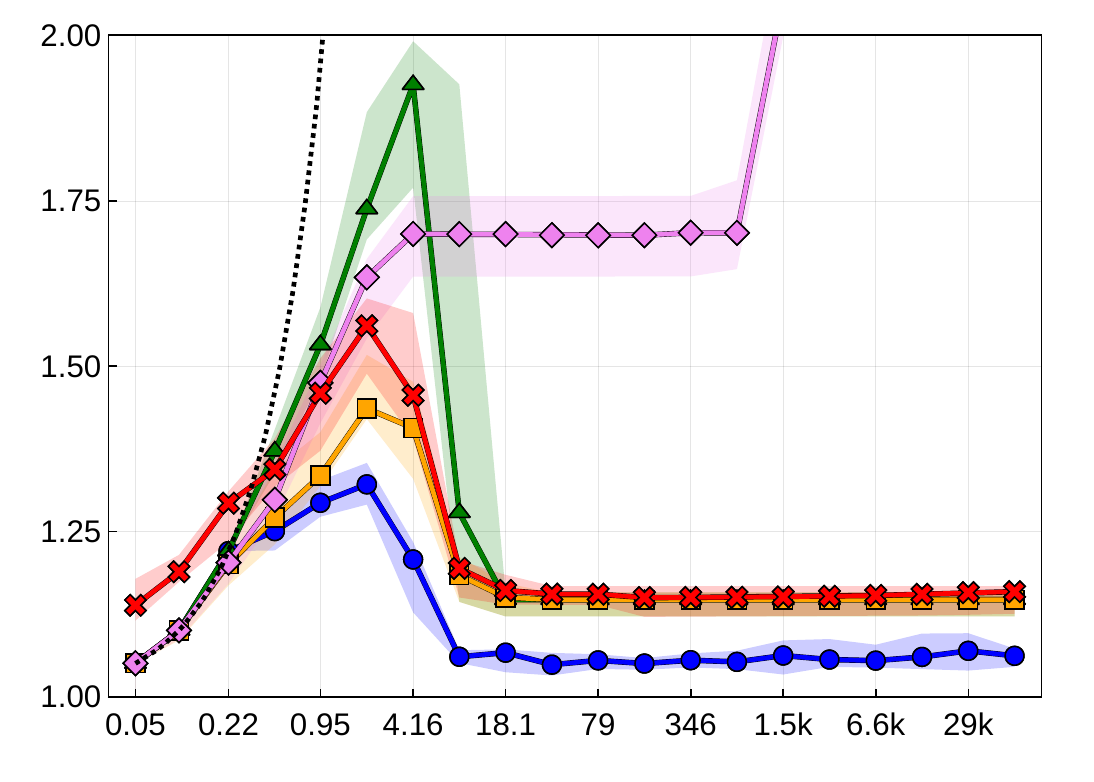}
    \end{subfigure}\\
    \begin{subfigure}[t]{0.45\textwidth}
        \includegraphics[width=\linewidth]{figures/EGP/RTE/RTE_s=4_rho=035_noLegend.pdf}
    \end{subfigure}
    \begin{subfigure}[t]{0.45\textwidth}
        \includegraphics[width=\linewidth]{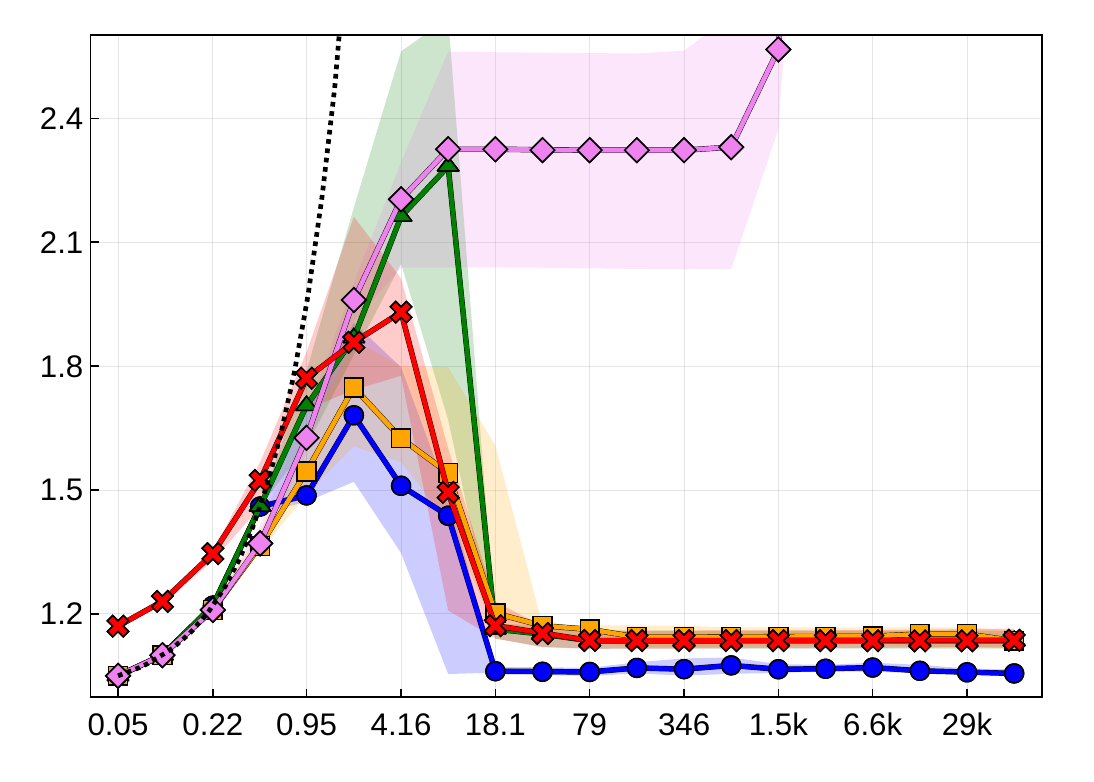}
    \end{subfigure}
    \caption{RTE as a function of SNR. The 4 rows correspond to the 4 settings described in \eqref{eq:settings}. (Auto)correlation level $\rho=0.35$. Left column: EGP with $\ell_1$ regularization. Right column: EGP without $\ell_1$ regularization. Medians over $10$ runs with quantile range between $0.3$ and $0.7$. Legend holds for all subfigures.}
    \label{fig:L1_ablation_2}
\end{figure}

\begin{figure}[ht]
    \centering
    \begin{subfigure}[t]{0.45\textwidth}
        \includegraphics[width=\linewidth]{figures/EGP/RTE/RTE_s=1_rho=07_noLegend.pdf}
    \end{subfigure}
    \begin{subfigure}[t]{0.45\textwidth}
        \includegraphics[width=\linewidth]{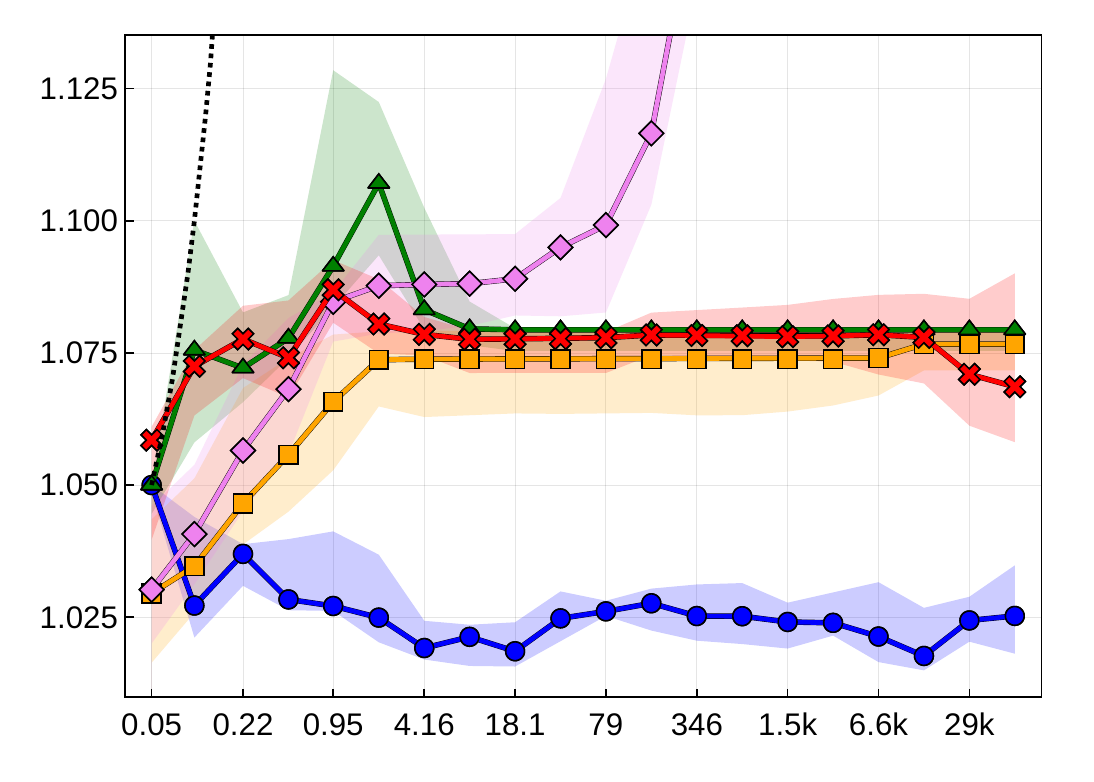}
    \end{subfigure}\\
    \begin{subfigure}[t]{0.45\textwidth}
        \includegraphics[width=\linewidth]{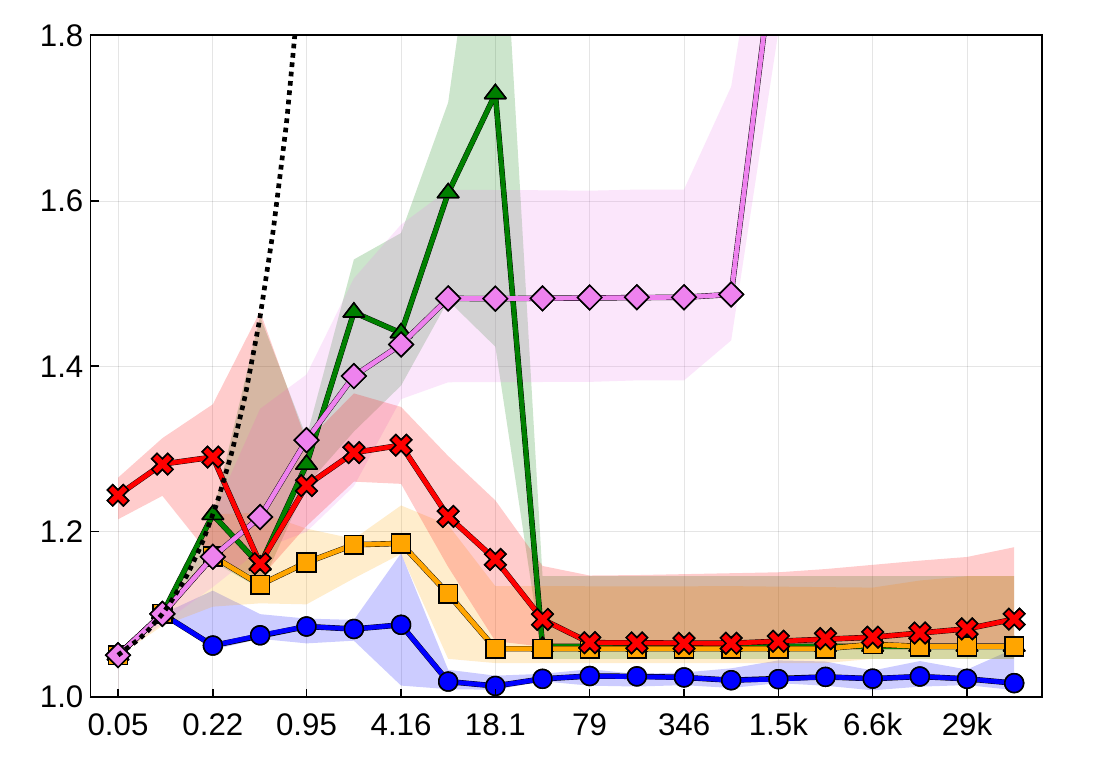}
    \end{subfigure}
    \begin{subfigure}[t]{0.45\textwidth}
        \includegraphics[width=\linewidth]{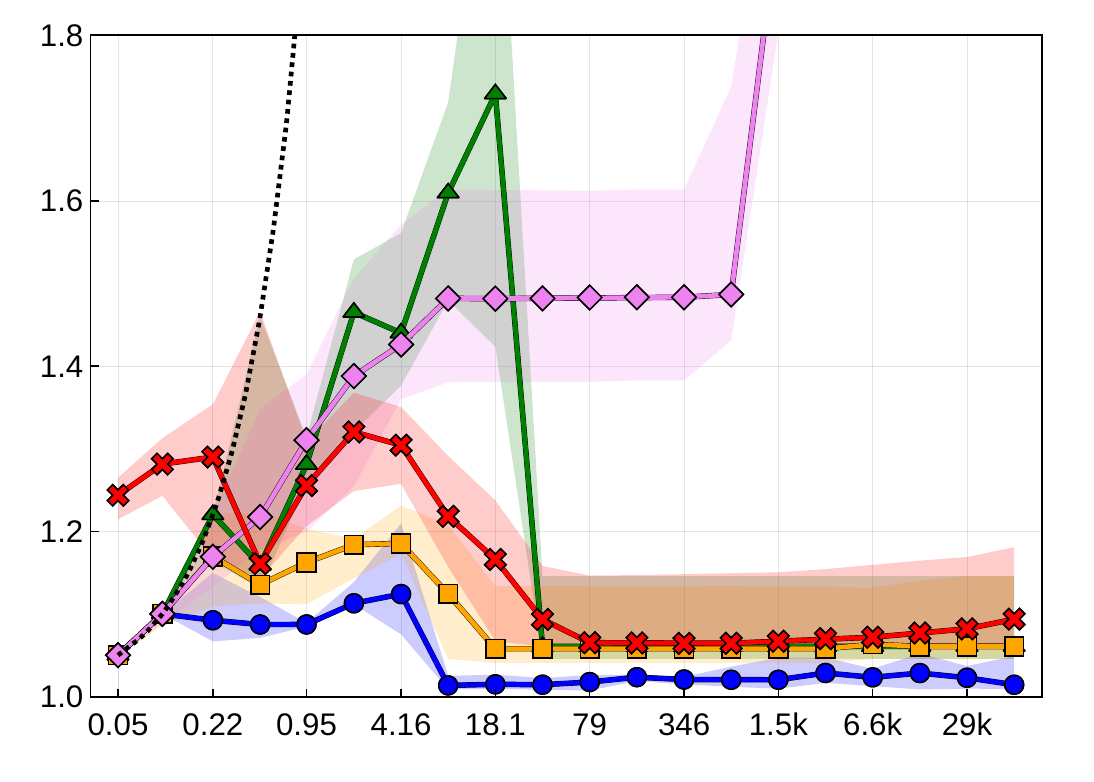}
    \end{subfigure}\\
    \begin{subfigure}[t]{0.45\textwidth}
        \includegraphics[width=\linewidth]{figures/EGP/RTE/RTE_s=3_rho=07_noLegend.pdf}
    \end{subfigure}
    \begin{subfigure}[t]{0.45\textwidth}
        \includegraphics[width=\linewidth]{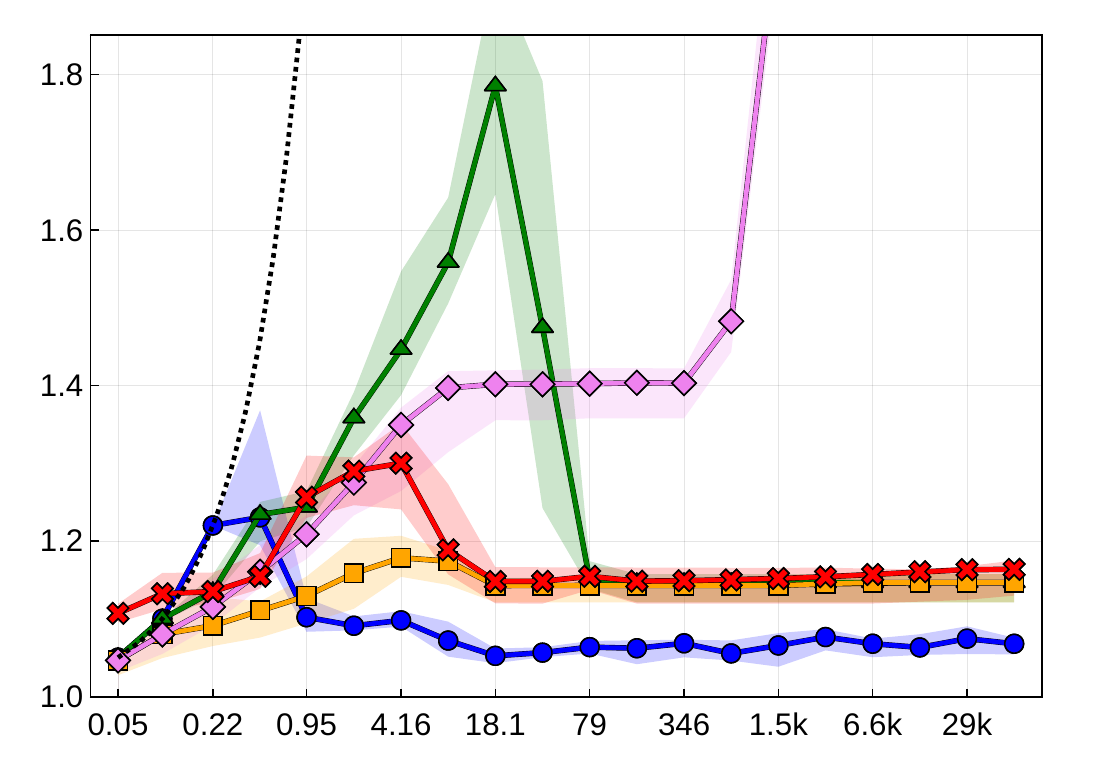}
    \end{subfigure}\\
    \begin{subfigure}[t]{0.45\textwidth}
        \includegraphics[width=\linewidth]{figures/EGP/RTE/RTE_s=4_rho=07_noLegend.pdf}
    \end{subfigure}
    \begin{subfigure}[t]{0.45\textwidth}
        \includegraphics[width=\linewidth]{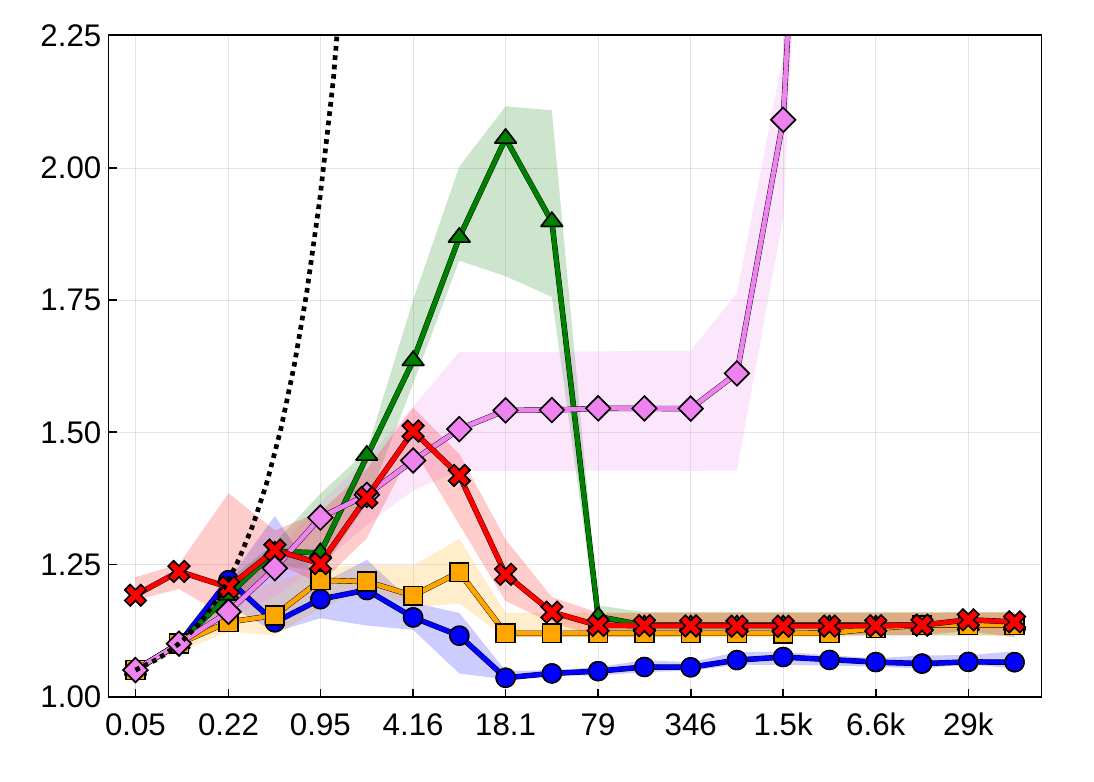}
    \end{subfigure}
    \caption{RTE as a function of SNR. The 4 rows correspond to the 4 settings described in \eqref{eq:settings}. (Auto)correlation level $\rho=0.7$. Left column: EGP with $\ell_1$ regularization. Right column: EGP without $\ell_1$ regularization. Medians over $10$ runs with quantile range between $0.3$ and $0.7$. Legend holds for all subfigures.}
    \label{fig:L1_ablation_3}
\end{figure}

\begin{figure}[ht]
    \centering
    \begin{subfigure}[t]{0.45\textwidth}
        \includegraphics[width=\linewidth]{figures/EGP/RTE/RTE_s=1_rho=09_noLegend.pdf}
    \end{subfigure}
    \begin{subfigure}[t]{0.45\textwidth}
        \includegraphics[width=\linewidth]{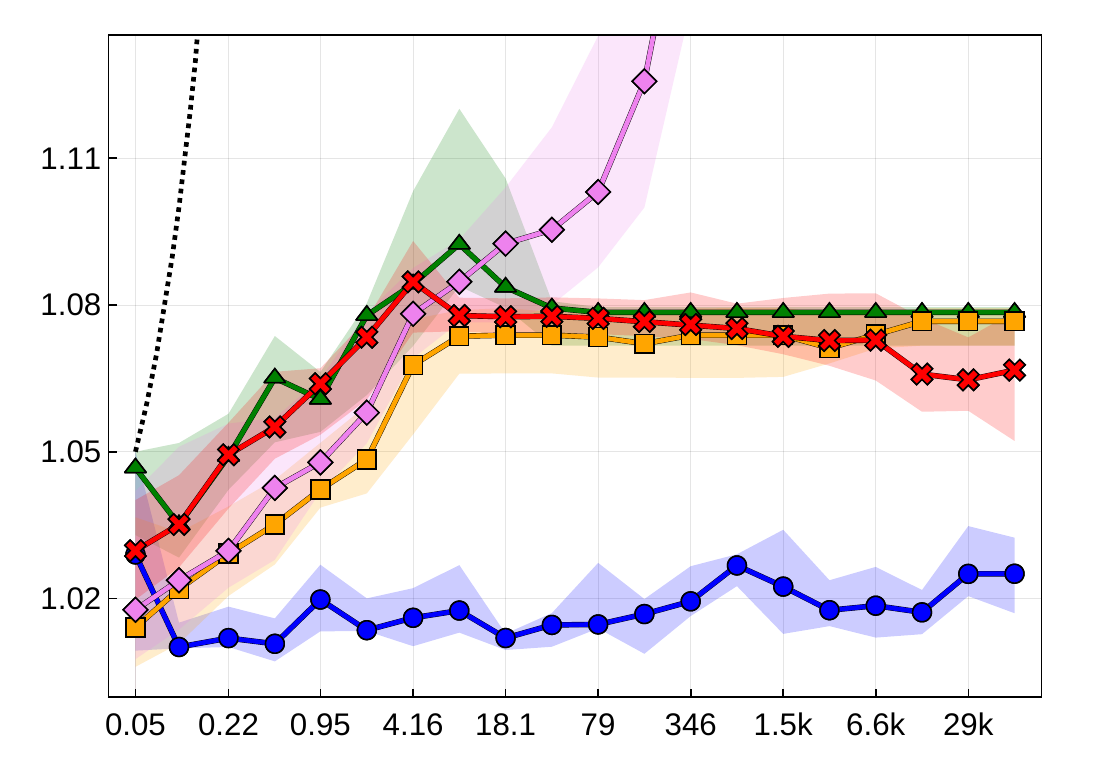}
    \end{subfigure}\\
    \begin{subfigure}[t]{0.45\textwidth}
        \includegraphics[width=\linewidth]{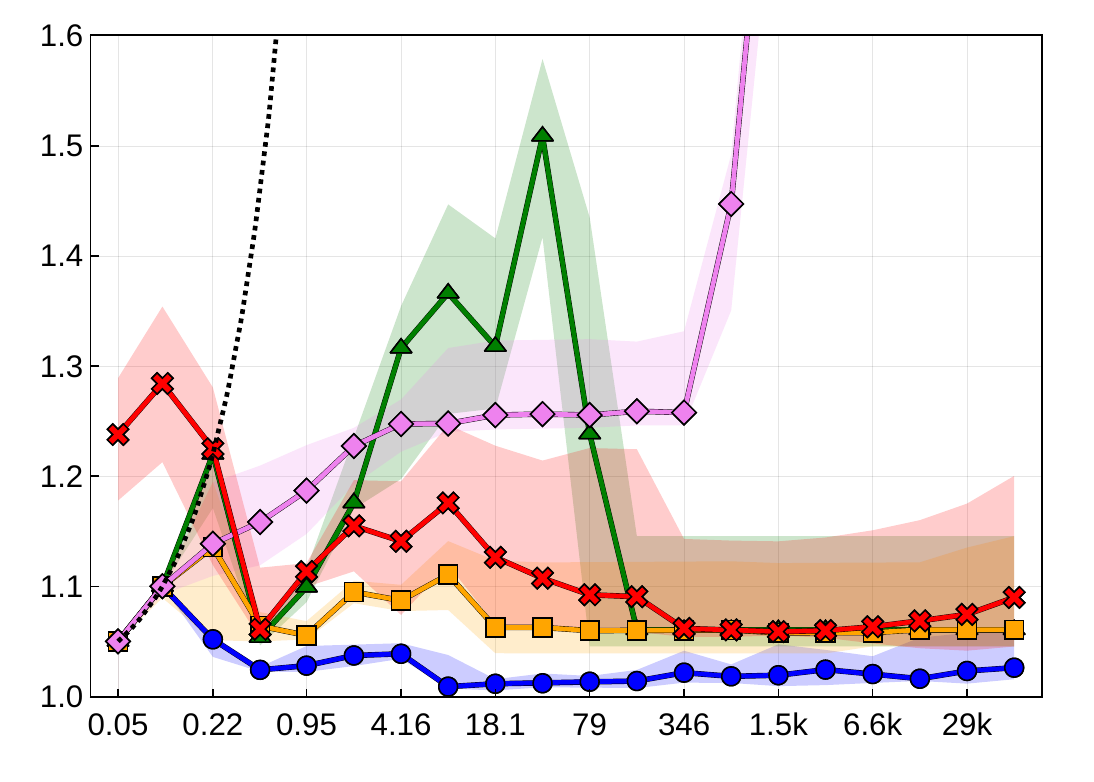}
    \end{subfigure}
    \begin{subfigure}[t]{0.45\textwidth}
        \includegraphics[width=\linewidth]{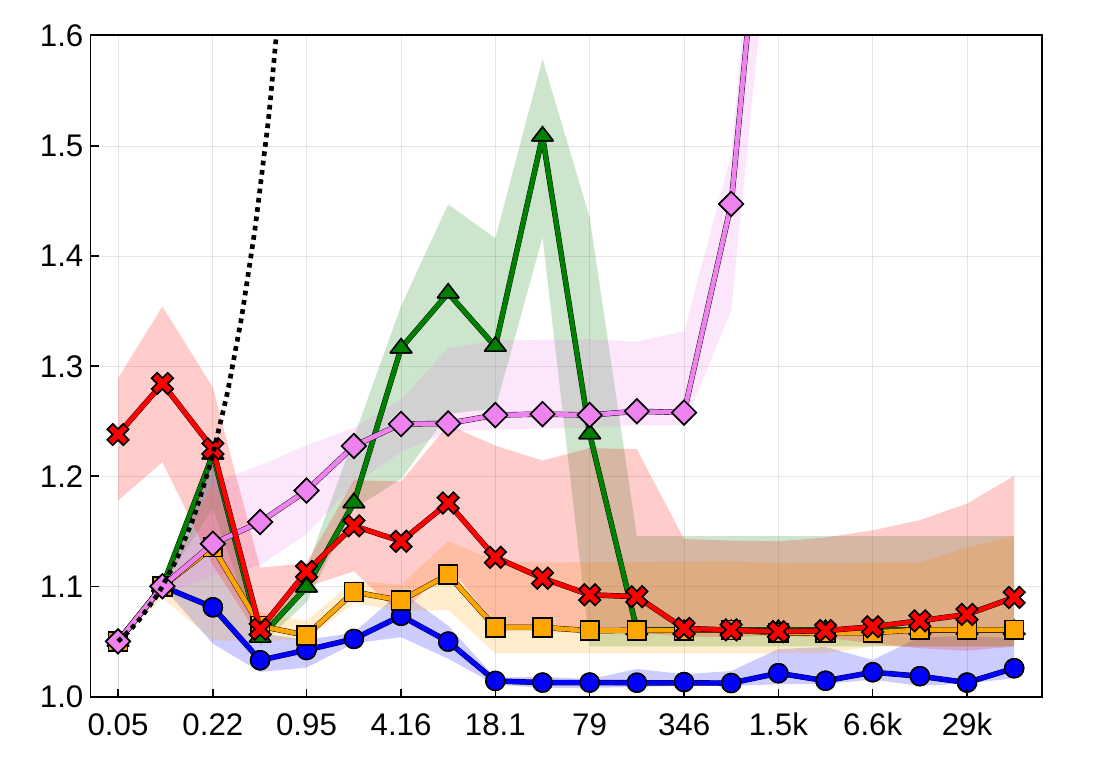}
    \end{subfigure}\\
    \begin{subfigure}[t]{0.45\textwidth}
        \includegraphics[width=\linewidth]{figures/EGP/RTE/RTE_s=3_rho=09_noLegend.pdf}
    \end{subfigure}
    \begin{subfigure}[t]{0.45\textwidth}
        \includegraphics[width=\linewidth]{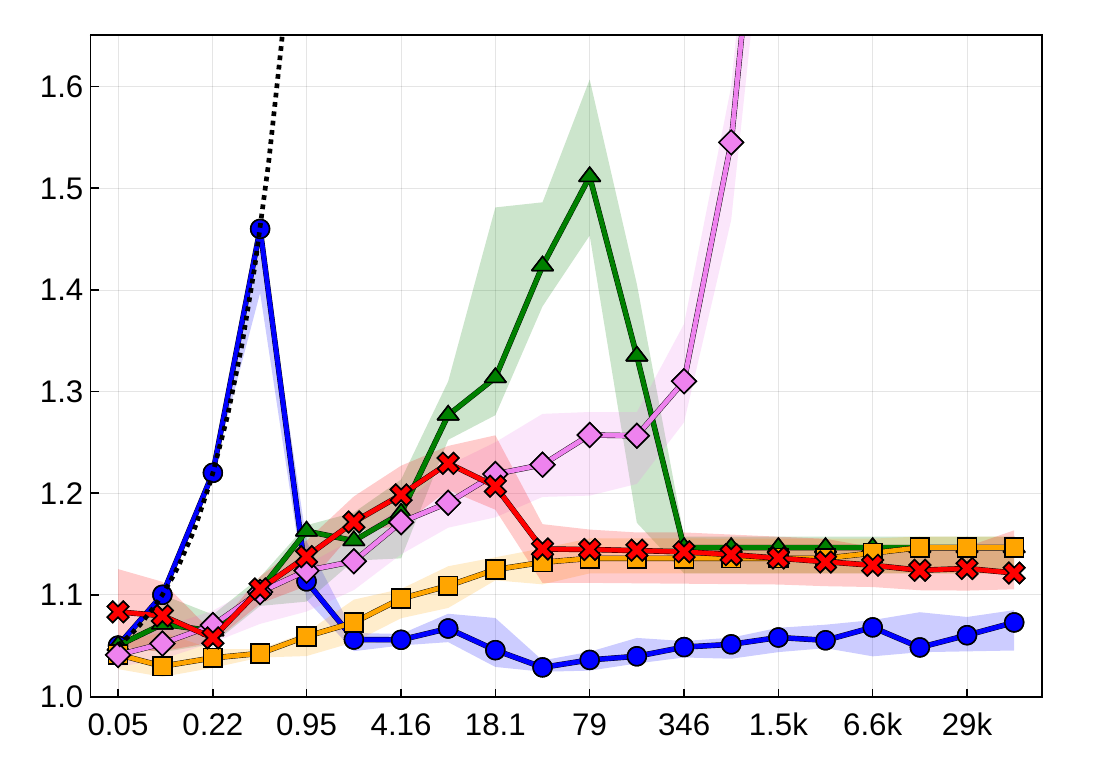}
    \end{subfigure}\\
    \begin{subfigure}[t]{0.45\textwidth}
        \includegraphics[width=\linewidth]{figures/EGP/RTE/RTE_s=4_rho=09_noLegend.pdf}
    \end{subfigure}
    \begin{subfigure}[t]{0.45\textwidth}
        \includegraphics[width=\linewidth]{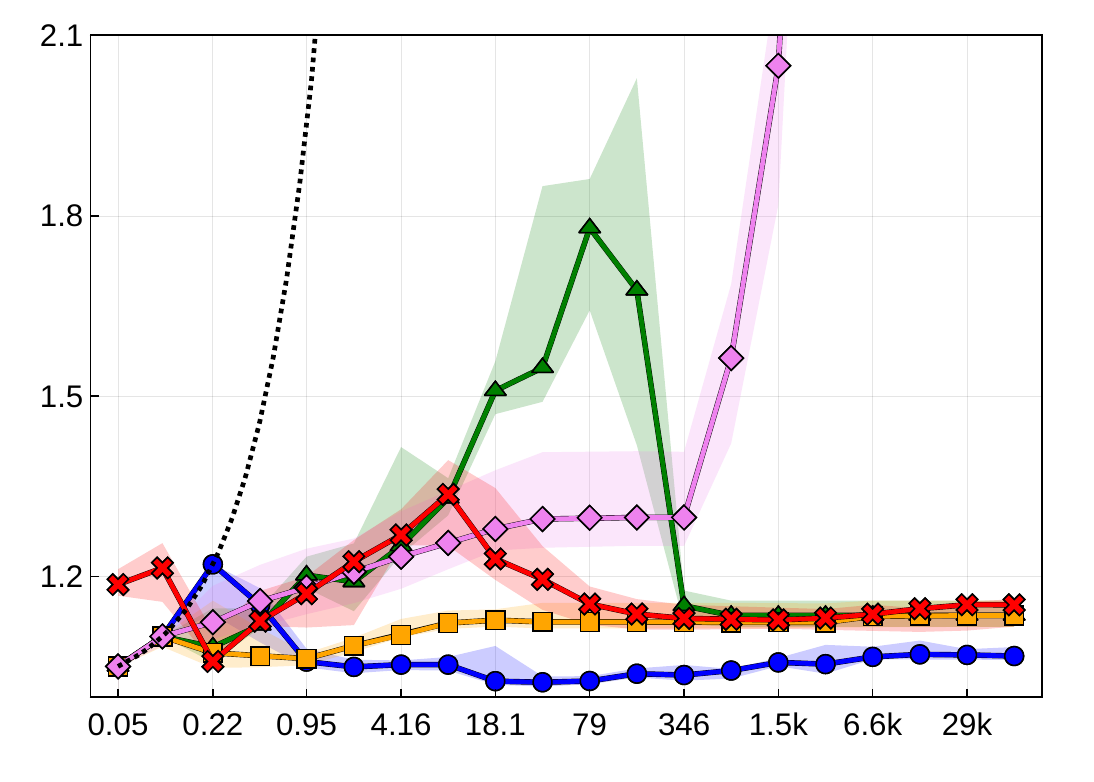}
    \end{subfigure}
    \caption{RTE as a function of SNR. The 4 rows correspond to the 4 settings described in \eqref{eq:settings}. (Auto)correlation level $\rho=0.9$. Left column: EGP with $\ell_1$ regularization. Right column: EGP without $\ell_1$ regularization. Medians over $10$ runs with quantile range between $0.3$ and $0.7$. Legend holds for all subfigures.}
    \label{fig:L1_ablation_4}
\end{figure}

Figures \ref{fig:L1_ablation_1}, \ref{fig:L1_ablation_2}, \ref{fig:L1_ablation_3} and \ref{fig:L1_ablation_4}  show the results for 4 different (auto)correlation levels $\rho$. The left and right sides of the figures respectively show the results with and without $\ell_1$ regularization.

One can observe that $\ell_1$ regularization can sometimes be helpful in the very-high-noise regime but overall the results are very similar to the simulations that perform combined ($\ell_0$  and $\ell_1$ ) regularization.

\begin{figure}[ht]
    \centering
    \begin{subfigure}[t]{0.45\textwidth}
        \includegraphics[width=\linewidth]{figures/EGP/RTE/RTE_s=1_rho=00_noLegend.pdf}
    \end{subfigure}
    \begin{subfigure}[t]{0.45\textwidth}
        \includegraphics[width=\linewidth]{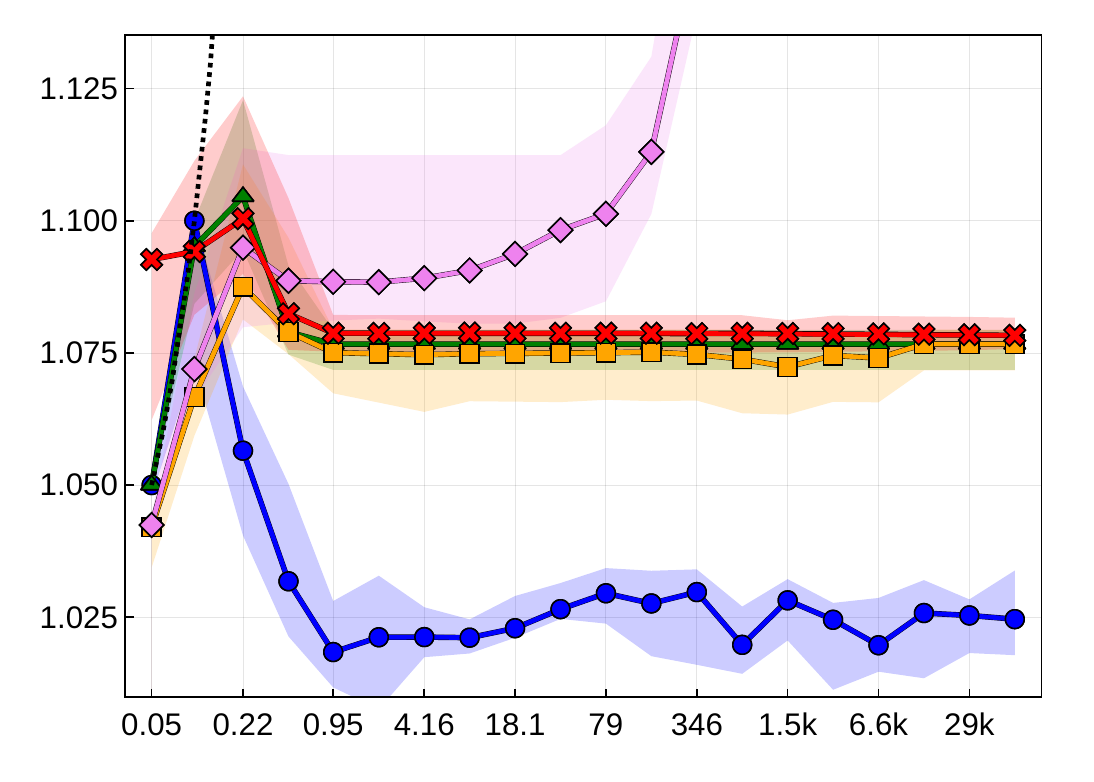}
    \end{subfigure}\\
    \begin{subfigure}[t]{0.45\textwidth}
        \includegraphics[width=\linewidth]{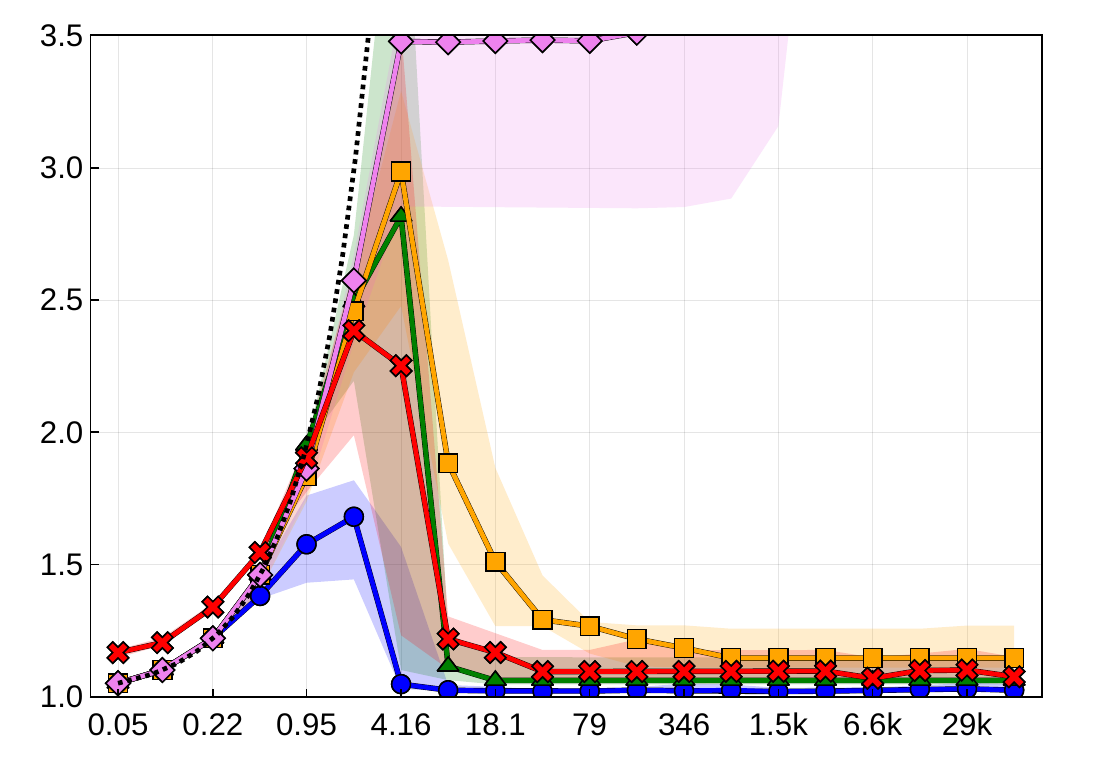}
    \end{subfigure}
    \begin{subfigure}[t]{0.45\textwidth}
        \includegraphics[width=\linewidth]{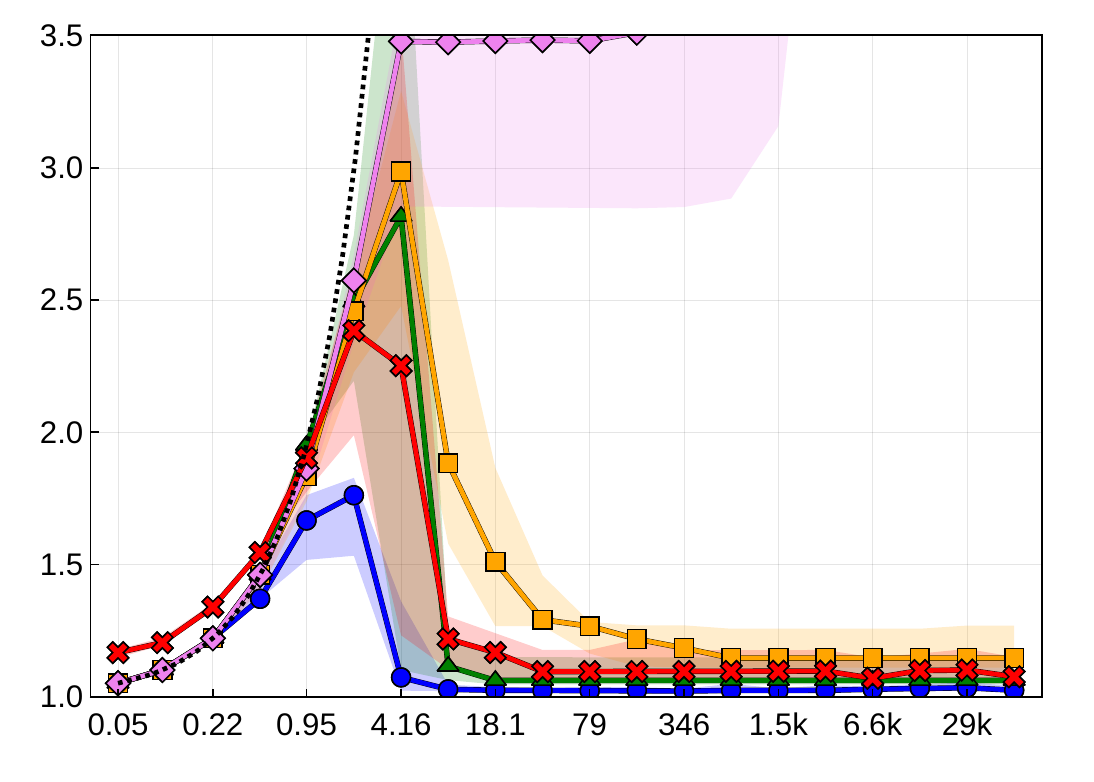}
    \end{subfigure}\\
    \begin{subfigure}[t]{0.45\textwidth}
        \includegraphics[width=\linewidth]{figures/EGP/RTE/RTE_s=3_rho=00_noLegend.pdf}
    \end{subfigure}
    \begin{subfigure}[t]{0.45\textwidth}
        \includegraphics[width=\linewidth]{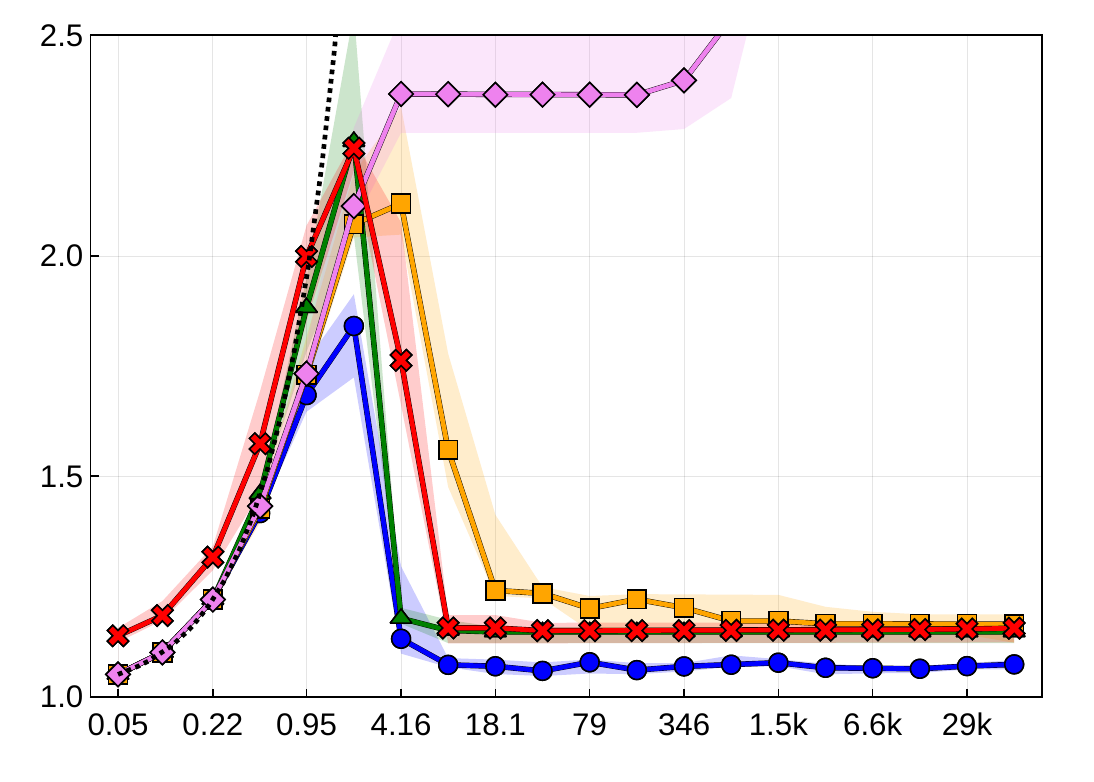}
    \end{subfigure}\\
    \begin{subfigure}[t]{0.45\textwidth}
        \includegraphics[width=\linewidth]{figures/EGP/RTE/RTE_s=4_rho=00_noLegend.pdf}
    \end{subfigure}
    \begin{subfigure}[t]{0.45\textwidth}
        \includegraphics[width=\linewidth]{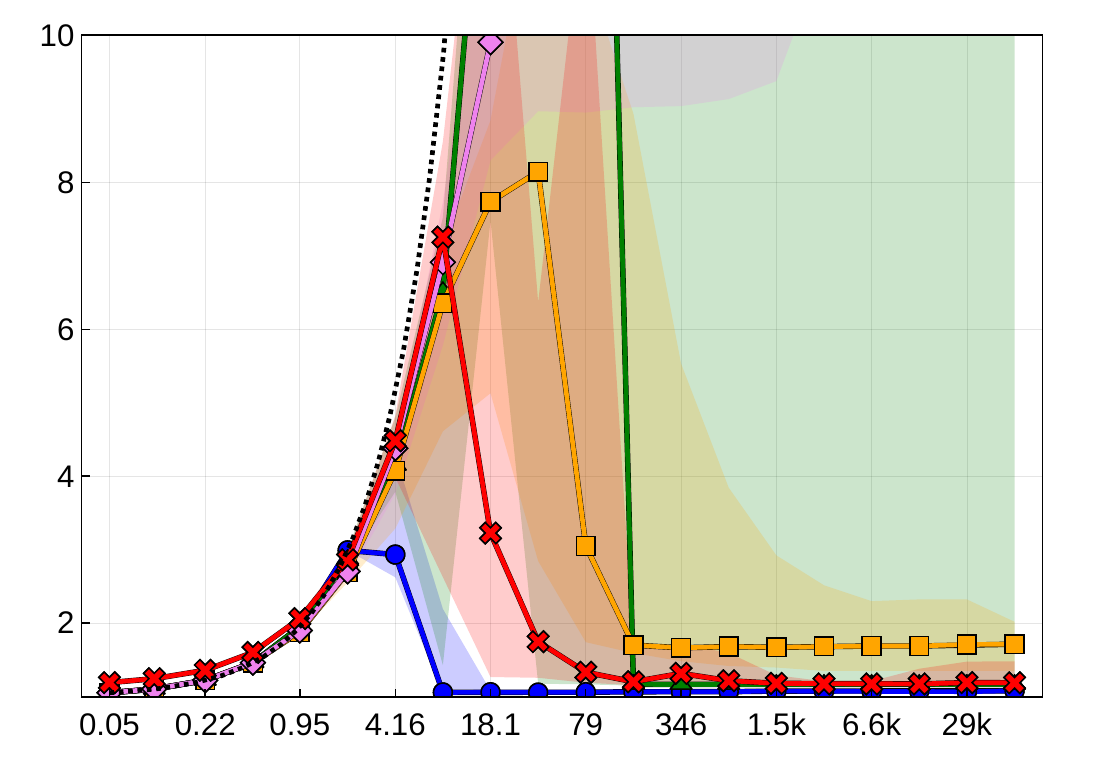}
    \end{subfigure}
    \caption{RTE as a function of SNR. The 4 rows correspond to the 4 settings described in \eqref{eq:settings}. (Auto)correlation level $\rho=0.0$. Left column: EGP with $\ell_1$ regularization. Right column: EGP without $\ell_1$ regularization. Medians over $10$ runs with quantile range between $0.3$ and $0.7$. Legend holds for all subfigures.}
    \label{fig:L1_ablation_1}
\end{figure}

\end{document}